\documentclass[journal]{IEEEtran}
\usepackage{amsmath}
\usepackage{algorithm}
\usepackage{algorithmic}
\usepackage{graphicx}
\usepackage{multirow}
\usepackage{tabularx}
\usepackage{xcolor}
\usepackage{array}
\usepackage{amssymb}
\usepackage{bbding}
\usepackage{subfigure}
\usepackage{diagbox}
\usepackage{rotating}
\usepackage{bm}
\usepackage{tikz} 
\usepackage{booktabs}
\usepackage{cite}
\usepackage{CJK}
\usepackage{overpic}
\usepackage[colorlinks,
            linkcolor=red,
            anchorcolor=blue,
            citecolor=green
            ]{hyperref}

\newcommand{\addFig}[1]{}
\newcommand{\addFigs}[1]{}

\usepackage{subfigure}
\newcommand{\etal}{\textit{et~al}.~}
\newcommand{\ie}{\textit{i}.\textit{e}.,~}
\newcommand{\eg}{\textit{e}.\textit{g}.,~}

\usepackage{balance}
\usepackage{cleveref}
\crefformat{section}{\S#2#1#3} 
\crefformat{subsection}{\S#2#1#3}
\crefformat{subsubsection}{\S#2#1#3}

\ifCLASSINFOpdf
\else
\fi

\begin{document}
%
\title{Context-Aware Interaction Network for RGB-T Semantic Segmentation}
%
%
%

\author{Ying~Lv,
Zhi~Liu,~\IEEEmembership{Senior Member,~IEEE},
Gongyang~Li

\thanks{This work was supported in part by the National Natural Science Foundation of China under Grant 62171269 and in part by the China Postdoctoral Science Foundation under Grant 2022M722037. \textit{(Corresponding author: Zhi Liu.)}}
\thanks{All authors are with the Key Laboratory of Specialty Fiber Optics and Optical Access Networks, Joint International Research Laboratory of Specialty Fiber Optics and Advanced Communication, Shanghai Institute for Advanced Communication and Data Science, School of Communication and Information Engineering, Shanghai University, Shanghai 200444, China. Zhi Liu and Gongyang Li are also with Wenzhou Institute of Shanghai University, Wenzhou 325000, China (e-mail: yinglv@shu.edu.cn; liuzhisjtu@163.com; ligongyang@shu.edu.cn).}
}

\markboth{IEEE TRANSACTIONS ON Multimedia}%
{Shell \MakeLowercase{\textit{et al.}}: Bare Demo of IEEEtran.cls for IEEE Journals}

\maketitle
\begin{abstract}
RGB-T semantic segmentation is a key technique for autonomous driving scenes understanding. 
For the existing RGB-T semantic segmentation methods, however, the effective exploration of the complementary relationship between different modalities is not implemented in the information interaction between multiple levels.
To address such an issue, the Context-Aware Interaction Network (CAINet) is proposed for RGB-T semantic segmentation, which constructs interaction space to exploit auxiliary tasks and global context for explicitly guided learning.
Specifically, we propose a Context-Aware Complementary Reasoning (CACR) module aimed at establishing the complementary relationship between multimodal features with the long-term context in both spatial and channel dimensions.
Further, considering the importance of global contextual and detailed information, we propose the Global Context Modeling (GCM) module and Detail Aggregation (DA) module, and we introduce specific auxiliary supervision to explicitly guide the context interaction and refine the segmentation map. 
Extensive experiments on two benchmark datasets of MFNet and PST900 demonstrate that the proposed CAINet achieves state-of-the-art performance. The code is available at https://github.com/YingLv1106/CAINet.
%
\end{abstract}


\IEEEpeerreviewmaketitle

\section{Introduction}
\begin{figure}[t!]
  \centering
  \footnotesize
  \begin{overpic}[width=0.95\columnwidth]
  {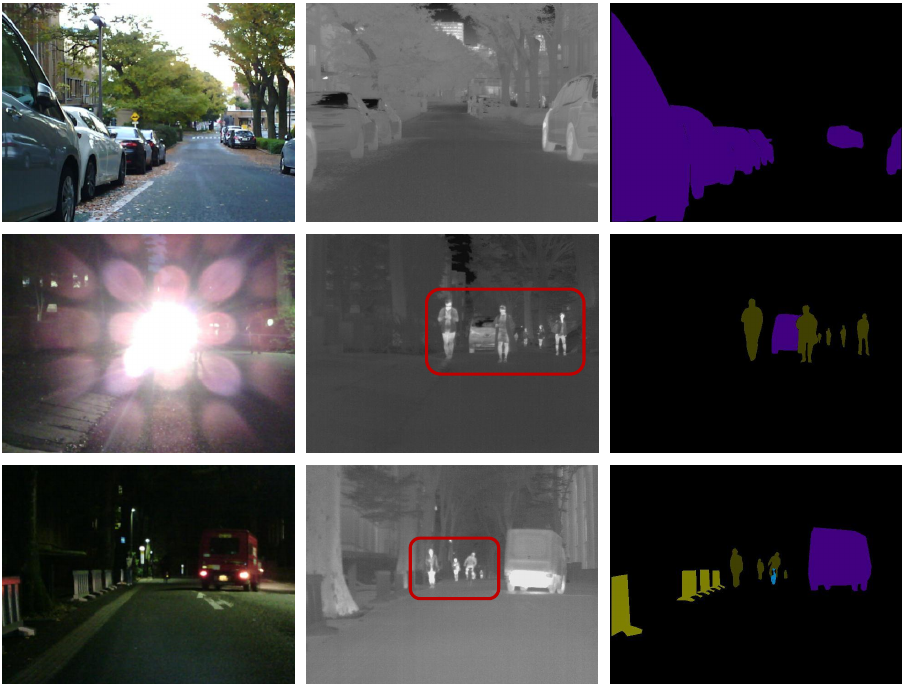}
  \end{overpic}
    \begin{tabular}{ccc}
     RGB \quad\quad\quad\quad\quad\quad Thermal \quad\quad\quad\quad\quad\quad\quad         GT \\
    \end{tabular}
    \caption{Samples of RGB-T semantic segmentation from MFNet~\cite{2017MFNet} dataset, which contains daytime, bright light and nighttime images from top to bottom.}
\label{fig:example}
\end{figure}

\IEEEPARstart{S}{emantic} segmentation~\cite{2015FCN,DeepLabV1,17SegNet,2021Segformer,gao2022fbsnet,9362263,2022fewmetalearn,ma2022learning,zhao2023lif} is a fundamental computer vision task that assigns a class to each pixel in an image, and thus forms dense semantic regions and provides scene understanding for many real-world applications such as autonomous driving~\cite{1autonomous,2autonomous}, robotic manipulation~\cite{ainetter2021end}, medical diagnosis~\cite{2015Unet}, and virtual reality~\cite{VirtualReality}.
Recently, semantic segmentation performance has been improved significantly~\cite{DeepLabV3+,2021Segformer,cheng2021maskformer,lgysod}. However, unimodal methods perform poorly in realistic driving scenarios,~\eg cluttered backgrounds, distant small targets, and unfavorable lighting conditions (dazzle lamps or even darkness), as shown in Fig.~\ref{fig:example}. 
It is difficult to identify pedestrians and vehicles from RGB images, but they can be clearly identified from thermal images. RGB-based semantic segmentation is inadequate, and to overcome this problem, thermal images are introduced. However, this brings new challenges, such as the effective fusion of cross-modal information for RGB-T semantic segmentation. 

\begin{figure}
\centering
\footnotesize
  \begin{overpic}[width=1\columnwidth]{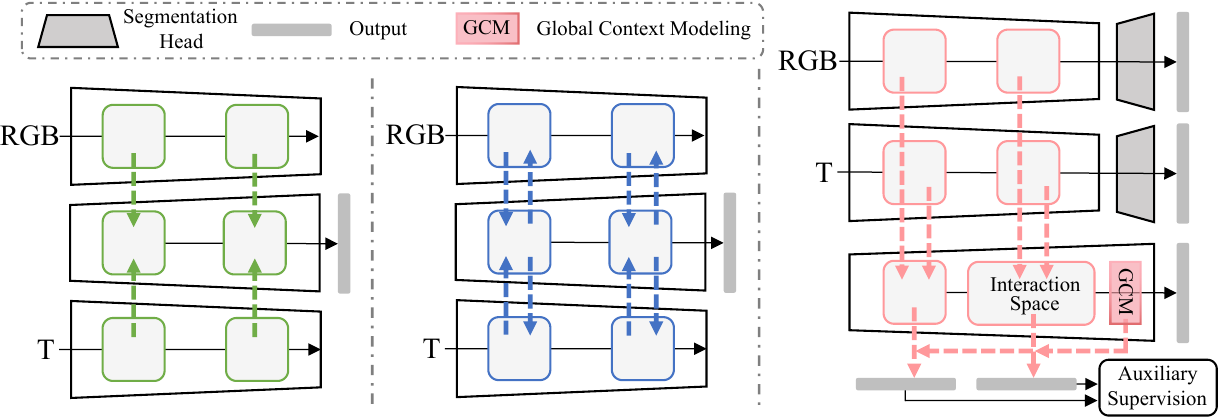}
  \end{overpic}
    \begin{tabular}{ccc}
    (a)\quad\quad\quad\quad\quad
    \quad\quad\quad
    (b) \quad\quad\quad\quad\quad
    \quad\quad\quad\quad
    (c)  
    \end{tabular}
    
\caption{Three typical categories of architectures in RGB-T semantic segmentation, including 
            a) direct fusion~\cite{2017MFNet,
                    2019RTFNet,2020PSTNet,
                    2021MLFNet,2021FuseSeg,2021ABMDRNet,
                    2022EGFNet,2021FEANet,2022MTANet,
                    2021GMNet,zhou2023embedded,2023LASNet,2022MFFENet},
            b) feedback fusion~\cite{2022CCFFNet,
        liu2022cmx,zhou2022multispectral}, and c) our proposed context-aware interaction fusion network.}
    \label{fig:FusionStrategy}
\end{figure}

%
The mainstream RGB-T semantic segmentation methods can be divided into two categories (as shown in Fig.~\ref{fig:FusionStrategy}), \ie direct fusion~\cite{2017MFNet,2019RTFNet,2020PSTNet,2021MLFNet,2021FuseSeg,2021ABMDRNet,2022EGFNet,2021FEANet,2022MTANet,2021GMNet,2023LASNet,zhou2023embedded,2022MFFENet} and feedback fusion~\cite{zhou2022multispectral,2022CCFFNet,liu2022cmx}.
The first category (Fig.~\ref{fig:FusionStrategy}(a)) consists of two backbones that are used to extract features separately. Then, specific modules are designed for direct fusion independent of the backbones. 
The second category (Fig.~\ref{fig:FusionStrategy}(b)) differs from the first one in that fusion features independent of the backbone are fed back to the respective backbone, completing the interaction between the encoder and the fusion layer at multiple levels, respectively.
Both of the above fusion paradigms have greatly explored the fusion of features based on RGB and thermal modalities.

However, both categories have their shortcomings. Most direct fusion models use the same module for feature fusion at different levels, which neglects the characteristics of different levels. Meanwhile, level-by-level fusion is not conducive to information interaction between levels. 
Although the feedback fusion models complete the feature interaction, the feedback of interactive information to the backbone leads to information bottleneck~\cite{2018InformBottle}, which makes it impossible to recur shallow features to deep levels of the model to form high-level semantic information in the feature extraction stages. 
In addition, the effective exploration of complementary relationship between different modalities is not implemented for information interaction between multiple levels. Moreover, the supervision is only set at the end of the above two categories of models, and there is no explicit guiding supervision in the multi-level feature interaction thus making the representation of multi-level fused features ambiguous and uncertain.

To mitigate the above shortcomings, we propose a Context-Aware Interaction Network (CAINet) for RGB-T semantic segmentation, aiming to achieve an effective exploration of the complementary relationship between different modalities for the information interaction between multiple levels. We design a new fusion paradigm to implement CAINet by combining the advantages of both direct fusion and feedback fusion paradigms, as shown in Fig.~\ref{fig:FusionStrategy}(c). Since multiple low-level features can provide boundary detail features for refining the segmentation results, we also fully explore them in our CAINet. In addition, we adopt different auxiliary supervision in CAINet to improve the feature representation at multiple levels.


Speciﬁcally, we propose the Context-Aware Complementary Reasoning (CACR) module to exploit complementary relationship between multiple modalities and their long-range dependencies along spatial and channel dimensions. Then the proposed Global Context Modeling (GCM) module provides global context to guide multi-level feature interactions.
The contextual complementary information and global guidance are obtained from the CACR and GCM modules at three high-level stages. 
In order to explicitly convey the complementary information of multiple modalities, we assign specific auxiliary supervision to multi-level features as shown in Fig.~\ref{fig:FusionStrategy}(c), which has the advantage of explicitly guiding each level of feature representation. Hence long-range dependencies and multimodal complementary information are fully utilized with the cooperation of CACR and GCM.
%
%
%
Moreover, the Detail Aggregation (DA) module is proposed to explore boundary information and binary mask, which can solve some problems,~\ie binary mask supervision to alleviate the interference of cluttered background and boundary supervision to further improve performance by performing boundary detail refinement.
Thanks to all this, our model achieves state-of-the-art performance on MFNet and PST900 datasets.
%
%

Our main contributions are summarized as follows:
\begin{itemize}
\item We fully explore the complementary relationship between high-level multimodal features and propose a novel context-aware interaction fusion paradigm to implement CAINet, which achieves state-of-the-art performance on MFNet and PST900 datasets.
\item We propose the CACR module for establishing complementary relationship between multimodal features with long-term dependence in spatial and channel dimensions, GCM module to explore global context clues to provide global guidance for feature interactions, and DA module to aggregate detailed features to further promote segmentation performance.
\item We introduce auxiliary supervision
 and residual learning into CAINet. The auxiliary supervision ensures explicit guidance step by step and the residual learning retains the high-level global context information.

\end{itemize}

The rest of this paper is organized as follows.
Section~\ref{sec:related} explains the related work on RGB and RGB-T semantic segmentation. The proposed CAINet is elaborated in Section~\ref{sec:OurMethod}.
Section~\ref{sec:exp} presents the experimental results and analysis of ablation experiments.
Finally, Section~\ref{sec:con} concludes this paper.

\section{Related Work}
\label{sec:related}

\subsection{RGB Semantic Segmentation}
\label{sec:Tra_ORSI_SOD}
Since the introduction of the fully convolutional network (FCN) for semantic segmentation, various CNN-based semantic segmentation models have been proposed and achieved satisfactory performance. However, there are limitations in the receptive field. Many works attempted to alleviate this weakness.
For instance, Chen~\etal\cite{DeepLabV2} proposed a deep convolutional network, which utilizes multiple sampling rates of dilated convolution in atrous spatial pyramid pooling. Similarly, Zhao~\etal\cite{2017PSP} proposed a pyramid scene parsing network that aggregates contextual information at different scales.

Sequentially, many semantic segmentation methods integrate attention mechanisms into the convolutional structure. Wang~\etal\cite{2018NLNet} proposed the non-local network, which leverages weighted sum to aggregate the features of all points to obtain global features. 
Hu~\etal\cite{SENet} introduced a novel technique called SENet, which incorporates global context information to adjust channel dependencies and rescale channel weights.
Woo~\etal\cite{2018CBAM} proposed the convolution-based attention module, which splits the attention process into two independent parts, namely, the channel attention and spatial attention modules. 
Cao~\etal\cite{cao2019gcnet} investigated a new global context modeling framework known as GCNet, which can  build long-range dependencies while being computationally efficient.
Fu~\etal\cite{20DANet} presented a dual attention network to capture feature dependencies in both the spatial and channel dimensions. 

Following the introduction of visual transformers, subsequent advancements in semantic segmentation have led to the emergence of transformer-based techniques.
Zheng~\etal\cite{Zheng2021SETR} adopted an encoder with transformer structure to tackle semantic segmentation as a sequence-to-sequence prediction task. 
Xie~\etal\cite{2021Segformer} introduced a hierarchical structure transformer encoder combining both local and global attention mechanisms to produce powerful representations. 
Cheng~\etal\cite{cheng2021maskformer} proposed a mask classification paradigm that leverages both instance segmentation and semantic segmentation tasks. 
Furthermore, Zhang~\etal\cite{zhang2022segvit} transferred a similar mapping strategy between a set of learnable class tokens and spatial feature mappings to segmentation masks. 
Lastly, Kim~\etal\cite{kim2023VTM} presented a generalized sample-less learner for arbitrarily dense prediction tasks called Visual Token Matching (VTM).
Overall, these unimodal segmentation methods have demonstrated their efficacy in generating powerful representations for semantic segmentation.  However, in situations with reduced visibility or obstacles such as fog, rain, snow, or dust caused by adverse weather conditions, as well as in low-light or no-light conditions like nighttime or inside tunnels, visible light imaging systems may struggle to effectively detect objects against the background. This is where the supplementation of infrared technology becomes necessary.

\subsection{RGB-T Semantic Segmentation}
\label{sec:CNN_ORSI_SOD}
Although the RGB semantic segmentation methods show superior performance, they still perform poorly when encountering scenes with extreme lighting conditions, as depicted in Fig.~\ref{fig:example}. 
So many researchers additionally introduce thermal images to improve the segmentation performance. 
Ha~\etal\cite{2017MFNet} proposed MFNet as a pioneer of RGB-T semantic segmentation, which uses a two-stream structure for feature extraction and then fuses these features using concatenation.
RTFNet~\cite{2019RTFNet} and FuseSeg~\cite{2021FuseSeg} adopted a comparable architecture as MFNet, albeit fusing features through an element-wise summation.
Shivakumar~\etal\cite{2020PSTNet} proposed a two-stage structure and fused the output of the first stage with thermal and color images.
However, these methods almost used simple fusion strategies, \ie the element-wise summation and concatenation are used to capture cross-modal features, which may lead to information redundancy by ignoring the differences between cross-modal information.

Many studies have attempted to design specialized feature fusion operations.
Guo~\etal\cite{2021MLFNet} proposed a multi-stage skip connection between the encoder and decoder.
Zhang~\etal\cite{2021ABMDRNet} introduced a strategy of 
bridging-then-fusing to reduce multimodal differences and achieved multi-scale contextual feature fusion.
Lan~\etal\cite{2022MMNet} introduced a two-stage multimodal multi-stage network (MMNet) to extract features from different modalities separately and the second stage performed feature fusion and refinement.
Wang~\etal\cite{wang2023facedark} proposed a double fusion embedded learning approach to force each task to adapt to the feature representation of the other tasks.
Deng~\etal\cite{2021FEANet} introduced channel and spatial attention modules in the encoder to enhance the feature fusion. 
%
%
Zhou~\etal\cite{2021GMNet,2022GCNet,2022EGFNet,2023GCGLNet,2022MFFENet,2022MTANet} fused cross-modal features in a specially designed module and binary boundary-assisted supervision for feature refinement.
%
%
Yi~\etal\cite{2022CCAFFMNet} proposed an attentional feature fusion network of channel coordinates to obtain channel and coordinate correlations between multimodal features.
Xu~\etal\cite{2022DSGBINet} introduced  graph convolution to fuse multimodal high-level features from a global and semantic perspective and also uses multitasking supervision.
Zhao~\etal\cite{2022FDCNet} performed unimodal feature extraction separately, spatial attention in shallow layer, shared convolutional blocks in the high layer, and channel attention to fuse them.
%
%
Frigo~\etal\cite{2022DooDLeNet} explored match correlation and confidence weights to explicitly estimate feature matches for decoders by computing feature correlations. And confidence weights are used for multimodal feature fusion.
Wang~\etal\cite{wang2023distribution,wang2023incomplete} proposed modality recovery methods that effectively predict the distribution-consistent space of missing modalities, achieving consistency in multimodal distributions and semantic disambiguation.

Fu~\etal\cite{fu2022cgfnet} presented the attention mechanism to fuse multi-level features and added the extracted global information to the decoding stage.
Feng~\etal\cite{2023CENet} introduced a teacher-student distillation model that used a boundary-assisted teacher model to learn cross-modal features and a student model to learn only thermal image features.
Cai~\etal\cite{2023DHFNet} proposed a two-branch network enhanced by boundary refinement and guided foreground background features and fused with high-level features to preserve semantic information.
Zhou~\etal\cite{zhou2023embedded} proposed an embedded control gate fusion paradigm, and developed an attention residual learning strategy to ensure effective instruction of high-level features.
Li~\etal\cite{2023LASNet} treated multi-level features differently,~\ie the high-, middle- and low-level semantic features were used to localize, activate target regions, and refine boundary, respectively.
%
%

Above mentioned feature fusion method may ignore the interaction of hierarchical features and cannot sufficiently explore multimodal complementary information. In recent years, some interaction fusion methods have been proposed.
Zhou~\etal\cite{2022MFTNet} proposed a transformer-based framework to handle the correlation and complementation between multimodal, and later return the fused features to their respective backbone networks. 
This method differs from the feature fusion method mentioned above as shown in Fig.~\ref{fig:FusionStrategy}{(a)}, which feeds the fused features to their respective backbones.
%
Wu~\etal\cite{2022CCFFNet} proposed a complementary-aware cross-modal feature fusion network (CCFFNet) to select complementary information from multimodal features and fused them through a channel weighting mechanism. However, feature fusion in spatial details is not sufficient.
Liu~\etal\cite{liu2022cmx} proposed a vision-transformer-based cross-modal feature rectify fusion network that exploited global max pooling in terms of channel and spatial interaction, but the pooling layer may lead to loss of spatially detailed features.
The complementary relationship between the two modalities may not be well-exploited by these models. 

Aiming to effectively explore multimodal complementary information, we combine the benefits of two fusion paradigms,~\ie direct fusion and feedback fusion, we propose a novel fusion paradigm that builds an interaction space for reasoning about complementary relationship between different modalities at multi-level stages. And we propose CACR, GCM, and 
DA for exploring complementary relationship, harvesting global
contextual cue, and refining segmentation results,
respectively. Furthermore, auxiliary supervisions assist high-level contextual information to guide feature interaction in an explicit manner. Based on the above, we propose the Context-Aware Interaction Network, \ie CAINet, for RGB-T semantic
segmentation.
%

\begin{figure*}
	\centering
	\begin{overpic}[width=0.95\textwidth]{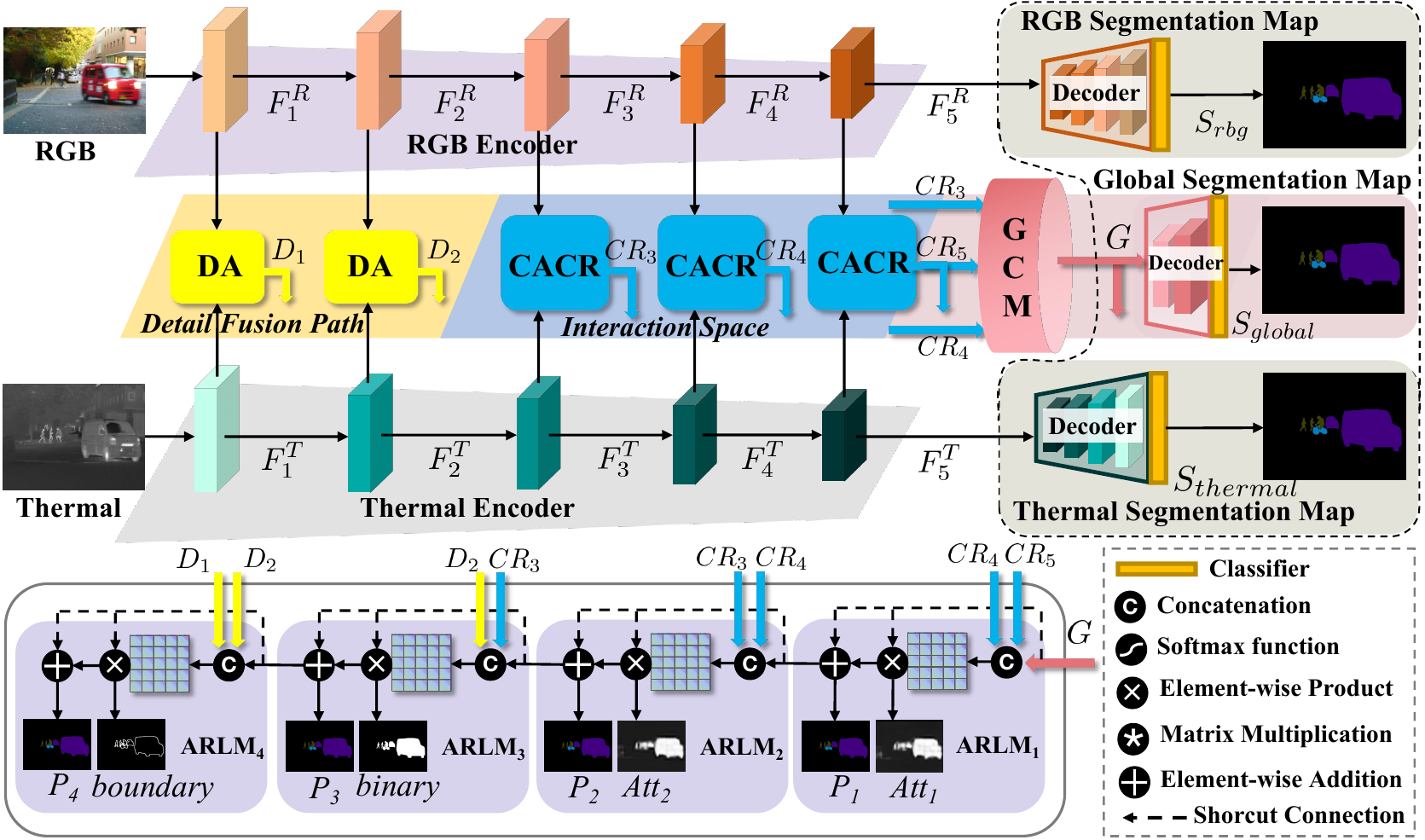}
    \end{overpic}
	\caption{The overview of proposed CAINet.  
 Specifically, CAINet consists of six components including RGB and thermal encoders, interaction space reasoning, global context modeling, three-decoder supervision, detailed feature fusion, and residual learning of multiple auxiliary tasks supervision. The residual learning~\cite{zhou2023embedded} (ARLM) module is to assist context-aware complementary reasoning (CACR) and global context modeling (GCM) to implement multimodal feature interaction; the detail aggregation (DA) module refines the final segmentation map. During inference, we can remove the other supervised branches and retain the final predictive semantic segmentation map $P_4$, suggesting that the performance enhancement comes with no added inference cost.
}
    \label{fig:Framework}
\end{figure*}
%
\section{Proposed Method}
\label{sec:OurMethod}
The main pipeline of our Context-Aware Interaction Network (CAINet) can be divided into five stages: RGB and thermal stream encoder-decoder, context-aware  complementary reasoning (Section~\ref{sec:CACR}), global context modeling (Section~\ref{sec:GCM}), detail aggregation (Section~\ref{sec:DAM}) and residual learning with auxiliary supervision (Section~\ref{sec:RLATT}). The overall framework is shown in Fig.~\ref{fig:Framework}.

\subsection{Architecture Overview}
\label{sec:Overview}
%
We employ two MobileNet-V2~\cite{MobileNet2} networks pretrained on ImageNet~\cite{deng2009imagenet} as the backbone. As suggested in ECGFNet~\cite{zhou2023embedded}, we remove the last down-sampling operations, hence enlarging the resolution of the feature map, so the output stride becomes 16 instead of 32. Let the multistage RGB and thermal features obtained from the $i$-th ($i\in\{1,2,3,4,5\})$ block of MobileNet-V2 be $F^{R}_{i}$ and $F^{T}_{i}$, respectively. 
They are followed by the decoder layers and independently supervised to obtain the coarse segmentation maps denoted as $S_{rgb}$ and $S_{thermal}$, respectively.
To establish a multimodal complementary relationship for RGB and thermal features with long-term dependencies in both spatial and channel dimensions, we use the CACR module at three levels (\ie $F^{R}_{3-5}$ and $F^{T}_{3-5}$). Additionally, we introduce the GCM module to explore global context clues and provide global guidance for feature interactions. Following the GCM module is a decoder to generate a global coarse segmentation map $S_{global}$.
The DA module aggregates detailed features and further improves segmentation performance.
We employ the Attention Residual Learning Module (ARLM)~\cite{zhou2023embedded} to establish a upsample stream with auxiliary supervision and target supervision running in parallel. This upsample stream consists of four stages, where the auxiliary supervision is carried out sequentially for attention, binary, and boundary supervision. In contrast, the target supervision exclusively focuses on semantic segmentation supervision. Ultimately, through residual learning, we obtain progressively refined semantic segmentation maps. The predicted segmentation map $P_4$ from the fourth stage serves as the final segmentation result.
%
\subsection{Context-Aware Complementary Reasoning}
\label{sec:CACR} 

There exists a complementary relationship between multimodal semantic regions, and how to effectively tap this complementary relationship is crucial for multimodal information fusion~\cite{2022DSGBINet}. We propose the CACR module to construct an interaction space, which reasons about the complementary relationship on a long range of the interaction space, including the spatial and channel dimensions. 
Specifically, the CACR module views the inputs as a set of features (\ie pixels) with the corresponding feature dimensions (\eg spatial structure, color, surface temperature, and local and global texture), which enable the exchange of long-range contexts, enhancing multimodal features at the local level and global level simultaneously. 
We apply the CACR module in three high-level stages of our model, \ie $\{F^{R}_{3-5}, F^{T}_{3-5}\}$, and their outputs are $CR_3$, $CR_4$, and $CR_5$, respectively, as shown in Fig.~\ref{fig:Framework}.
%
%
\begin{figure*}
	\centering
	\begin{overpic}[width=0.95\textwidth]{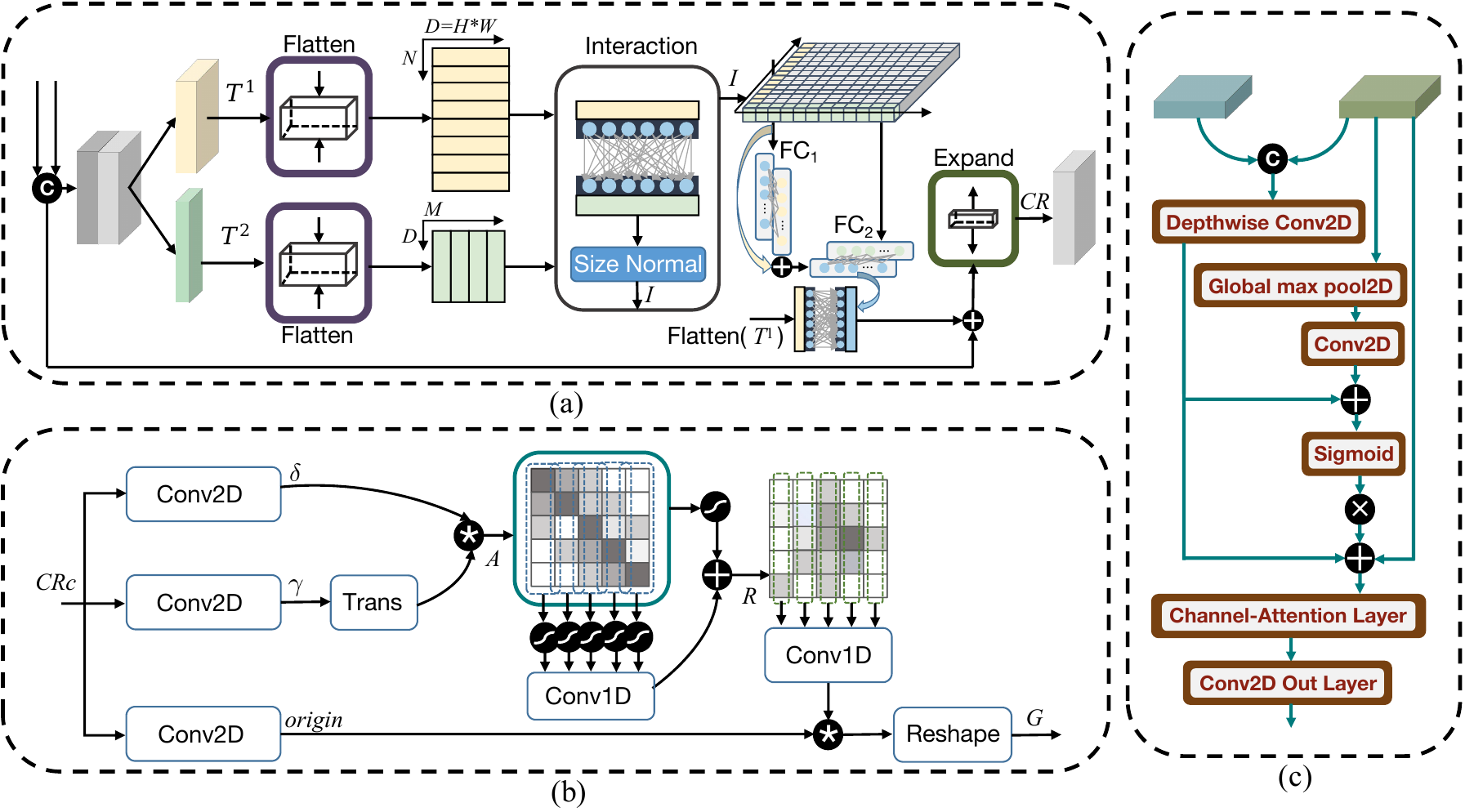}
    \end{overpic}
\caption{Illustration of a) the Context-Aware Complementary Reasoning (CACR) module, b) the Global Context Modeling (GCM) module, where $\delta$, $\gamma$, and $origin$ represent feature maps from different convolution layers and c) the Detail Aggregation (DA) module.
    }
    \label{fig:block}
\end{figure*}

The detailed structure of the CACR module is shown in Fig.~\ref{fig:block}(a). The inputs to the CACR module are the high-level features $F_{3}^{R/T}$, $F_{4}^{R/T}$, and $F_{5}^{R/T}$. Given an input feature $\mathbf{X}\in\mathbb{R}^{C\!\times\!H\!\times\!W}$ converted to a collection of features $\mathbf{P}\in\mathbb{R}^{C\!\times\!D}$, where $D = H\!\times\!W$ is the number of features, and each channel contain information. The image of $C\!\times\!H\!\times\!W$ can only provide two-dimensional information of objects and cannot be used for geometry analysis directly. 
In contrast, the features $C\!\times\!D$ represent the spatial and attribute information of the feature~\cite{ma2023setpoint}, so it can be easily used for geometry analysis and processing. For context interaction space transformation, we let the model learn a transfer function $\boldsymbol{f}(\cdot)$:
\begin{equation}
    T(X)=\boldsymbol{f}(X;\omega)=X\omega,
    \label{eq:1}
\end{equation}
where $\omega$ is the learnable convolution layer, two different dimensional features are then obtained, which are $T^{1}\in\mathbb{R}^{N\!\times\!H\!\times\!W}$ and $T^{2}\in\mathbb{R}^{M\!\times\!H\!\times\!W}$. We set $N=C$ and $M=C/2$. We aim to build a similarity correlation between the regions of two features and further complement multimodal features in the interaction space based on regional correlations:
\begin{equation}
    I=
\frac{ [\operatorname{Flatten}\left(T^2\right)]^T * \operatorname{Flatten}\left(T^1\right)}{N\times M},
    \label{eq:2}
\end{equation}
where Flatten$(\cdot)$ denotes that map these features to feature set space and $*$ denotes the matrix multiplication. We get the similarity correlation features $I\in\mathbb{R}^{M\!\times\!N}$ of the interaction space, which contains the multimodal feature description of each pixel and can capture the relationship between any regions, and then reason about the complete multimodal feature map. 
In practice, ‘Size Normal’ as shown in~Fig.~\ref{fig:block}(a) is to simply divide the feature map by its size $N\times M$ to control the numerical amplitude of the feature map.

%

In particular, the similarity correlation $I$ is then adaptively assigned to each feature of the region in the interaction space. Fig.~\ref{fig:block}(a) shows the implementation of this process and the inference along two different dimensions through two fully connected layers. By doing so, the features can communicate with each other based on complementary relationship between multimodal features, which allows the exchange of long-range contexts. Therefore, the complementary reasoning $CR$ is given as follows: 
\begin{equation}
    CR’=\sigma\left(\left(\mathrm{FC}_2\left(\sigma\left(\mathrm{FC}_1(I)+\I\right)\right)\right)\right),
    \label{eq:c'}
\end{equation}
\begin{equation}
    CR=\operatorname{Expand}(X+\operatorname{Flatten}(T^1)^T * CR’),
    \label{eq:4}
\end{equation}
where $\sigma$ denotes the ReLU activation function, $\mathrm{FC}_1$ represents the fully connected layer along the $N$ dimension, and $\mathrm{FC}_2$ denotes the $M$ dimension. Expand$(\cdot)$ represents to extend these features to restore the $H$ and $W$ dimensions. Here, we follow similar correlations to reason about complementarity and apply two fully connected layers to assign feature dimensions (from the interaction space dimension to the original dimension).  
After this, we obtain three high-level complementary features $CR_3, CR_4$ and $CR_5$.
Specifically, CACR module is different from convolutional blocks and vision transformers~\cite{dosovitskiy2020image}, where the former ones perform a sliding window on the image to extract features from local regions, and the latter ones treat an image as a sequence of patches and extract features via self-attention mechanism~\cite{2021ViT} in a global range, as well as different from the GCN~\cite{chen2019graph} module with edge and adjacency matrix. The CACR module establishes the similarity correlation between the multimodal feature regions and is a novel approach that constructs an interaction space to reason about the complementary relationship, allowing it to capture both local and global contextual information in a flexible and adaptive way.
\subsection{Global Context Modeling}
\label{sec:GCM} 

Many efforts try to overcome the limitation of local convolution operators by introducing a global context, such as global pooling of SENet~\cite{SENet}, ASPP~\cite{2018DenseASPP}, and self-attention mechanism~\cite{2021ViT}. Although we also incorporate a global paradigm, in the proposed approach, we go one step further and perform  higher-level global reasoning on the complementary relationship between multi-level features in the GCM module as shown in Fig.~\ref{fig:block}(b).
The GCM module yields the global context modeling feature $G$, which will be used as a global guide for joint auxiliary supervision with the ARLM. 

Auxiliary and target tasks require global contextual information to provide object location cues, as stated in GLCNet~\cite{zhou2019global}, global information is essential to enhance semantic prediction. The key is to improve the semantic representation of each channel using global context modeling.
To account for this, As shown in Fig.~\ref{fig:block}(b), we first aggregate three high-level complementary features by channel concatenation to produce semantic guidance $CR_c$.
Then, we use $RM$ to represent the relationship modeling, which can be obtained as follows:
\begin{equation}
    A=[\operatorname{Conv2D}(CRc)]^{T} * \operatorname{Conv2D}(CRc), 
\label{eq:6}
\end{equation}
\begin{equation}
RM=\operatorname{Conv1D}\left(\beta\left(A\right)\right)+\frac{\exp (A)}{\sum_j^c \sum_i^c \exp (A)},
\label{eq:7}
\end{equation}
where  $Conv2D$ denotes 2D
convolution layer with a $1\times1$ kernel and 
stride 1, $\beta$ represents the Softmax function along with channel dimension, and $Conv1D$ denotes the 1D convolution with a $1\times1$ kernel and stride 1.
Further, global context  features $G$ can be obtained as follows:
\begin{equation}
\mathrm{G}=\operatorname{Reshape}\left[\operatorname{Conv} 2 \mathrm{D}_3(\mathrm{CRc})\right]^{\mathrm{T}} * \operatorname{Conv} 1 \mathrm{D}_2(\mathrm{RM}),
\label{eq:8}
\end{equation}
where $Reshape$ represents feature dimension reshaping, $Conv2D_3$ denotes a 2D convolution with a $1\times1$ kernel and stride 1, $Conv1D_2$ denotes the 1D convolution with a $1\times1$ kernel and stride 1. And $G$ as the global guidance, combines with residual learning of ARLM to associate auxiliary and target supervision.
In addition, $G$ passes through the decoder to get a global coarse segmentation map after the interaction, which may lose detail information, so next we will construct the DA module for feature map refinement. 
\subsection{Detail Aggregation Module}
\label{sec:DAM}
 The features play different roles at different levels in the convolutional neural network (CNN). The low-level ones (\eg $F_{1}^{R/T}$ and $F_{2}^{R/T}$) contain boundary information and abundant texture.  Inspired by this observation, we propose a simple yet effective DA fusion scheme to well utilize multimodal features in uniting the ARLM module for auxiliary boundary detail supervision, further boosting the performance of our CAINet.
As shown in Fig.~\ref{fig:block}(c), we combine the RGB and thermal features $F^R_i$ and $F^T_i$ ($i=1,2$) by concatenation along the channel direction in the AD module. In order to maintain the performance while greatly reducing computational costs, we use depthwise separable convolution to get the combined features $F^c_i$.
%
%
For the thermal features, we utilize a global max pooling layer and convolution with $7\times7$ kernel size to improve the extraction of spatial information, and here the extracted thermal features are denoted as $T_i^{s}$.
%
%
Finally, the detail aggregation features $D_i$ are obtained by a channel attention layer calculated as follows:
\begin{equation}
    D_i=\operatorname{CA} \left(F^c_i\left(\sigma\left(F^c_i+T^s_i\right)\right)+F^T_i\right),
\label{eq:11}
\end{equation}
where $CA$ denotes the channel attention layer. 
After we obtain the detail aggregation features $D_i$, and through subsequent ARLM steps, we obtained the boundary and binary map, refining the final semantic segmentation result.
\begin{table*}[t!]
  \centering
  \small
  \renewcommand{\arraystretch}{1.4}
  \renewcommand{\tabcolsep}{0.9mm}
  \caption{
  Quantitative comparisons (\%) on the test set of MFNet dataset. The value 0.0 represents that there are no true positives. `-' means that the authors do not provide the corresponding results. The top two results in each column are highlighted in \textcolor{red}{\textbf{red}} and \textcolor{blue}{\textbf{blue}}.
    }
\label{table:QuantitativeResults_MFNet}
  
\begin{tabular}{r|c|cccccccccccccccccc}
\midrule[1pt]    
 \multirow{2}{*}{\normalsize{Methods}} &  \multirow{2}{*}{\normalsize{Type}}
 & \multicolumn{2}{c}{Car} & \multicolumn{2}{c}{Person} & \multicolumn{2}{c}{Bike} & \multicolumn{2}{c}{Curve} & \multicolumn{2}{c}{Car Stop} & \multicolumn{2}{c}{Guardrail} & \multicolumn{2}{c}{Color Cone} & \multicolumn{2}{c}{Bump}  
 & \multirow{2}{*}{\normalsize{mAcc}} & \multirow{2}{*}{\normalsize{mIoU}} \\
 
 \cline{3-18} 
     &  & Acc & IoU & Acc & IoU & Acc & IoU & Acc & IoU & Acc & IoU & Acc & IoU & Acc & IoU & Acc & IoU \\
	     
\midrule[1pt] 
MFNet$_{17}$~\cite{2017MFNet}  & RGB-T & 77.2 & 65.9 & 67.0 & 58.9 & 53.9 & 42.9 & 36.2 & 29.9 & 19.1 & 9.9 & 0.1 & 8.5 & 30.3 & 25.2 & 30.0 & 27.7 & 45.1 & 39.7 \\	
RTFNet50$_{19}$~\cite{2019RTFNet}     & RGB-T & 91.3 & 86.3 & 78.2 & 67.8 & 71.5 & 58.2 & 69.8 & 43.7 & 32.1 & 24.3 & 13.4 & 3.6 & 40.4 & 26.0 & 73.5 & 57.2 & 62.2 & 51.7 \\	
RTFNet152$_{19}$~\cite{2019RTFNet}   & RGB-T & 93.0 & 87.4 & 79.3 & 70.3 & 76.8 & 62.7 & 60.7 & 45.3 & 38.5 & 29.8 & 0.0 & 0.0 & 45.5 & 29.1 & 74.7 & 55.7 & 63.1 & 53.2 \\
PSTNet$_{20}$~\cite{2020PSTNet}  & RGB-T & - & 76.8 & - & 52.6 & - & 55.3 & - & 29.6 & - & 25.1 & - & 15.1 & - & 39.4 & - & 45.0 & - & 48.4 \\	
MLFNet$_{21}$~\cite{2021MLFNet}  & RGB-T & - & 82.3 & - & 68.1 & - & 67.3 & - & 27.3 & - & 30.4 & - & 15.7 & - & \textcolor{blue}{\textbf{55.6}} & - & 40.1 & - & 53.8 \\

FuseSeg$_{21}$~\cite{2021FuseSeg}   & RGB-T & 93.1 & 87.9 & 81.4 & 71.7 & 78.5 & 64.6 & 68.4 & 44.8 & 29.1 & 22.7 & 63.7 & 6.4 & 55.8 & 46.9 & 66.4 & 47.9 & 70.6 & 54.5 \\	
ABMDRNet$_{21}$~\cite{2021ABMDRNet}    & RGB-T & 94.3 & 84.8 & 90.0 & 69.6 & 75.7 & 60.3 & 64.0 & 45.1 & 44.1 & 33.1 & 31.0 & 5.1 & 61.7 & 47.4 & 66.2 & 50.0 & 69.5 & 54.8 \\	
MMNet$_{22}$~\cite{2022MMNet}  & RGB-T & - & 83.9 & - & 69.3 & - & 59.0 & - & 43.2 & - & 24.7 & - & 4.6 & - & 42.2 & - & 50.7 & 62.7 & 52.8 \\	
EGFNet$_{22}$~\cite{2022EGFNet}           & RGB-T & \textcolor{red}{\textbf{95.8}}& 87.6 & 89.0 & 69.8 & 80.6 & 58.8 & \textcolor{blue}{\textbf{71.5}} & 42.8 & 48.7 & 33.8 & 33.6 & 7.0 & 65.3 & 48.3 & 71.1 & 47.1 & 72.7 & 54.8 \\

FEANet$_{21}$~\cite{2021FEANet}          & RGB-T & 93.3 & 87.8 & 82.7 & 71.1 & 76.7 & 61.1 & 65.5 & 46.5 & 26.6 & 22.1 & \textcolor{blue}{\textbf{70.8}} & 6.6 & \textcolor{blue}{\textbf{66.6}} & 55.3 & \textcolor{blue}{\textbf{77.3}} & 48.9 & 73.2 & 55.3 \\								   
MTANet$_{22}$~\cite{2022MTANet} 	& RGB-T & \textcolor{red}{\textbf{95.8}} & 88.1 & \textcolor{blue}{\textbf{90.9}} & 71.5 & 80.3 & 60.7 & \textcolor{red}{\textbf{75.3}} & 40.9 & \textcolor{red}{\textbf{62.8}} & 38.9 & 38.7 & 13.7 & 63.8 & 45.9 & 70.8 & 47.2 & \textcolor{blue}{\textbf{75.2}} & 56.1 \\		
MFFENet161$_{21}$~\cite{2022MFFENet}  & RGB-T & 93.1 & 88.2 & 83.2 & 74.1 & 77.1 & 62.9 & 67.2 & 46.2 & 52.3 & 37.1 & 65.0 & 7.6 & 58.5 & 52.4 & 73.4 & 47.4 & 74.3 & 57.1 \\	
GMNet$_{21}$~\cite{2021GMNet}           & RGB-T & 94.1 & 86.5 & 83.0 & 73.1 & 76.9 & 61.7 & 59.7 & 44.0 & 55.0 & \textcolor{blue}{\textbf{42.3}} & \textcolor{red}{\textbf{71.2}} & 14.5 & 54.7 & 48.7 & 73.1 & 47.4 & 74.1 & 57.3 \\	

CCFFNet101$_{22}$~\cite{2022CCFFNet}           & RGB-T & 94.5 & \textcolor{red}{\textbf{89.6}} & 83.6 & \textcolor{blue}{\textbf{74.2}} & 73.2 & 63.1 & 67.2 & \textcolor{blue}{\textbf{50.5}} & 38.7 & 31.9 & 30.6 & 4.8 & 55.2 & 49.7 & 72.9 & 56.3 & 68.3 & 57.6 \\	

CCAFFMNet$_{22}$~\cite{2022CCAFFMNet} 	& RGB-T & \textcolor{blue}{\textbf{95.2}} & 89.1 & 85.9 & 72.5 & \textcolor{blue}{\textbf{82.3}} & \textcolor{blue}{\textbf{67.5}} & 71.8 & 46.3 & 32.5 & 25.2 & 56.8 & \textcolor{blue}{\textbf{17.3}} & 58.3 & 50.6 & 76.6 & 58.3 & 72.9 & \textcolor{blue}{\textbf{58.2}} \\

DSGBINet$_{22}$~\cite{2022DSGBINet}  	& RGB-T & \textcolor{blue}{\textbf{95.2}} & 87.4 & 89.2 & 69.5 & \textcolor{red}{\textbf{85.2}} & 64.7 & 66.0 & 46.3 & 56.7 & \textcolor{red}{\textbf{43.4}} & 7.8 & 3.3 & \textcolor{red}{\textbf{82.0}} & \textcolor{red}{\textbf{61.7}} & 72.8 & 48.9 & 72.6 & 58.1 \\

CMX-SegF-B2$_{22}$~\cite{liu2022cmx} 	& RGB-T & - & \textcolor{blue}{\textbf{89.4}} & - & \textcolor{red}{\textbf{74.8}} & - & 64.7 & - & 47.3 & - & 30.1 & - & 8.1 & - & 52.4 & - & \textcolor{blue}{\textbf{59.4}} & - & \textcolor{blue}{\textbf{58.2}} \\				   						
FDCNet$_{22}$~\cite{2022FDCNet}  	& RGB-T & 94.1 & 87.5 & \textcolor{red}{\textbf{91.4}} & 72.4 & 78.1 & 61.7 & 70.1 & 43.8  & 34.4 & 27.2 & 61.5 & 7.3 & 64.0 & 52.0 & 74.5 & 56.6 & 74.1 & 56.3 \\	

ECGFNet$_{23}$~\cite{zhou2023embedded} & RGB-T & 89.4	& 83.5	& 85.2	& 72.1	& 72.9 &	61.6	& 62.8	& 40.5 &	44.8	& 30.8	& 45.2	& 11.1	& 57.2	& 49.7	& 65.1	& 50.9	& 69.1	& 55.3 \\
LASNet$_{23}$~\cite{2023LASNet}	 & RGB-T & 94.9 & 84.2 & 81.7 & 67.1 & 82.1 & 56.9 & 70.7 & 41.1 & \textcolor{blue}{\textbf{56.8}} & 39.6  & 59.5 &  \textcolor{red}{\textbf{18.9}} & 58.1 &  48.8 &  77.2 & 40.1 & \textcolor{red}{\textbf{75.4}} &  54.9 \\

\hline
\hline

\textbf{CAINet (Ours) } & RGB-T & 93.0	& 88.5	& 74.6	& 66.3	& \textcolor{red}{\textbf{85.2}}	& \textcolor{red}{\textbf{68.7}}	& 65.9	& \textcolor{red}{\textbf{55.4}}	& 34.7	& 31.5	& 65.6	& 9.0	& 55.6	& 48.9	& \textcolor{red}{\textbf{85.0}}	& \textcolor{red}{\textbf{60.7}}	& 73.2	& \textcolor{red}{\textbf{58.6}} \\ 

\toprule[1pt]
\end{tabular}
\end{table*}
\subsection{Residual Learning with Auxiliary and Target Tasks}
\label{sec:RLATT}

%

In order to explicitly convey the correlation between multiple modalities and target regions, we adopt the idea of residual learning to achieve cross-modal feature interaction in a collaborative manner with the CACR and GCM modules. 
Residual learning refers to the use of global context features $G$ to gradually guide the interactions of multi-level complementary features, and is performed step-by-step with multi-level interaction features and global information.
As shown at the bottom of Fig.~\ref{fig:Framework}, the ARLM has three inputs,~\eg the global context features $G$, and the complementary features $CR_5$ and $CR_4$ for ARLM$_1$, after which the predicted segmentation map $P_1$ and the predicted attention map $Att_1$ are obtained.  
%
After the same step-by-step calculation as above, predicted results can be obtained from residual learning, which are $P_2$, $Att_2$, $P_3$, the predicted binary map $binary$, $P_4$, and the predicted boundary map $boundary$.

In particular, in auxiliary supervision, the purpose of using attention maps for supervision is to explicitly reason about complementary relationship between multimodal features. 
The ground truth of attention map $Q$ is obtained by using corrosion expansion and Gaussian fuzzy operation on the ground truth of binary map\footnote{The ground truth of binary map is obtained by setting the non-zero elements of the label values in the semantic segmentation map to 1.}.
And we take the sum of mean square error and the modified linear correlation coefficient metric~\cite{zhou2019global} as the loss function: 
\begin{equation}
    L_{Att_i}=\frac{1}{N} \sum_{i=1}^n\left\|Att_i-Q\right\|^2-\frac{\operatorname{cov}(Att_i, \mathrm{Q})}{\sqrt{\operatorname{var}(Att_i)} \sqrt{\operatorname{var}(Q)}},
\label{eq:13}
\end{equation}
where $cov$ and $var$ represent covariance and variance, respectively.

For $P_1$ and $P_2$, we utilize Lov{\'a}sz-Softmax~\cite{berman2018lovasz} to respectively calculate their losses as $L_{seg1}$ and $L_{seg2}$, 
which has the perceptual quality and proportional invariance, and gives small objects proper correlation and false negative counts compared to the cross-entropy loss. The two semantic segmentation losses are  $L_{seg1}$ and $L_{seg2}$ as follows:
\begin{equation}
   \begin{aligned}
   L_{seg1}=\frac{1}{|C|} \sum_{c \in C} \overline{\Delta_{J_c}}(m(c)), \\
     m(c)= \begin{cases}1-P_1(c) & \text { if } c=Y(c) \\ P_1(c) & \text { otherwise },
    \end{cases}
    \\
    \end{aligned}
\label{eq:14}
\end{equation}
where $Y$ denotes the semantic segmentation label, $|C|$ represents the class number, $\Delta Jc$ defines the Lov{\'a}sz extension of the Jaccard index. $L_{seg2}$ is similar to $L_{seg1}$.
For $P_3$ and $P_4$, we use the weighted cross-entropy function to calculate their losses, denoted as $L_{seg3}$ and $L_{seg4}$ with the following function:
\begin{equation}
    \begin{aligned}
    L_{seg3}=-\frac{1}{N} \sum_{i=1}^N \omega\left(Y^i \log (P_3^i)\right),
    \end{aligned}
\label{eq:16}
\end{equation}
$L_{seg4}$ is similar to $L_{seg3}$.
For these four semantic segmentation losses, we add them together to obtain the target loss $L_{target}$.
For $binary$ and $boundary$\footnote{The ground truth of boundary map is obtained by performing Canny edge detection on the ground truth of binary map.}, we use the weighted binary cross-entropy loss function to calculate the binary loss and boundary loss, donated as $L_{binary}$ and $L_{boundary}$, respectively.
\begin{small}
\begin{equation}
    L_{\text {binary }}=-\frac{1}{N} \sum_{i=1}^N \omega(Y_{bi}^i \log (binary^i)+(1-Y_{bi}^i) \log (1-binary^i)),
\label{eq:15}
\end{equation}
\end{small}where $N$ denotes the total number of input pixels, $\omega$ denotes the class weights computed following~\cite{enet}, and $Y_{bi}$ denotes the ground truth of binary map. $L_{boundary}$ is similar to $L_{binary}$.
%
Furthermore, for the coarse segmentation maps $S_{rgb}$, $S_{thermal}$, and $S_{global}$ produced by the three decoders of the backbone network, we adopt the ordinary cross-entropy loss, and define the sum of these three losses as $L_{decoder}$.
%
To sum up, the final loss for auxiliary and target supervision is calculated
 as follows:
 \begin{equation}
    \begin{aligned}
    L_{total}=L_{target}+L_{Att1}+L_{Att2}+L_{binary}\\+L_{boundary}+L_{decoder}.
    \end{aligned}
\label{eq:10}
\end{equation}

%
%
\section{Experiments}
\label{sec:exp}

\subsection{Experimental Protocol}
\label{sec:ExpProtocol}
\textit{1) Datasets:}
We train and evaluate the proposed CAINet on two RGB-T semantic segmentation datasets: MFNet~\cite{2017MFNet} and PST900~\cite{2020PSTNet}. The MFNet dataset contains 1569 RGB-T image pairs  (820 taken during the day and 749 at night) and has 9 categories, including background, car, person, bike, curve, car stop, guardrail, color cone, and bump. 
The resolution size of the image pairs is $480\times 640$ pixels. We divide the dataset into three parts: training set, validation set, and test set, using the same splitting scheme as proposed in MFNet. 
The PST900 dataset contains 894 aligned RGB and thermal image pairs taken by the Stereolabs ZED Mini stereo camera and the FLIR Boson 320 camera, respectively. 
This dataset contains five categories: background, fire extinguisher, backpack,  hand drill, and survivor. The resolution of the RGB and thermal images is $1280\times 720$. 
This dataset is divided into two parts, the training set and the test set, with 597 pairs constituting the training and 288 pairs constituting the test set.

\textit{2) Evaluation Metrics:}
We used two metrics to quantitatively evaluate the performance of semantic segmentation, which are the mean accuracy  (mAcc) and the mean intersection over union (mIoU). The mean accuracy is calculated as the ratio of the number of correct pixels for each class to the number of all predicted pixels for that class, and then the average value is calculated by summation. The mean intersection over union is the average result of summing the intersection of the true values of the predictions for each class and the ratio of the sums. These measures are defined as follows: 
\begin{equation}
    mAcc =\frac{1}{k+1} \sum_{i=0}^k \frac{p_{i i}}{\sum_{j=0}^k p_{i j}},
    \label{eq:19}
\end{equation}
\begin{equation}
     mIoU =\frac{1}{k+1} \sum_{i=0}^k \frac{p_{i i}}{\sum_{j=0}^k p_{i j}+\sum_{j=0}^k p_{j i}-p_{i i}},
    \label{eq:20}
\end{equation}
where $k$ is the number of classes (including unlabeled) and class $id=0$ indicates “unlabeled”, $P_{ij}$ is the number of pixels belonging to class $i$ and is predicted as class $j$.

\renewcommand{\addFig}[1]{{\includegraphics[height=0.066\textwidth]{com_sota/#1}}}
\begin{figure*}[t]
  \centering
  \small
  \renewcommand{\arraystretch}{0.5}
  \setlength{\tabcolsep}{0.3mm}
  \begin{tabular}{ccccccccccc}
   \addFig{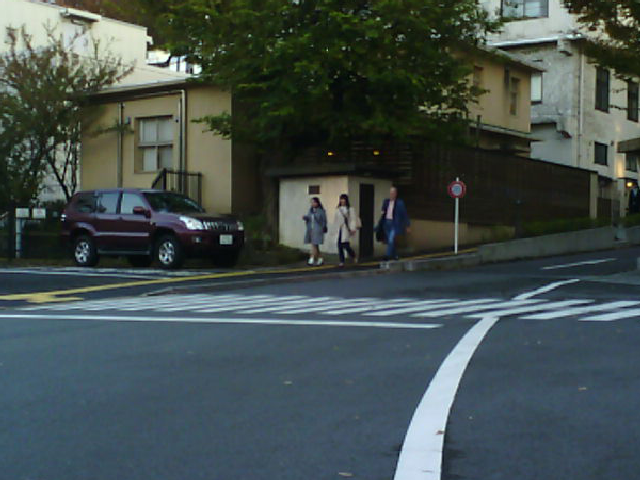}&
    \addFig{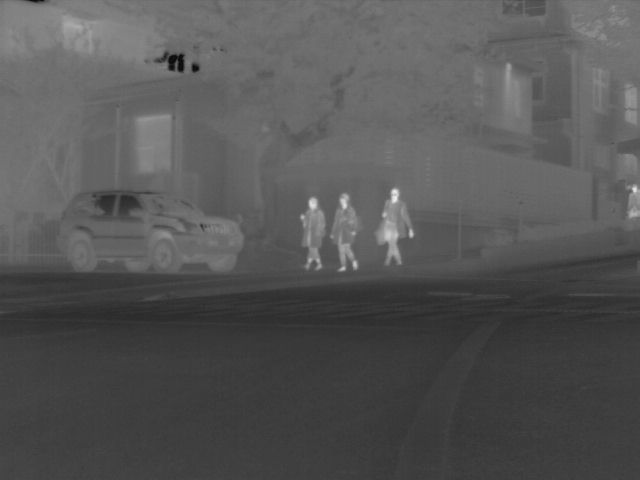}&
    \addFig{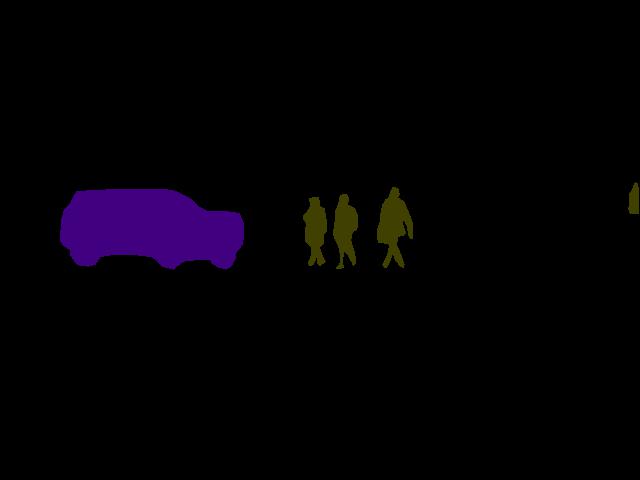}&
   \addFig{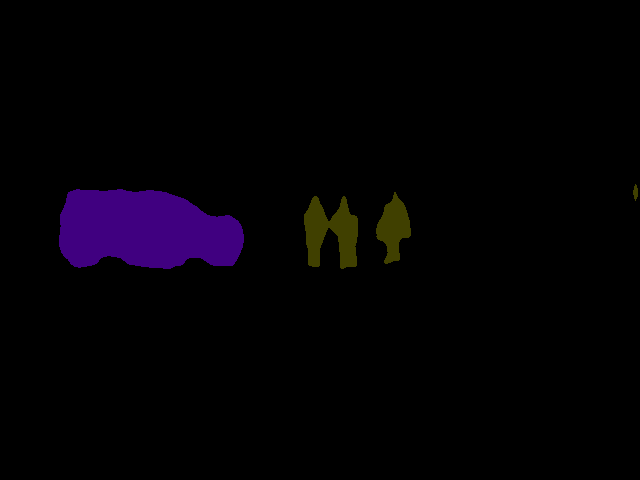}&
    \addFig{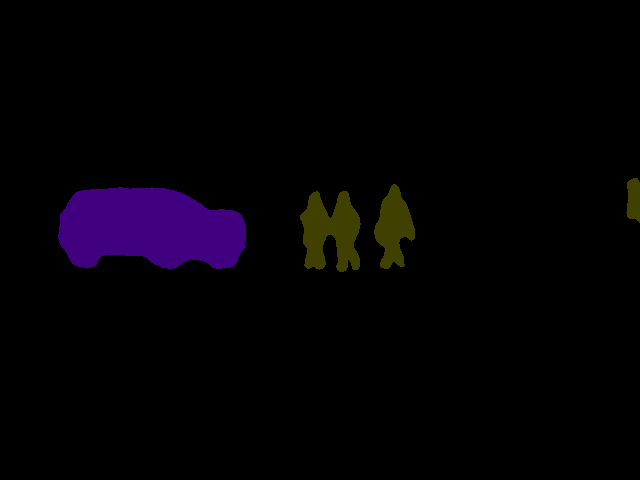}&
   \addFig{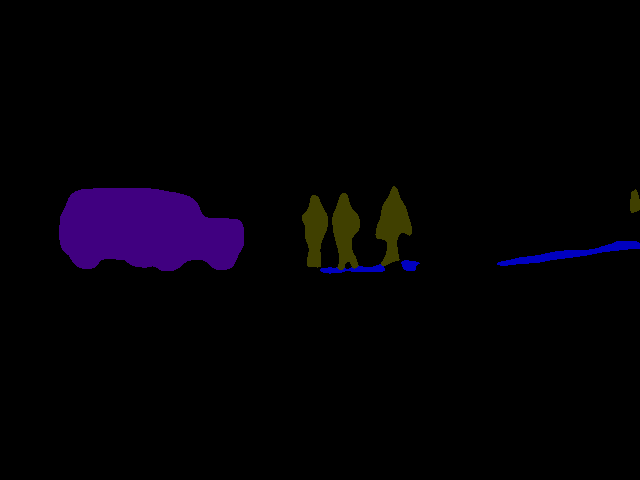}&
   \addFig{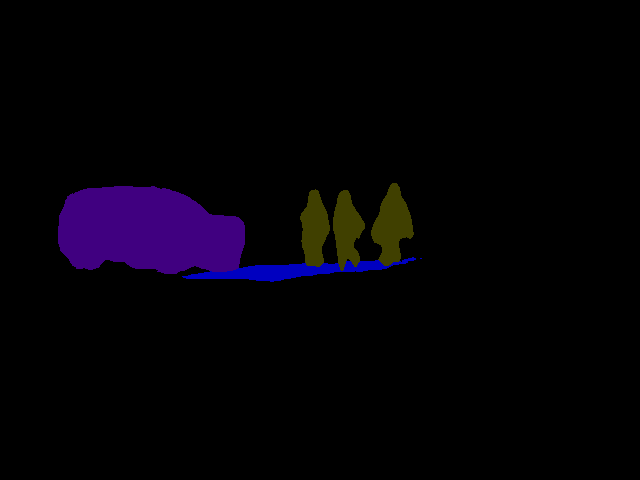}&
   \addFig{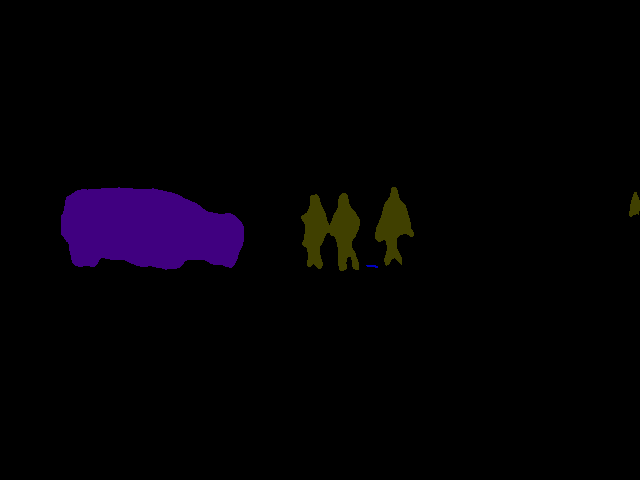}&
   \addFig{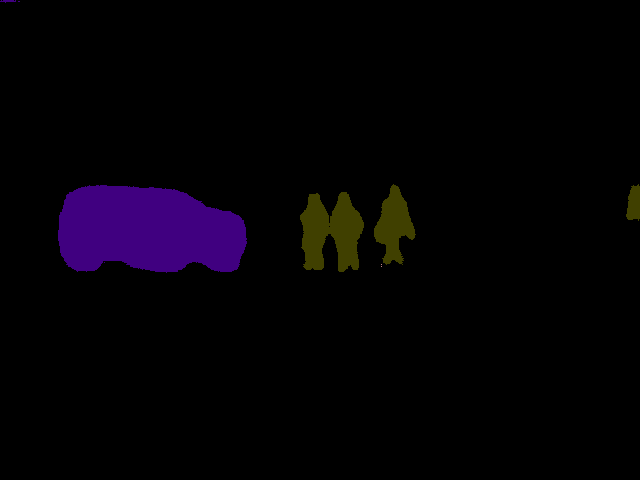}&
   \addFig{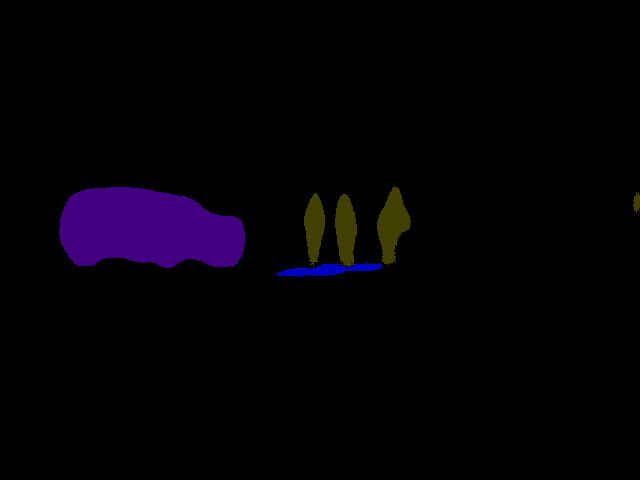}&
   \addFig{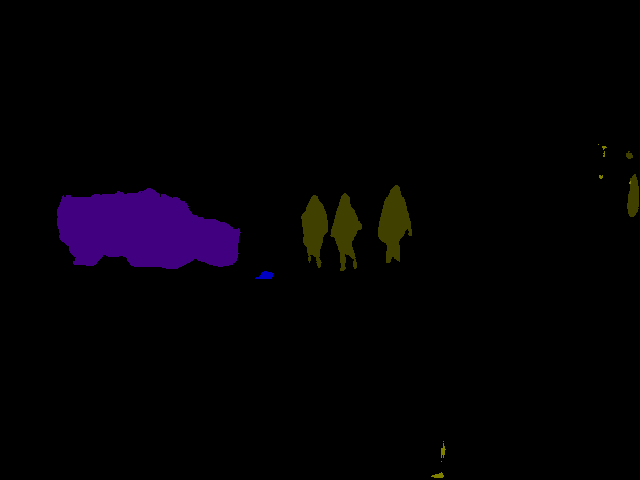}
   \\
    \addFig{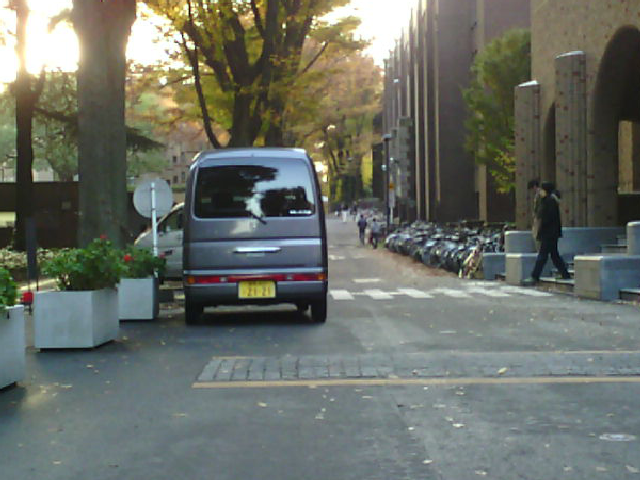}&
    \addFig{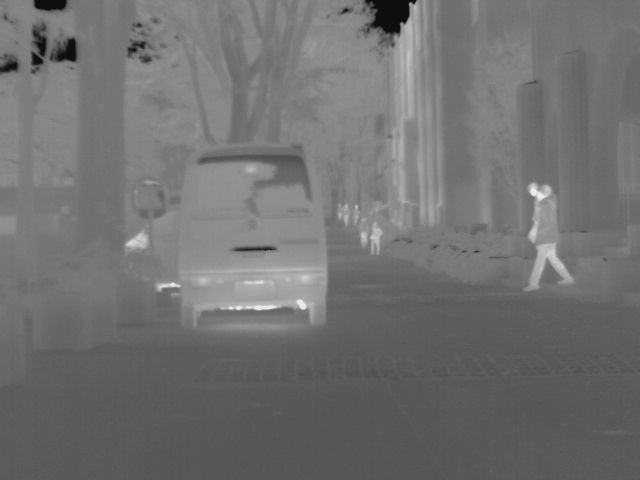}&
    \addFig{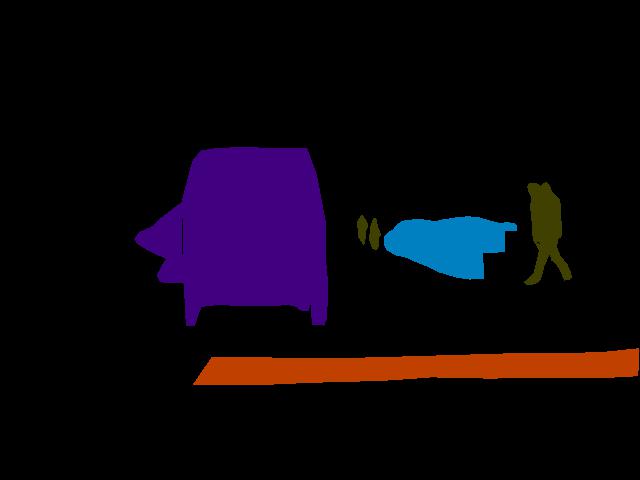}&
   \addFig{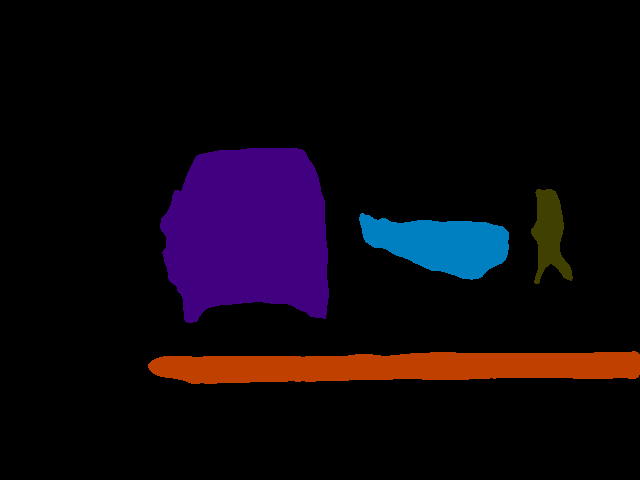}&
    \addFig{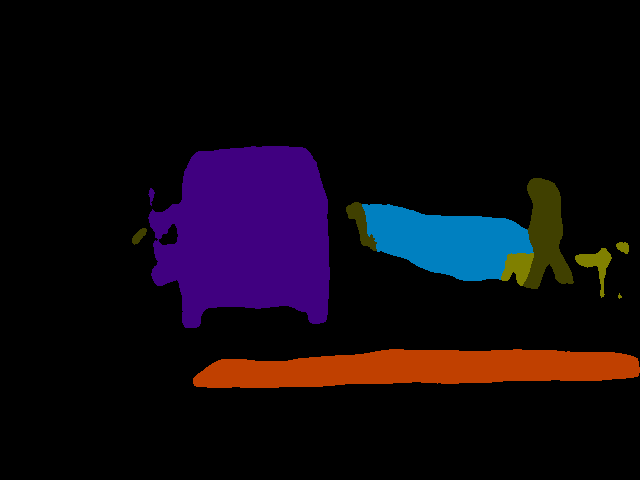}&
   \addFig{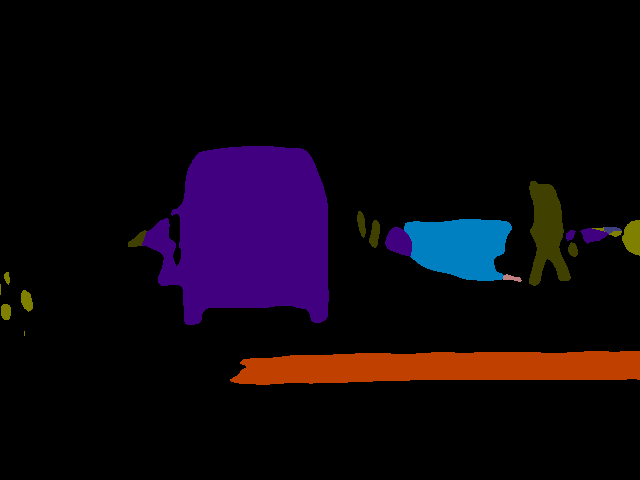}&
   \addFig{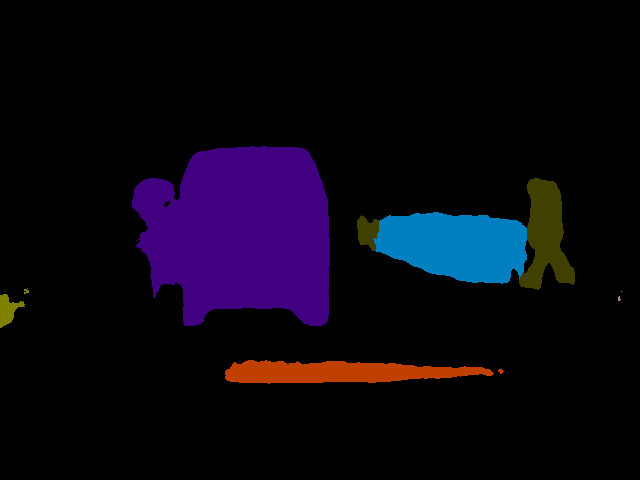}&
   \addFig{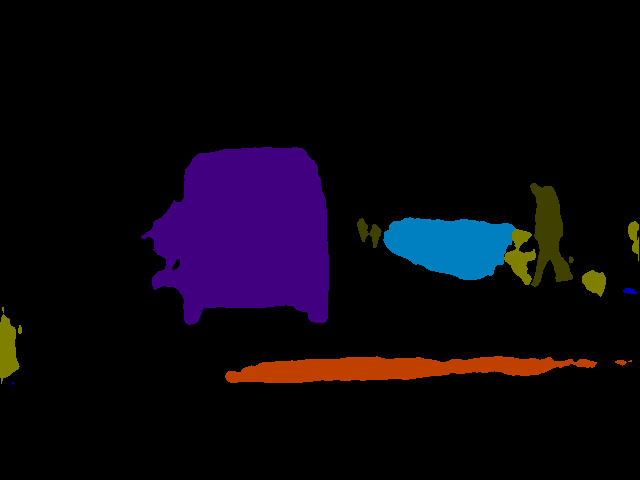}&
   \addFig{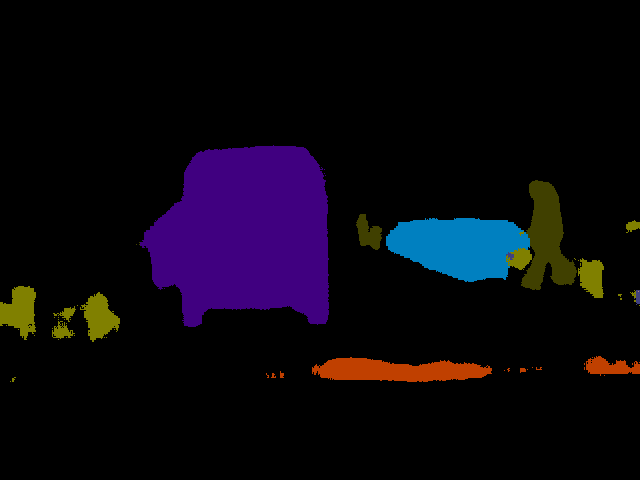}&
   \addFig{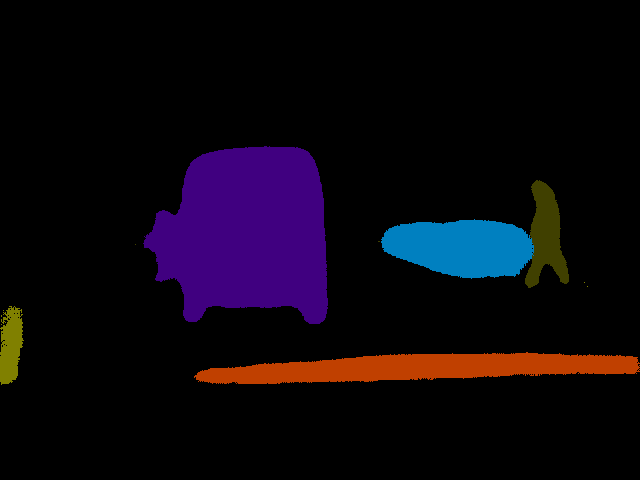}&
   \addFig{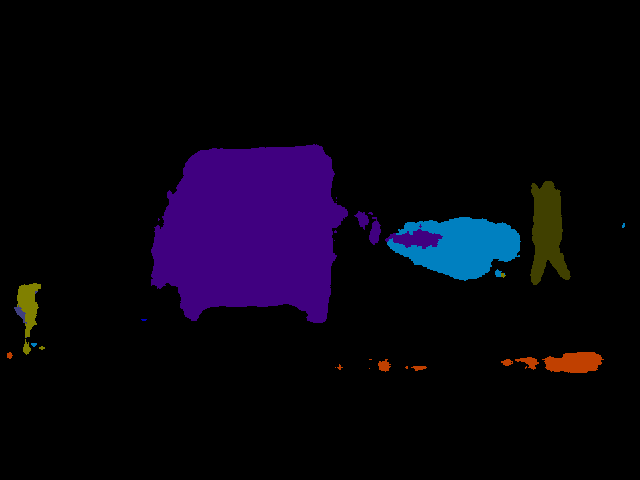}
   \\
    \addFig{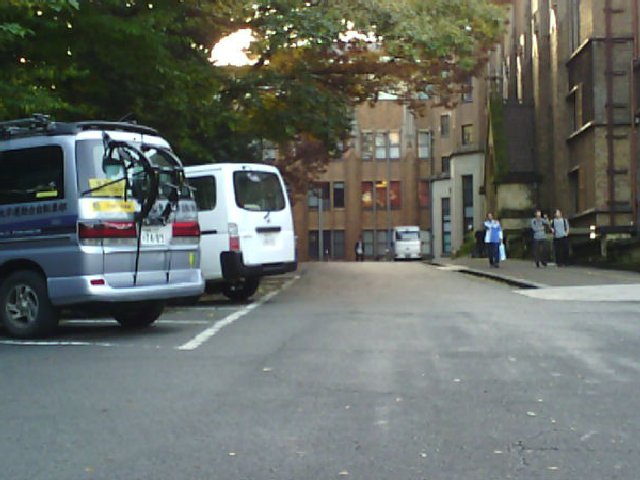}&
    \addFig{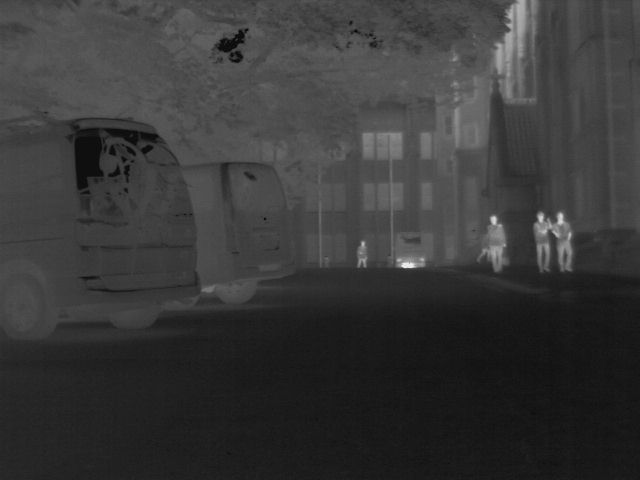}&
    \addFig{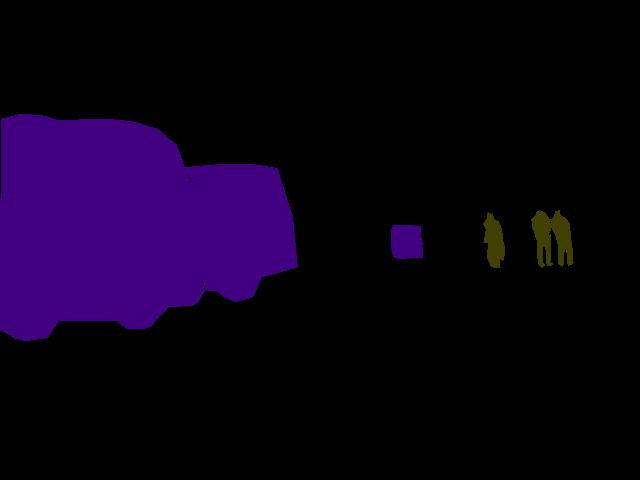}&
   \addFig{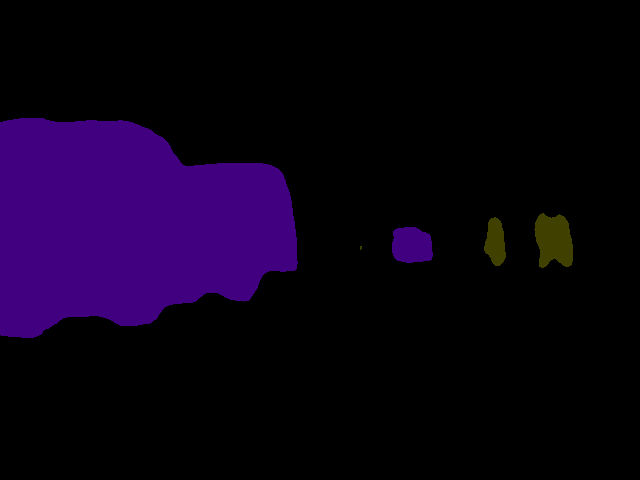}&
    \addFig{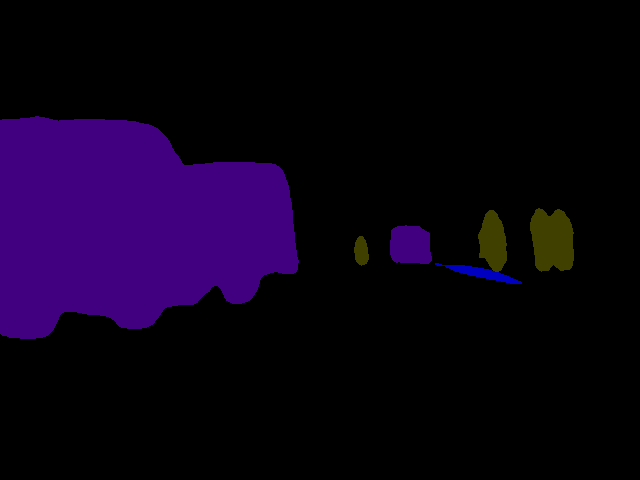}&
   \addFig{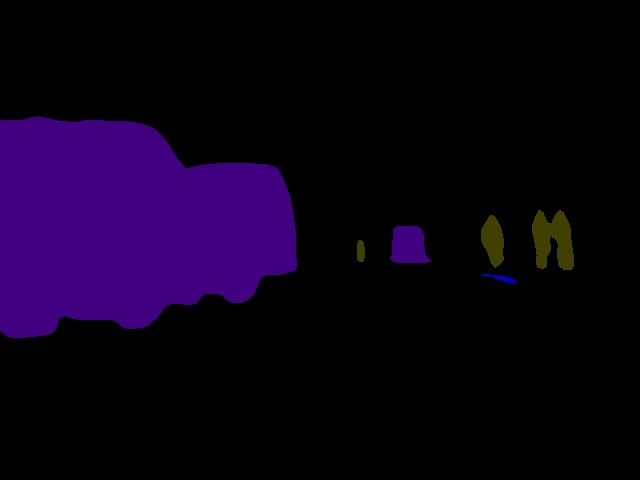}&
   \addFig{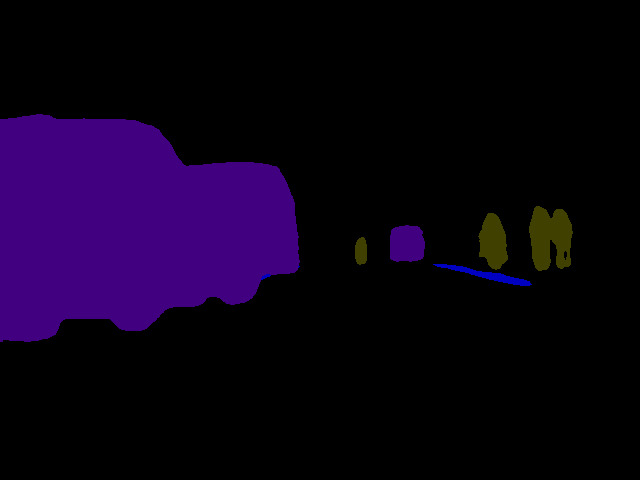}&
   \addFig{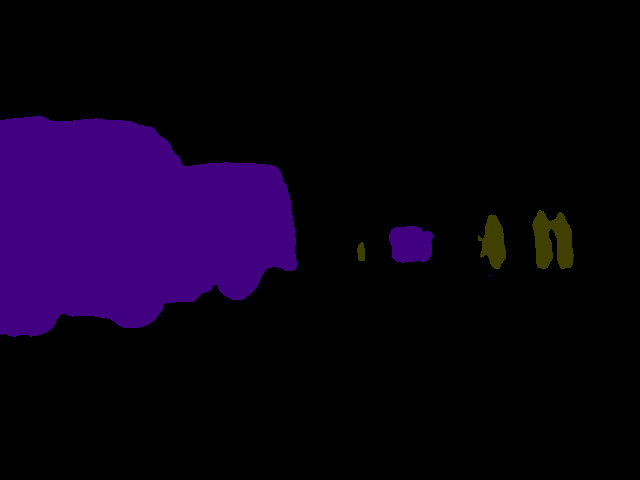}&
   \addFig{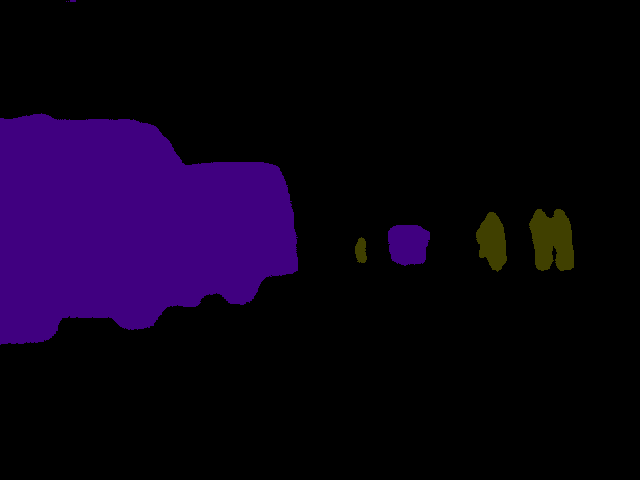}&
   \addFig{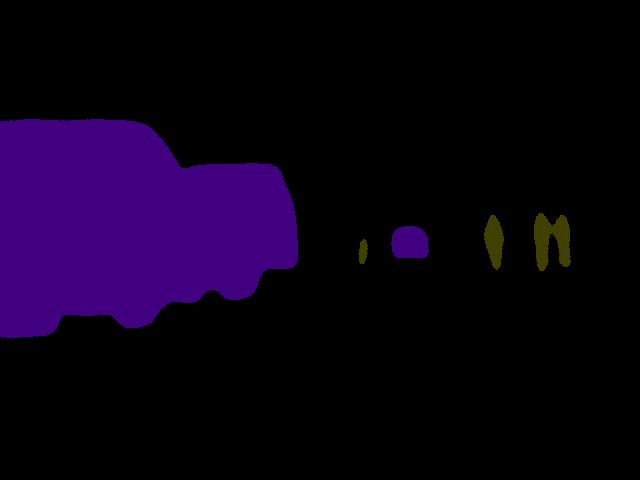}&
   \addFig{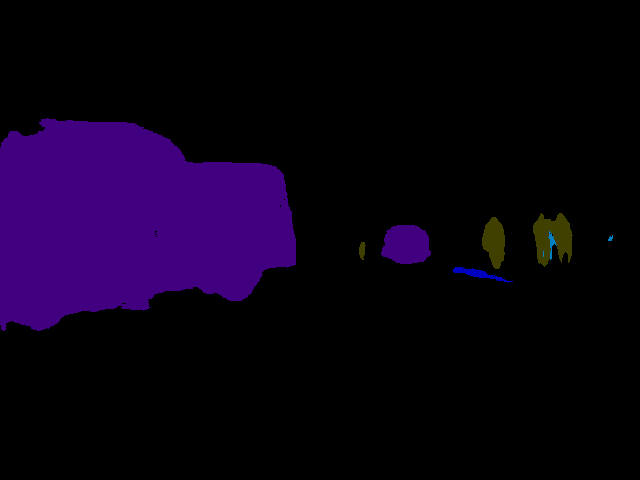}
   \\
    \addFig{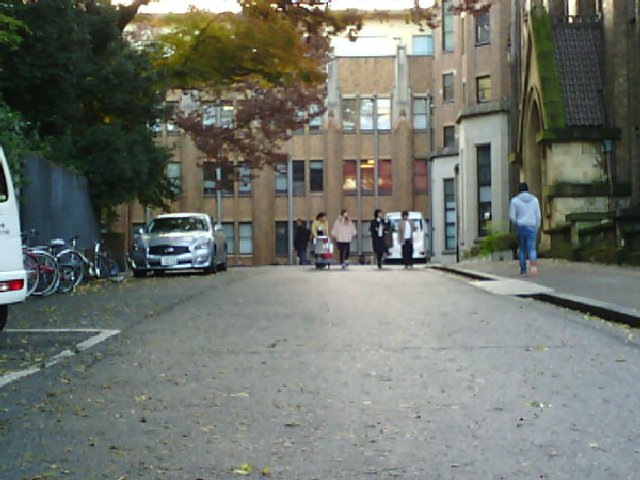}&
    \addFig{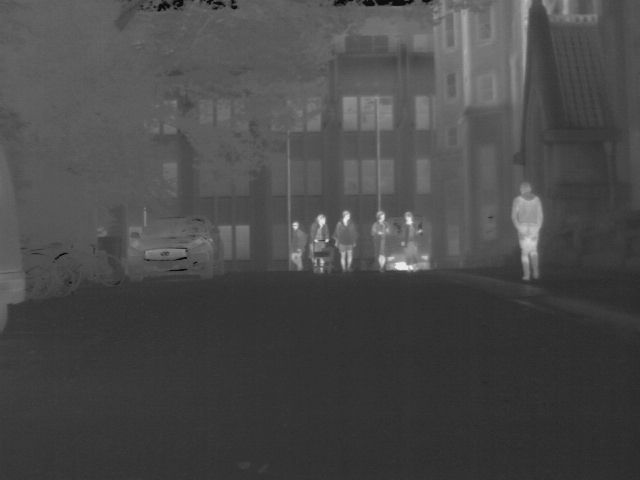}&
    \addFig{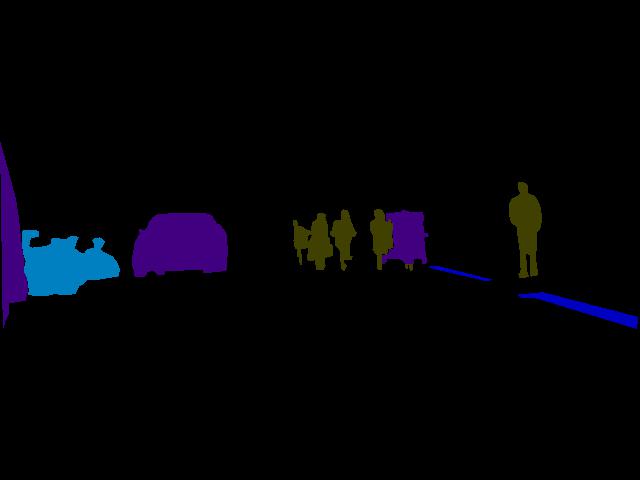}&
   \addFig{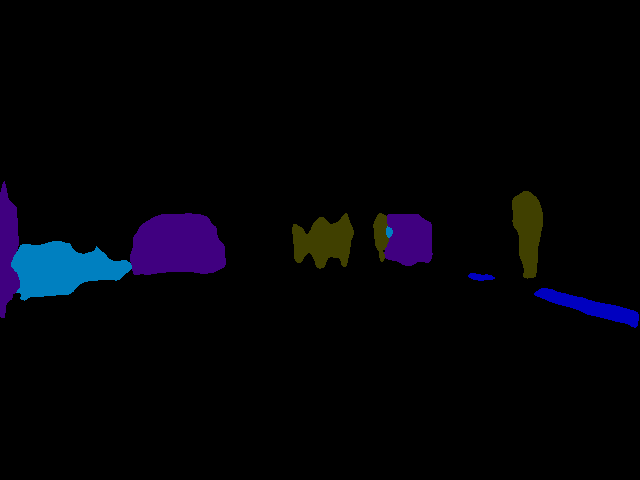}&
    \addFig{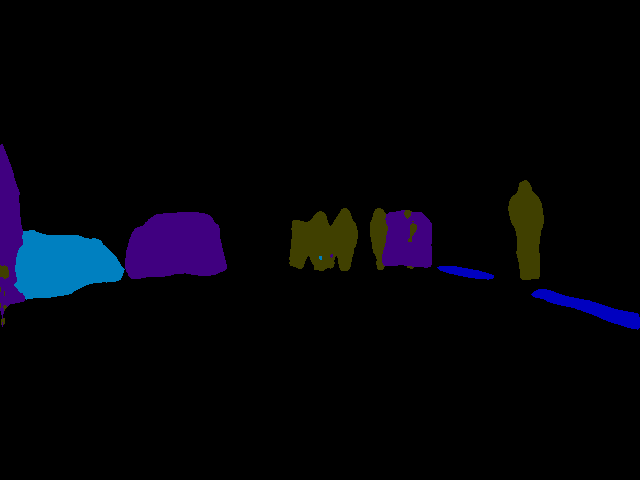}&
   \addFig{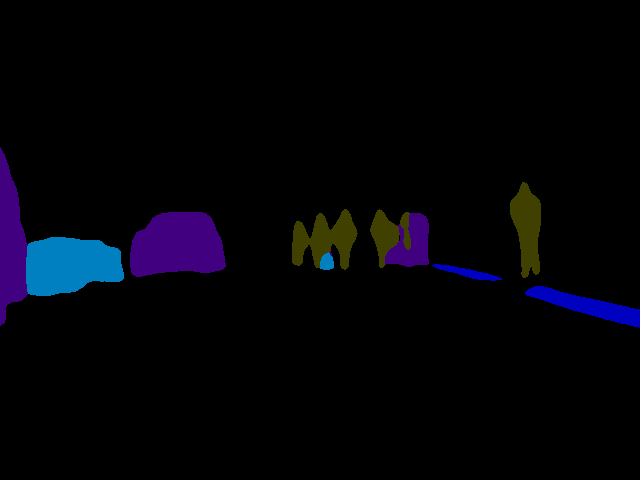}&
   \addFig{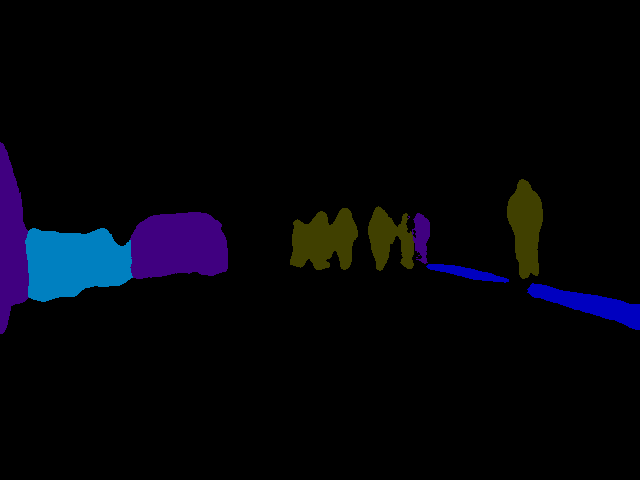}&
   \addFig{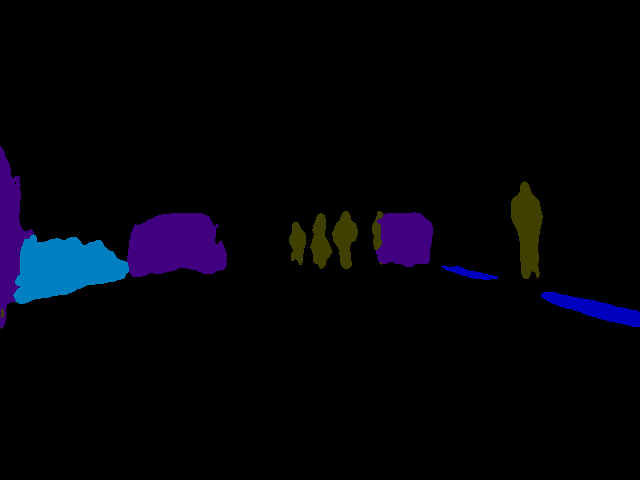}&
   \addFig{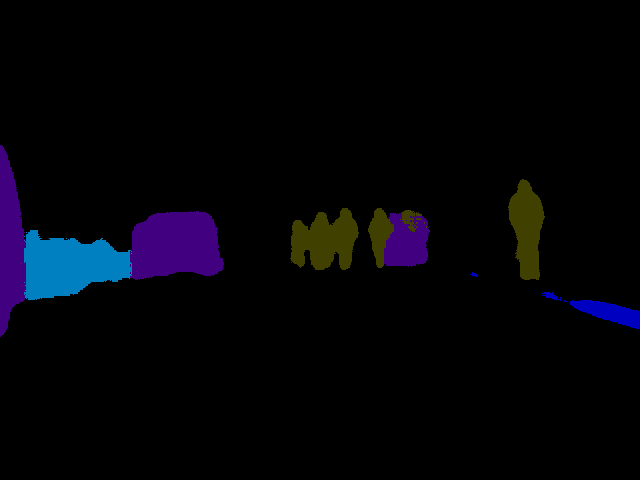}&
   \addFig{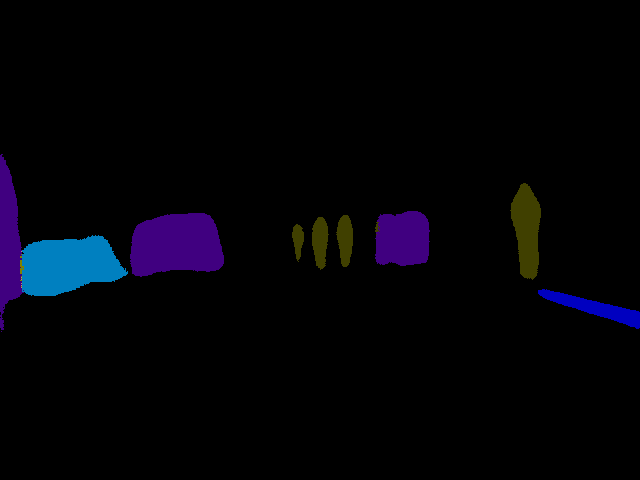}&
   \addFig{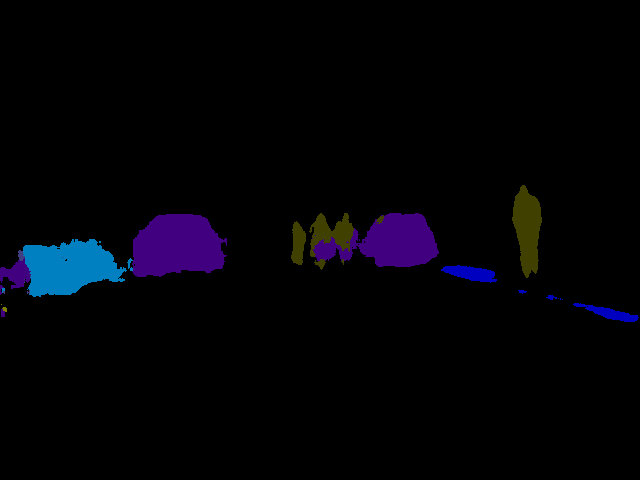}
   \\
    \addFig{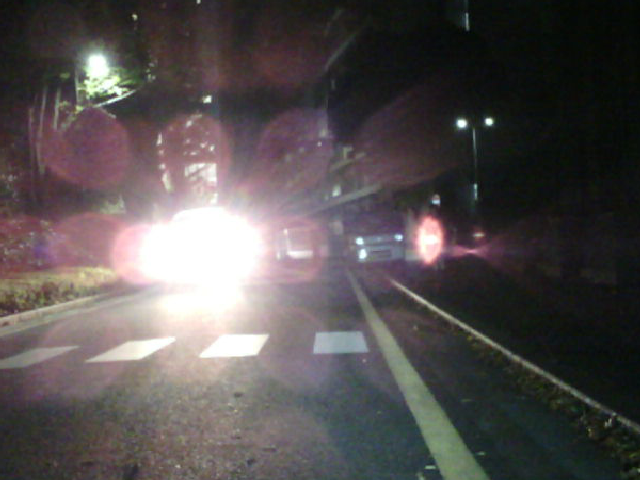}&
    \addFig{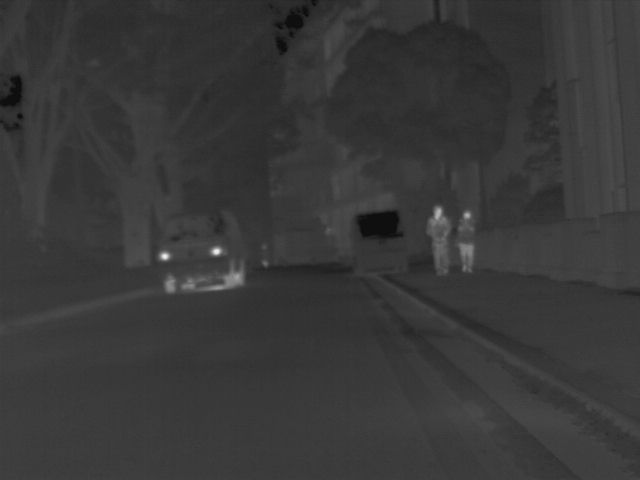}&
    \addFig{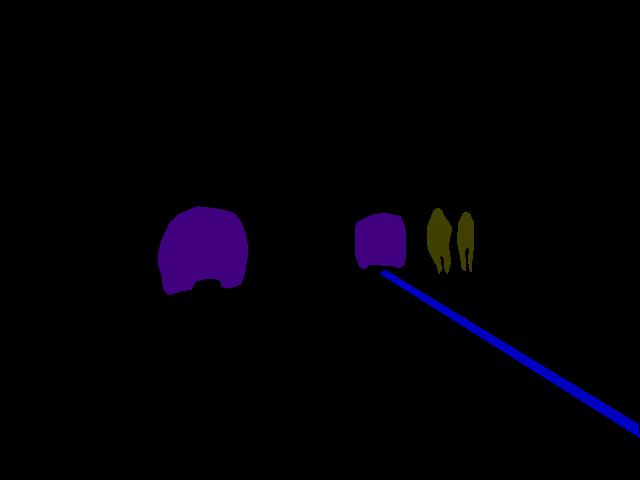}&
   \addFig{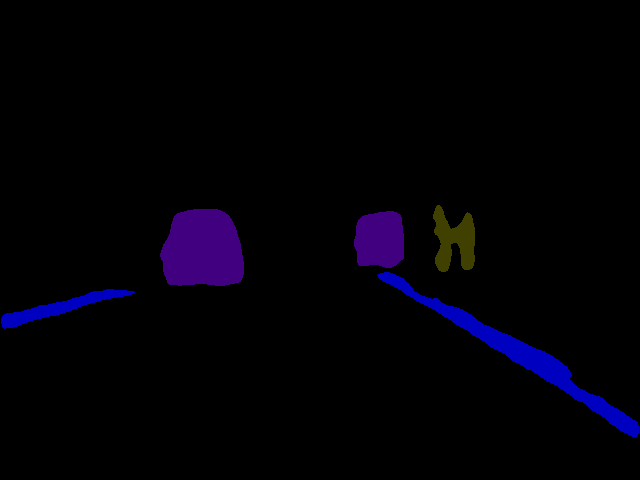}&
    \addFig{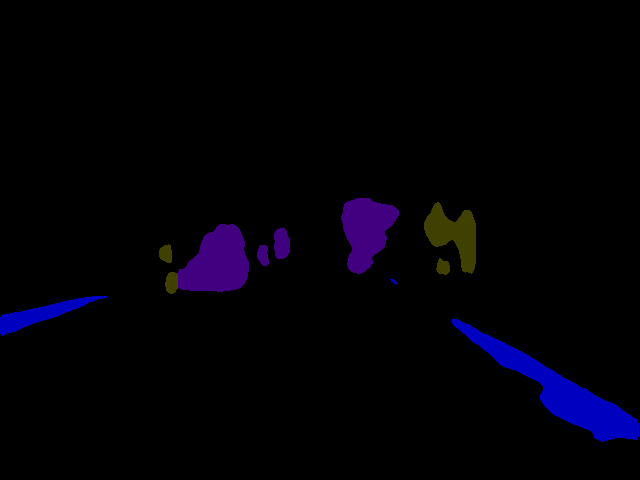}&
   \addFig{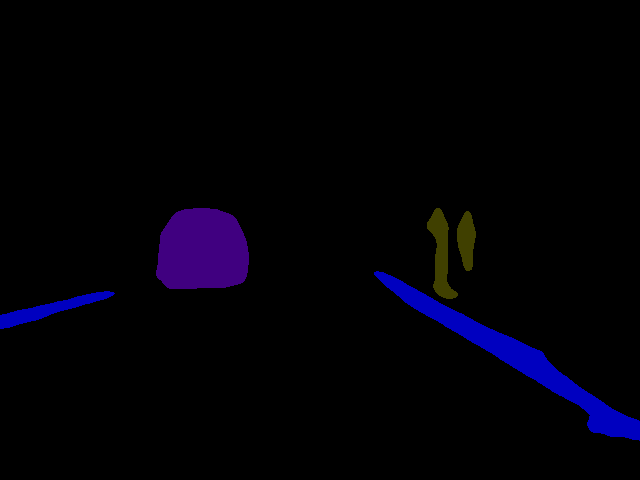}&
   \addFig{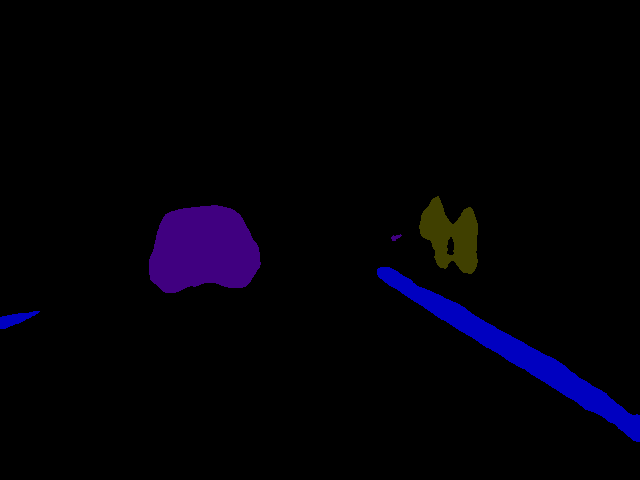}&
   \addFig{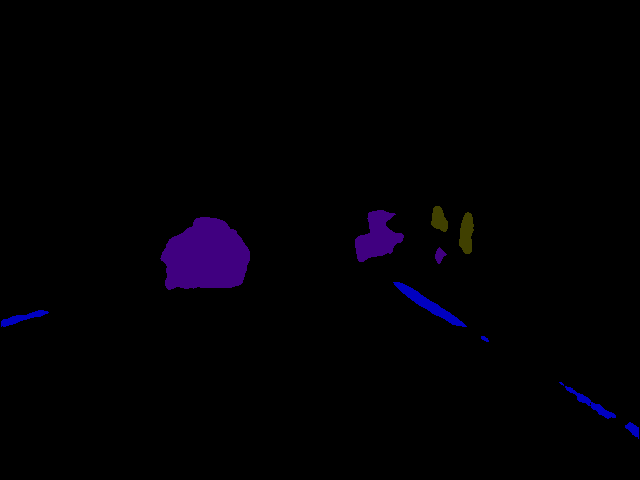}&
   \addFig{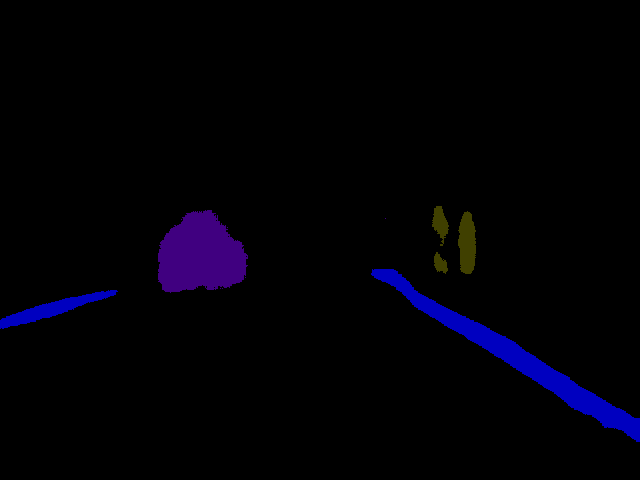}&
   \addFig{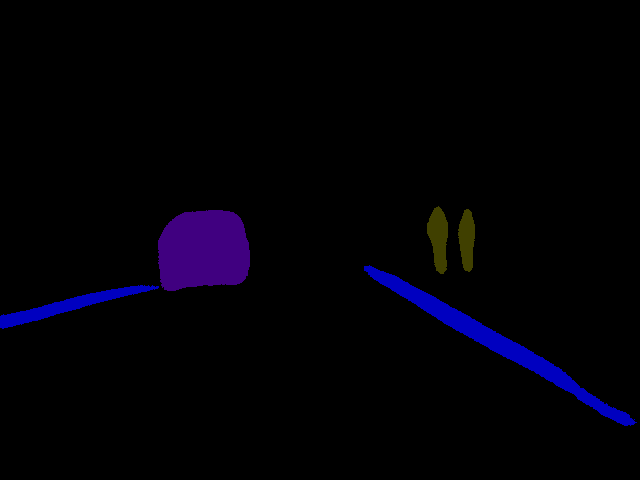}&
   \addFig{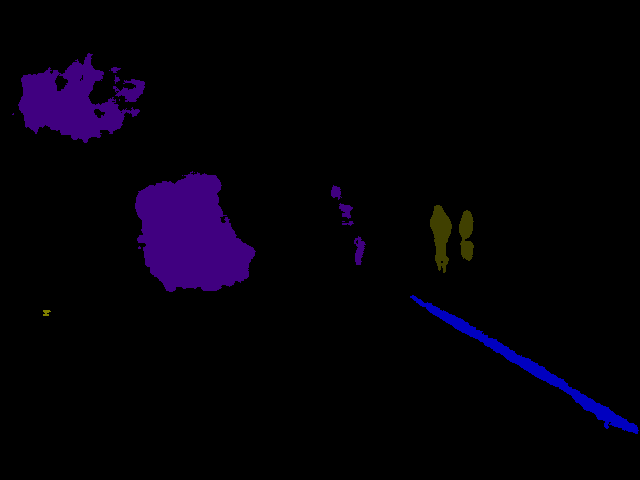}
   \\
    \addFig{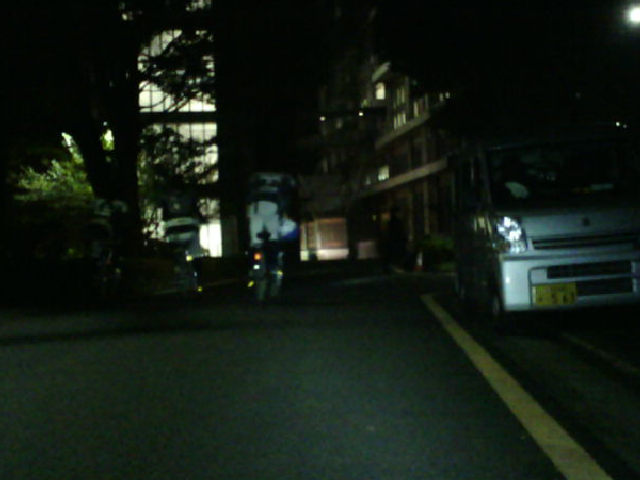}&
    \addFig{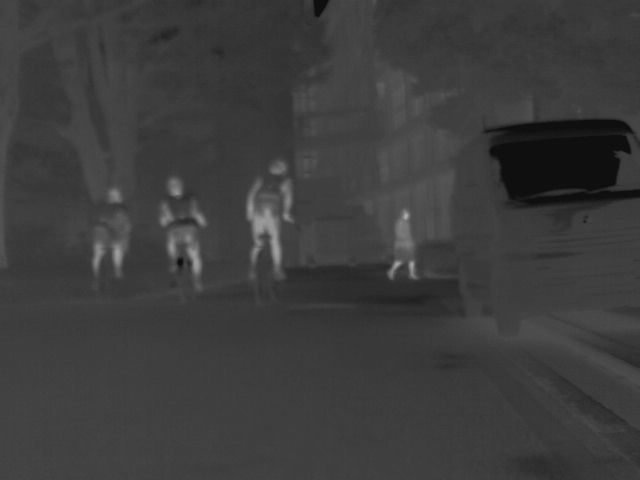}&
    \addFig{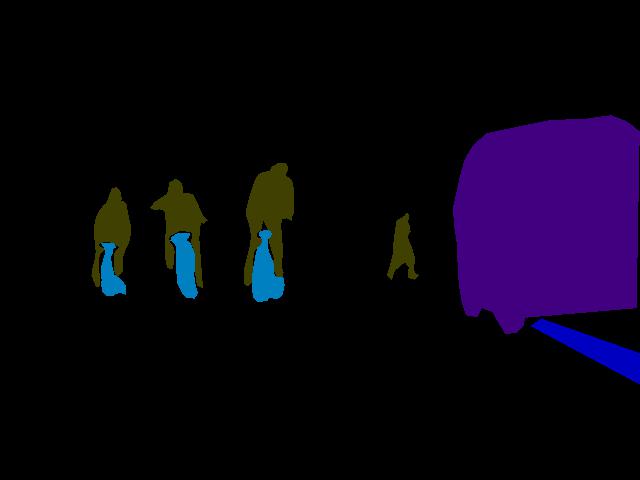}&
   \addFig{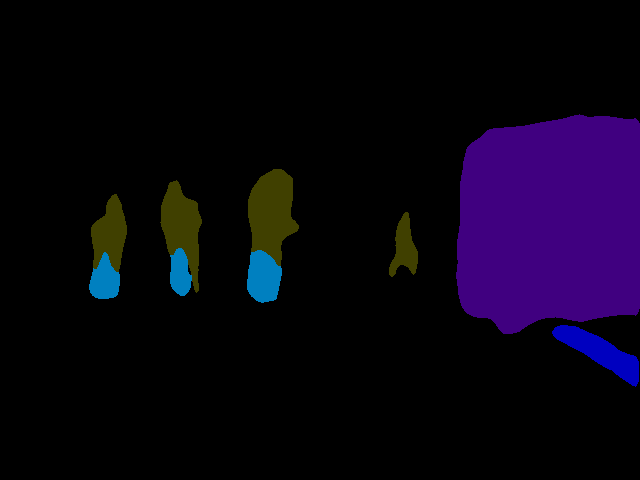}&
    \addFig{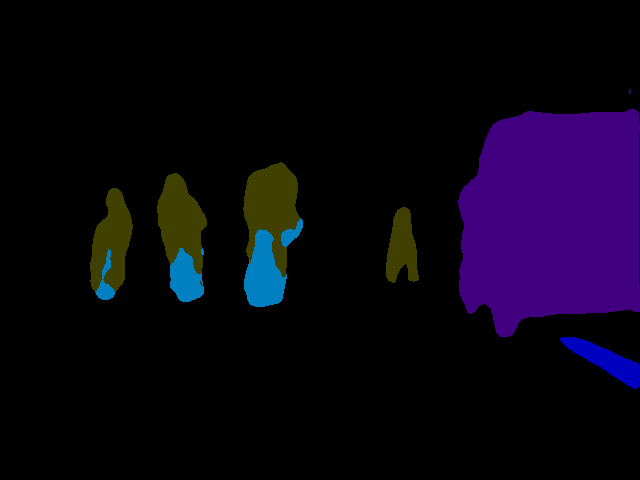}&
   \addFig{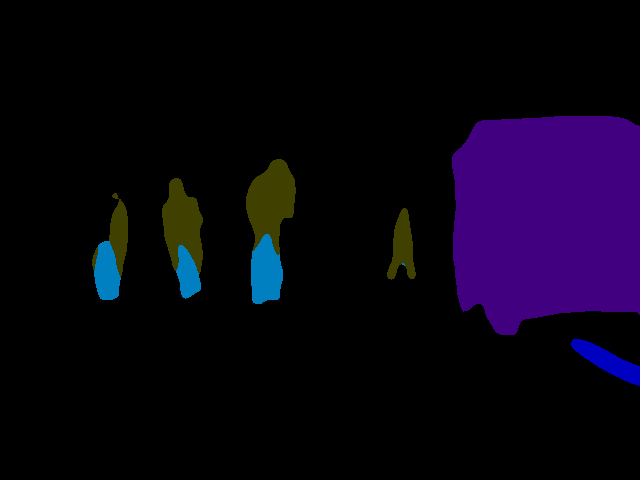}&
   \addFig{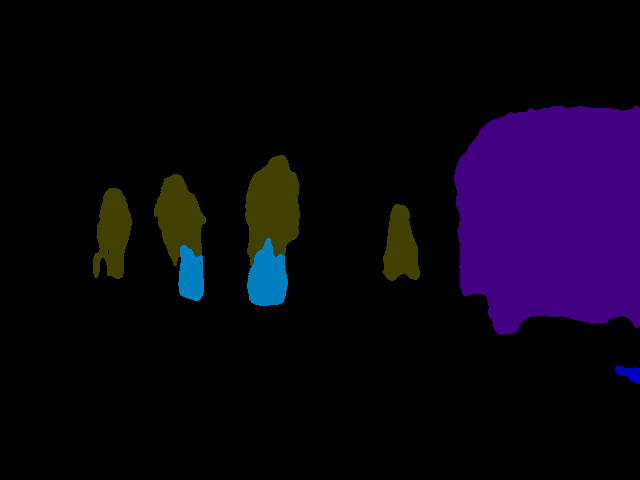}&
   \addFig{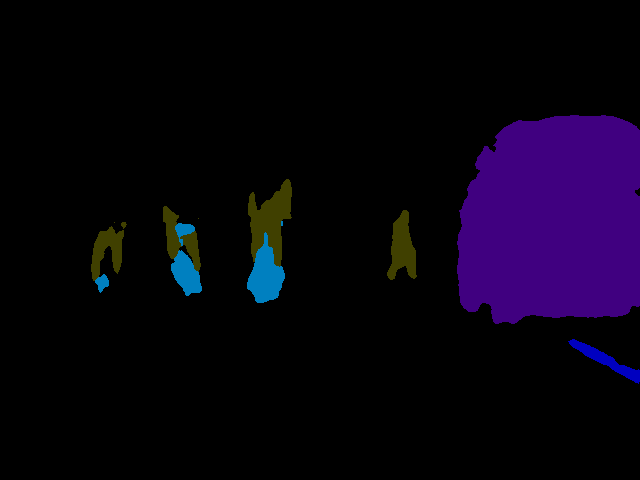}&
   \addFig{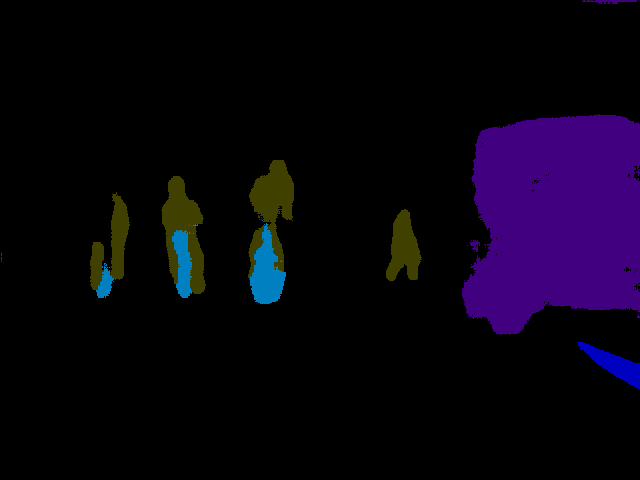}&
   \addFig{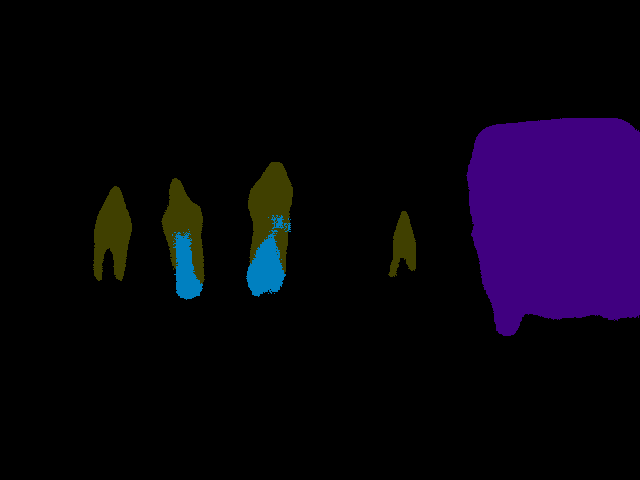}&
   \addFig{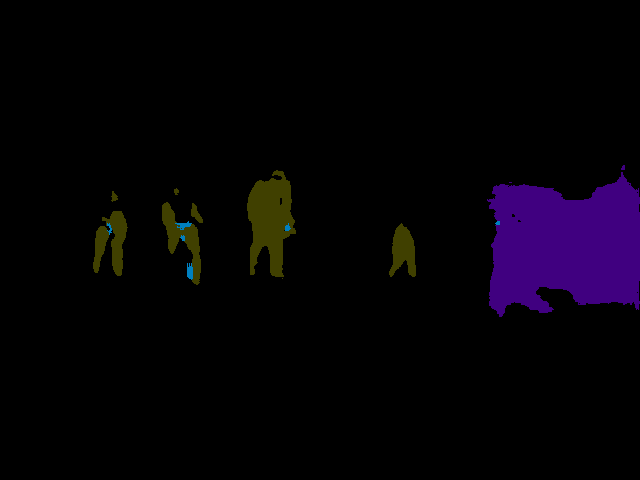}
   \\
    \addFig{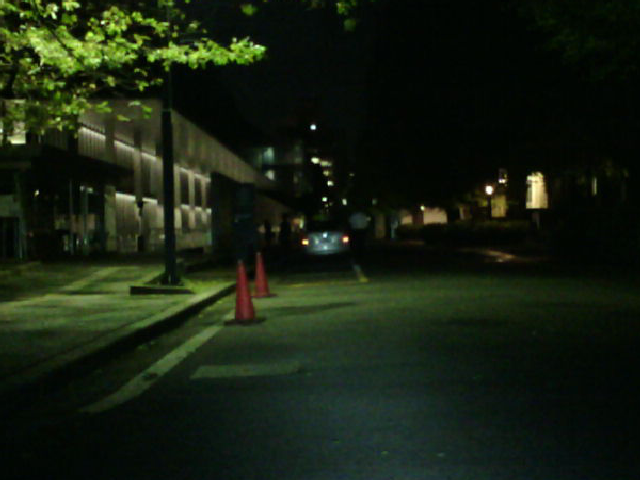}&
    \addFig{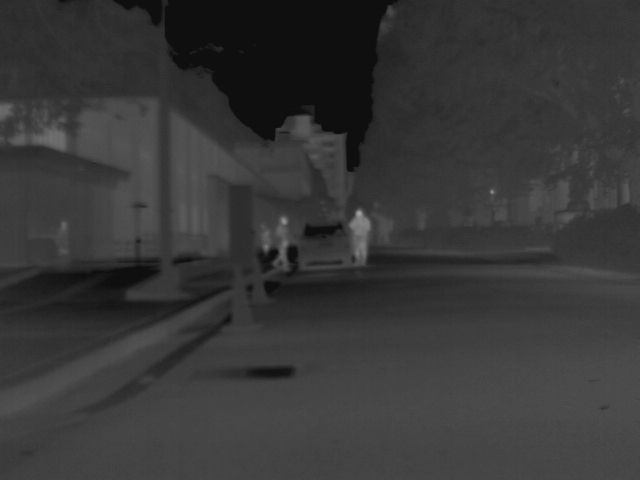}&
    \addFig{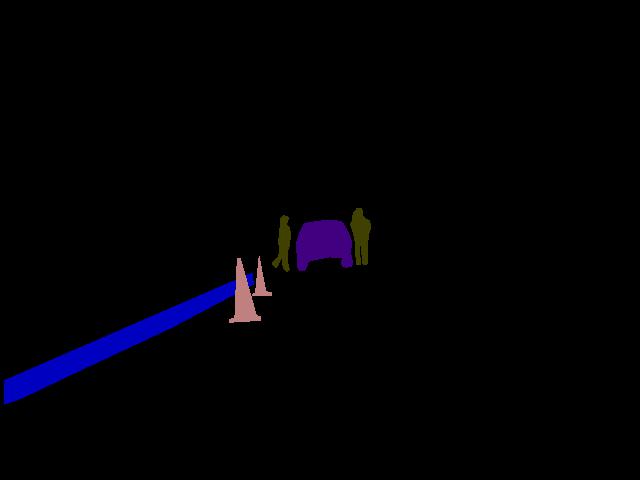}&
   \addFig{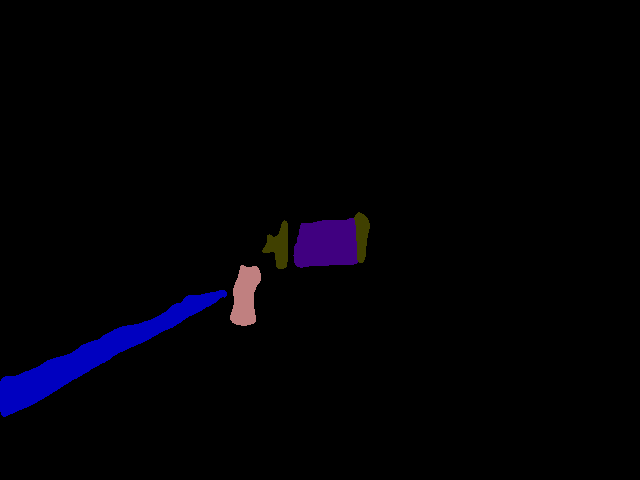}&
    \addFig{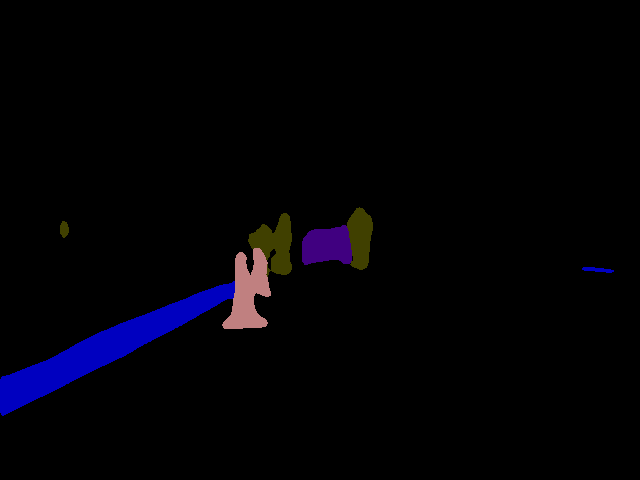}&
   \addFig{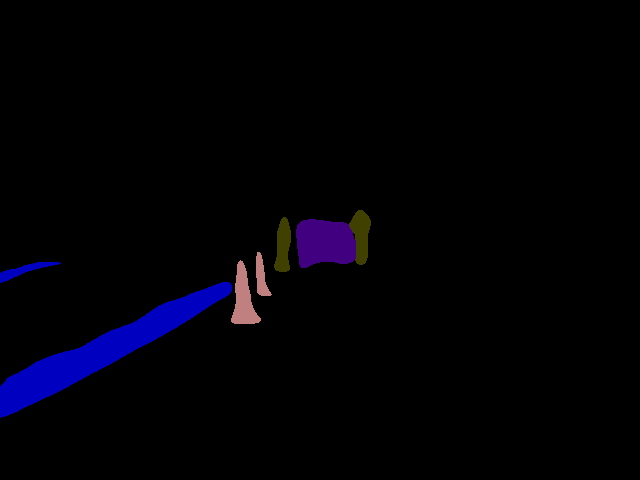}&
   \addFig{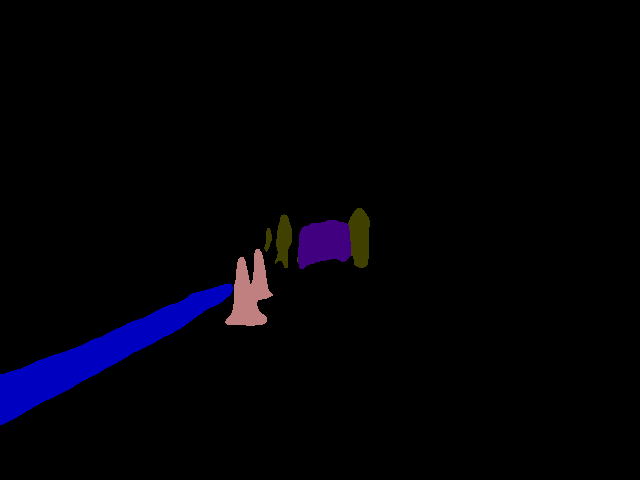}&
   \addFig{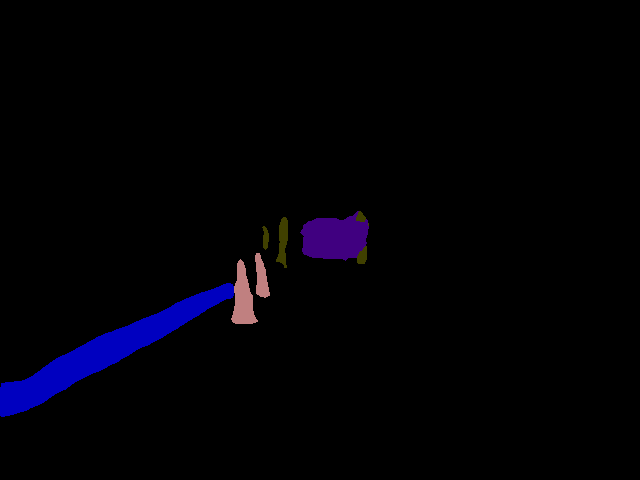}&
   \addFig{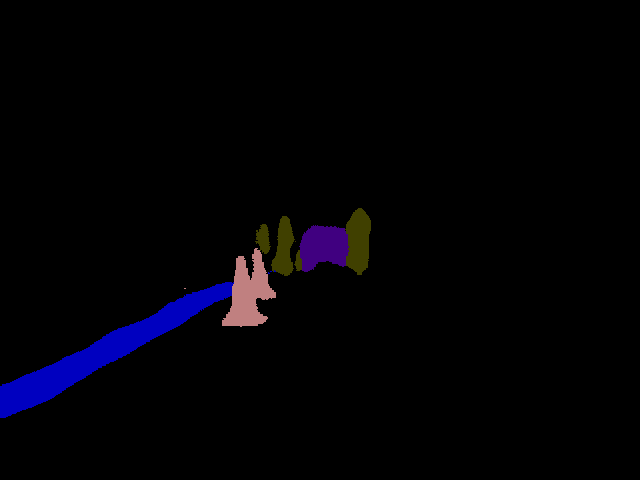}&
   \addFig{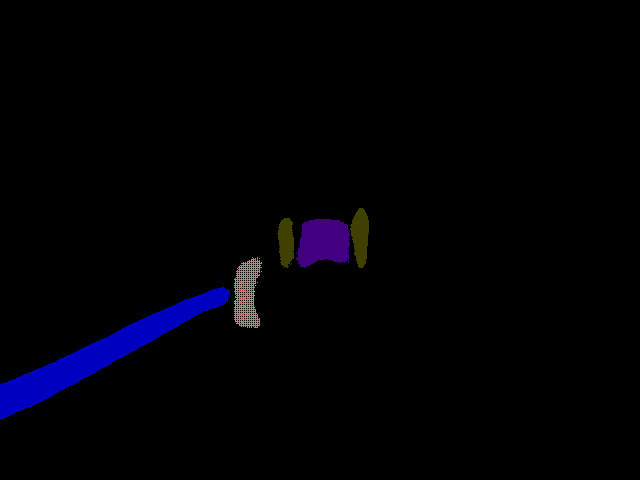}&
   \addFig{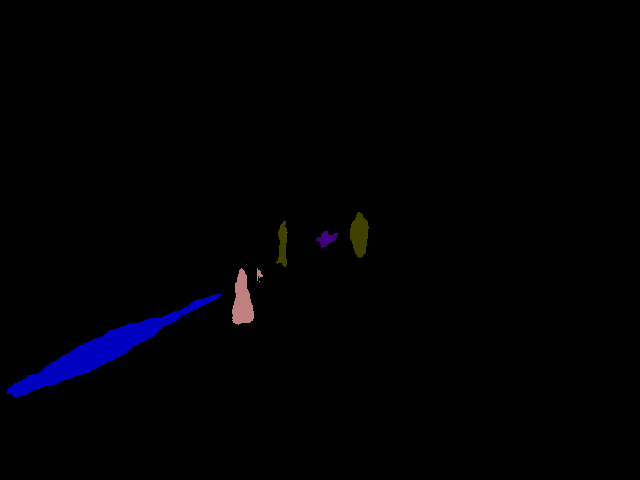}
   \\
    \addFig{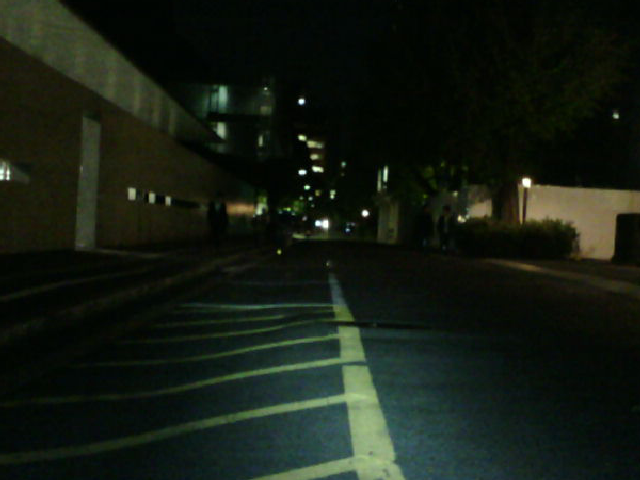}&
    \addFig{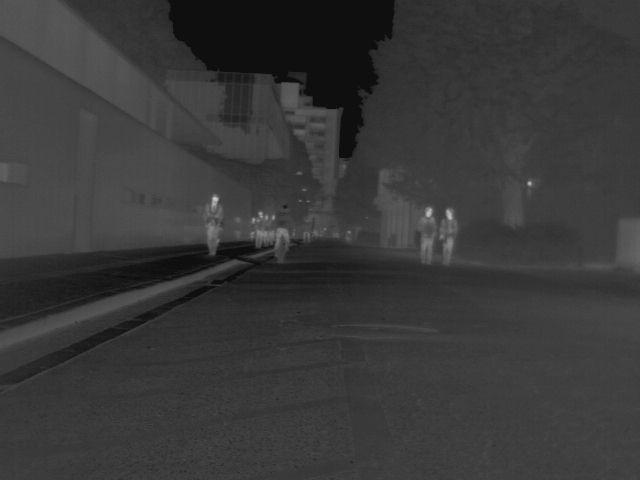}&
    \addFig{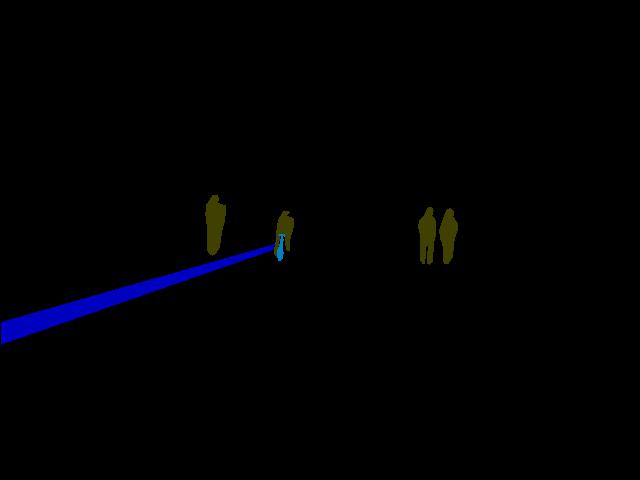}&
   \addFig{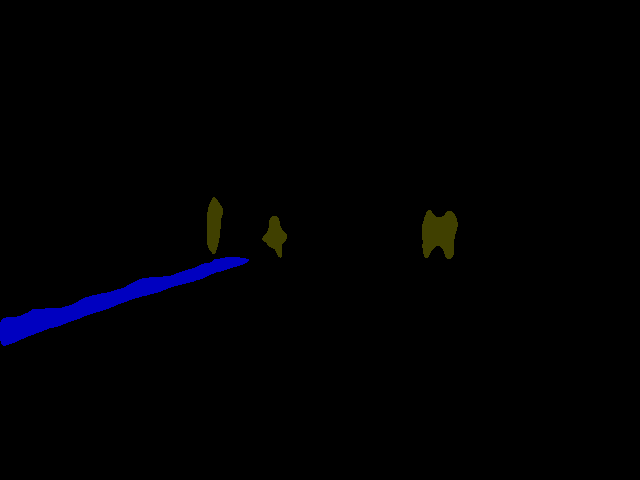}&
    \addFig{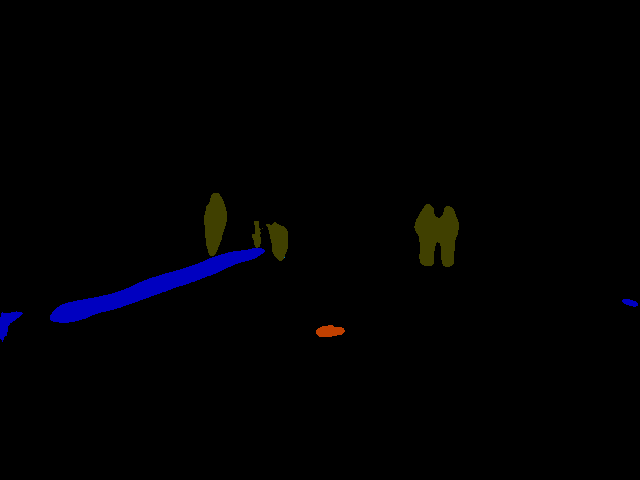}&
   \addFig{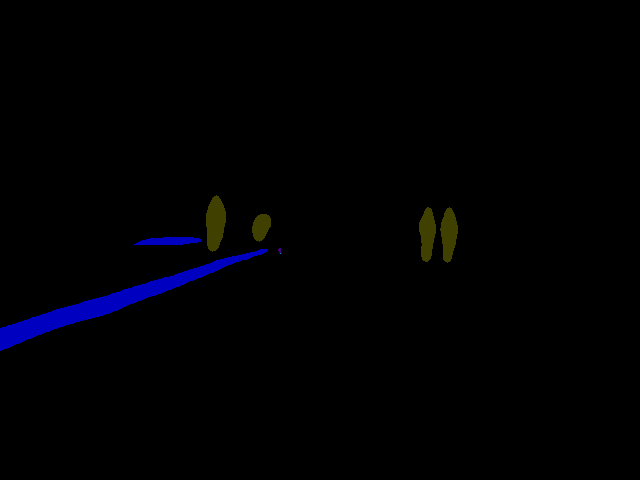}&
   \addFig{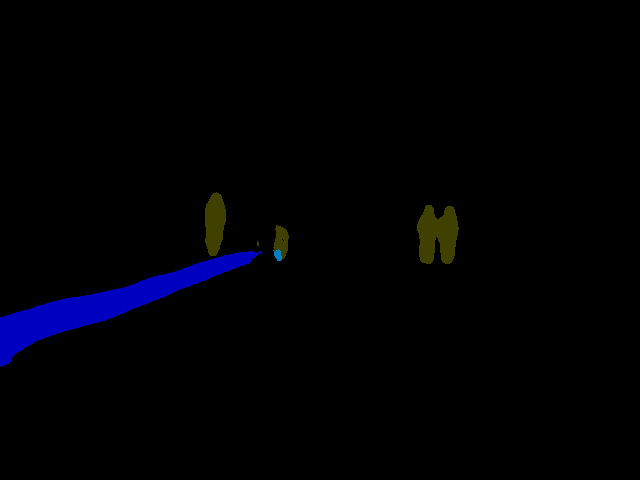}&
   \addFig{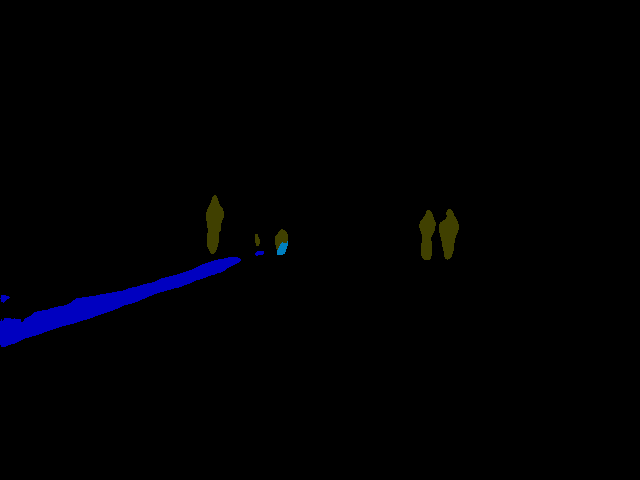}&
   \addFig{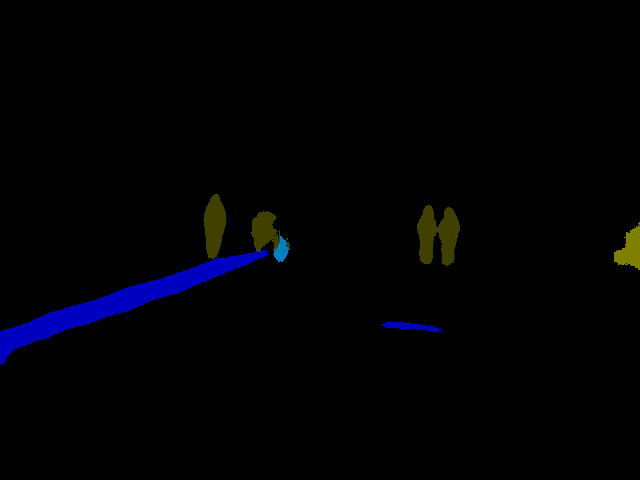}&
   \addFig{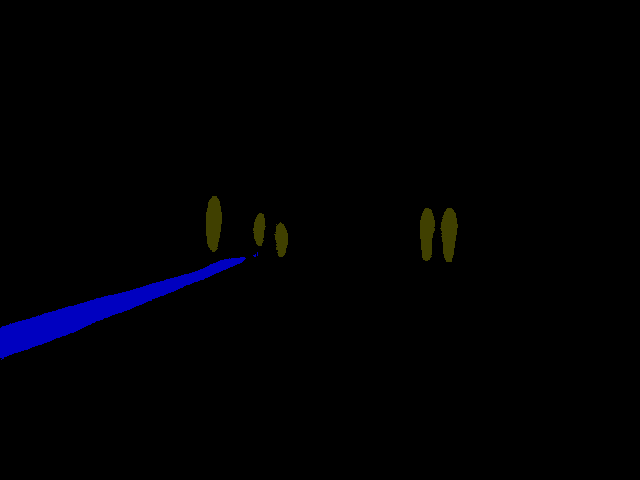}&
   \addFig{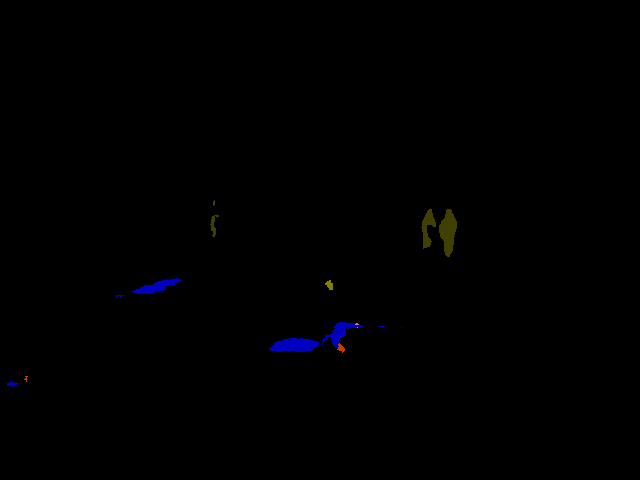}
   \\
   RGB & Thermal & GT & \textbf{Ours} & DSGBINet & GMNet & EGFNet & MMNet & ABMDRNet & RTFNet & MFNet \\
  \end{tabular}
  \caption{Visual comparisons of the proposed method and seven state-of-the-art methods in typical daytime and nighttime images of MFNet. The proposed CAINet provides suitable results under 
a variety of lighting conditions.}
  \label{fig:com_sota}
\end{figure*}
%
%
\textit{3) Implementation Details:}
We train our model by loading a pre-trained MobileNet-V2~\cite{MobileNet2} model on the ImageNet~\cite{deng2009imagenet} dataset. The RGB branch and the thermal image branch are first trained separately. Then their model parameters are loaded and the GCM decoder branch is trained. Finally, the trained basic model parameters are loaded and the whole model is trained together with the residual learning branch. The batch size of each iteration is 8 images.
The learning rate is initialized as $5\times 10^{-4}$. 
The network parameters are learned by back-propagating the model using the Adam optimization algorithm. We use an early termination algorithm to prevent network overfitting. The experiments are performed on the publicly available Pytorch 1.12.0~\cite{PyTorch} framework, using a workstation with a TITAN RTX GPU (24 GB RAM).
\begin{table}[!t]
  \centering
  \small
  \renewcommand{\arraystretch}{1.4}
  \renewcommand{\tabcolsep}{0.8mm}
\caption{Comparison of the number of parameters and execution efficiency of each method
}
\label{table:flops_and_params}
\begin{tabular}{ c | c c c c c }
\hline
Methods & Input size & FLOPs/G$\downarrow$ & Params/M$\downarrow$ &mAcc & mIoU \\
\hline
\hline
 RTFNet50~\cite{2019RTFNet} & 640$\times$480 & 245.71& 185.24 &62.2 & 51.7 \\
 RTFNet152~\cite{2019RTFNet} & 640$\times$480 &337.04 & 254.51 & 63.1 & 53.2  \\
 PSTNet~\cite{2020PSTNet} &640$\times$480 & 129.37 & 20.38 & - & 48.4\\
 FuseSeg~\cite{2021FuseSeg} &640$\times$480 & 193.40 & 141.52 & 70.6 & 54.5 \\
 ABMDRNet~\cite{2021ABMDRNet} &640$\times$480 &194.33& 64.60 &69.5& 54.8\\
 EGFNet~\cite{2022EGFNet} & 640$\times$480 &201.29 & 62.82 & 72.3 & 54.8 \\
 MTANet~\cite{2022MTANet} & 640$\times$480 &264.69 & 121.58 & 75.2 & 56.1 \\
 LASNet~\cite{2023LASNet} & 640$\times$480 & 233.81& 93.58 &\textbf{75.4} &54.9 \\
 Ours & 640$\times$480 & \textbf{123.62} & \textbf{12.16} & 73.2 & \textbf{58.6} \\
\hline
\end{tabular}
\end{table}

\subsection{Comparison with State-of-the-art Methods}
\textit{1) Evaluation on MFNet dataset:} We compare our CAINet with nineteen state-of-the-art methods, most of which are recently proposed RGB-T semantic segmentation methods, including MFNet~\cite{2017MFNet}, two versions of RTFNet~\cite{2019RTFNet}, PSTNet~\cite{2020PSTNet}, MLFNet~\cite{2021MLFNet}, FuseSeg~\cite{2021FuseSeg}, ABMDRNet~\cite{2021ABMDRNet},  MMNet~\cite{2022MMNet}, EGFNet~\cite{2022EGFNet}, FEANet~\cite{2021FEANet}, MTANet~\cite{2022MTANet}, MFFENet~\cite{2022MFFENet}, GMNet~\cite{2021GMNet}, CCFFNet~\cite{2022CCFFNet}, 
CCAFFMNet~\cite{2022CCAFFMNet},
DSGBINet~\cite{2022DSGBINet}, CMX-SegF-B2~\cite{liu2022cmx}, FDCNet~\cite{2022FDCNet}, and ECGFNet~\cite{zhou2023embedded}, and LASNet~\cite{2023LASNet}. 
The quantitative results on MFNet dataset are shown in Tab.~\ref{table:QuantitativeResults_MFNet}, including eight categories and overall average performance. 
Specifically, our proposed CAINet achieves the highest performance with 58.6\% in mIoU, outperforming the second-best methods by 0.4\% (\ie CCAFFMNet and CMX-SegF-B2 with 58.2\% in mIoU). 
Among all categories, our method performs very well in the bike and bump, outperforming the second-best method by 2.9\% in mAcc and 1.2\% in mIoU in the bike, and 7.7\% in mAcc and in 1.3\% mIoU in the bump, respectively.
%

Of course, our model also has a slight limitation in that for small and obscured objects, our method will segment the contiguous morphology of the target, but compared to other methods, most do not miss detection, as shown in Fig.~\ref{fig:com_sota}.
 Based on the visualizations, it can be inferred that the model proposed in this study outperforms the other methods, despite the possibility of some roadblocks
 being undetected. Overall, the CAINet model demonstrated favorable results in scenarios of high background lighting, as thermal imaging provides a reliable representation of heat that remains consistent across diverse lighting conditions.

\begin{table*}[t!]
  \centering
  \small
  \renewcommand{\arraystretch}{1.5}
  \renewcommand{\tabcolsep}{1.4mm}
  \caption{
    Quantitative comparisons (\%) on the test set of PST900 dataset. `-' means that the authors do not provide the corresponding results. The top two results in each column are highlighted in \textcolor{red}{\textbf{red}} and \textcolor{blue}{\textbf{blue}}.
    }
\label{table:QuantitativeResults_PST900}
\begin{tabular}{r|c|cccccccccccc}
\midrule[1pt]    
 \multirow{2}{*}{\normalsize{Methods}} &  \multirow{2}{*}{\normalsize{Methods}}
 & \multicolumn{2}{c}{Background} & \multicolumn{2}{c}{Hand-Drill} & \multicolumn{2}{c}{Backpack} & \multicolumn{2}{c}{Fire-Extinguisher} & \multicolumn{2}{c}{Survivor}   
 & \multirow{2}{*}{\normalsize{mAcc}} & \multirow{2}{*}{\normalsize{mIoU}} \\
 
 \cline{3-12} 
       & & Acc & IoU & Acc & IoU & Acc & IoU & Acc & IoU & Acc & IoU \\
	     
\midrule[1pt] 
MFNet$_{17}$~\cite{2017MFNet}          & RGB-T & - & 98.63 & - & 41.13 & - & 64.27 & - & 60.35 & - & 20.70 & - & 57.02 \\	
PSTNet$_{20}$~\cite{2020PSTNet}          & RGB-T & - & 98.85 & - & 53.60 & - & 69.20 & - & 70.12 & - & 50.03 & - & 68.36 \\	
EGFNet$_{22}$~\cite{2022EGFNet}           & RGB-T & 99.48 & 99.26 & \textcolor{red}{\textbf{97.99}} & 64.67 & \textcolor{blue}{\textbf{94.17}} & 83.05 & \textcolor{red}{\textbf{95.17}} & 71.29 & 83.30 & 74.30 & \textcolor{blue}{\textbf{94.02}} & 78.51 \\
MTANet$_{22}$~\cite{2022MTANet}          & RGB-T & - & 99.33 & - & 62.05 & - & \textcolor{blue}{\textbf{87.50}} & - & 64.95 & - & \textcolor{red}{\textbf{79.14}} & - & 78.60 \\
MFFENet$_{21}$~\cite{2022MFFENet}        & RGB-T & - & 99.40 & - & 72.50 & - & 81.02 & - & 66.38 & - & 75.60 & - & 78.98 \\	

GMNet$_{21}$~\cite{2021GMNet}            & RGB-T & \textcolor{blue}{\textbf{99.81}} & 99.44 & 90.29 & \textcolor{red}{\textbf{85.17}} & 89.01 & 83.82 & 88.28 & 73.79 & 80.86 & 78.36 & 89.61 & 84.12 \\	
			
CCFFNet50$_{22}^{*}$~\cite{2022CCFFNet}           & RGB-T & \textcolor{red}{\textbf{99.9}} & 99.4 & 89.7 & \textcolor{blue}{\textbf{82.8}} & 77.5 & 75.8 & 87.6 & \textcolor{red}{\textbf{79.9}} & 79.7 & 72.7 & 86.9 & 82.1 \\ 

DSGBINet$_{22}$~\cite{2022DSGBINet}  	& RGB-T & 99.73 & 99.39 & 94.53 & 74.99 & 88.65 & 85.11 & \textcolor{blue}{\textbf{94.78}} & \textcolor{blue}{\textbf{79.31}} & 81.37 & 75.56 & 91.81 & 82.87 \\
FDCNet$_{22}$~\cite{2022FDCNet}  	& RGB-T & 99.72 & 99.15 & 82.52 & 70.36 & 77.45 & 72.17 & 91.77 & 71.52 & 78.36 & 72.36 & 85.96 & 77.11 \\
LASNet$_{23}^{\#}$~\cite{2023LASNet}   & RGB-T & 99.77 & \textcolor{blue}{\textbf{99.46}} & \underline{91.81} &  \underline{82.80} &  90.80 &  86.48 &  \underline{92.36} & \underline{77.75} &  \textcolor{blue}{\textbf{83.43}} &  75.49 &  91.63 &  \textcolor{blue}{\textbf{84.40}} \\
\hline
\hline
\textbf{CAINet (Ours) }	 & RGB-T & 99.66 & \textcolor{red}{\textbf{99.50}} & \textcolor{blue}{\textbf{95.87}} & 80.30 & \textcolor{red}{\textbf{96.09}} & \textcolor{red}{\textbf{88.02}} & 88.38 & 77.21 & \textcolor{red}{\textbf{91.35}} & \textcolor{blue}{\textbf{78.69}} & \textcolor{red}{\textbf{94.27}} & \textcolor{red}{\textbf{84.74}} \\
\toprule[1pt]
\multicolumn{14}{l}{\footnotesize{$^{*}$: The performance of CCFFNet50 is excerpted from the original paper, which originally keep one decimal place.}} \\
\multicolumn{14}{l}{\footnotesize{$^{\#}$: We correct the values of “Hand-Drill” and “Fire-Extinguish” in LASNet, as indicated by underlining.}}
\end{tabular}
\end{table*}

%
%

%
\renewcommand{\addFig}[1]{{\includegraphics[height=0.06\textwidth]{com_pst9001/#1}}}
\begin{figure*}[t]
  \centering
  \small
  \renewcommand{\arraystretch}{0.5}
  \setlength{\tabcolsep}{0.3mm}
  \begin{tabular}{cccccccccccc}
   \addFig{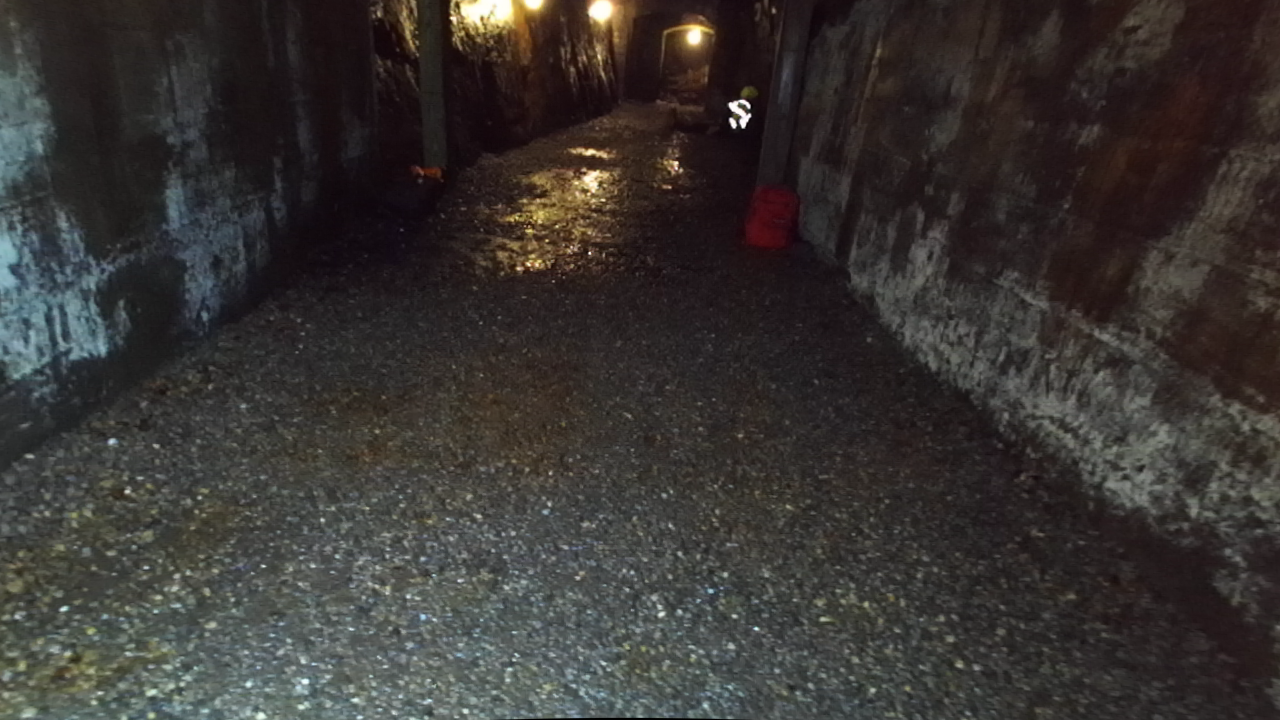}&
    \addFig{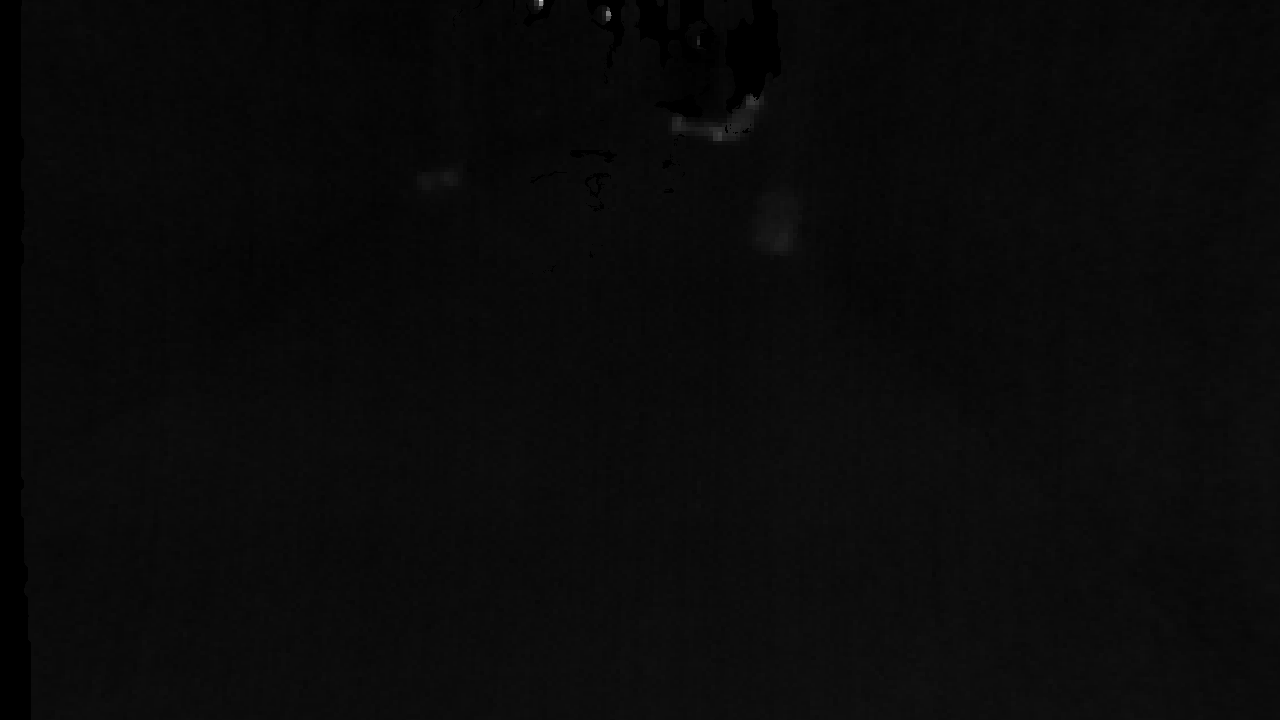}&
    \addFig{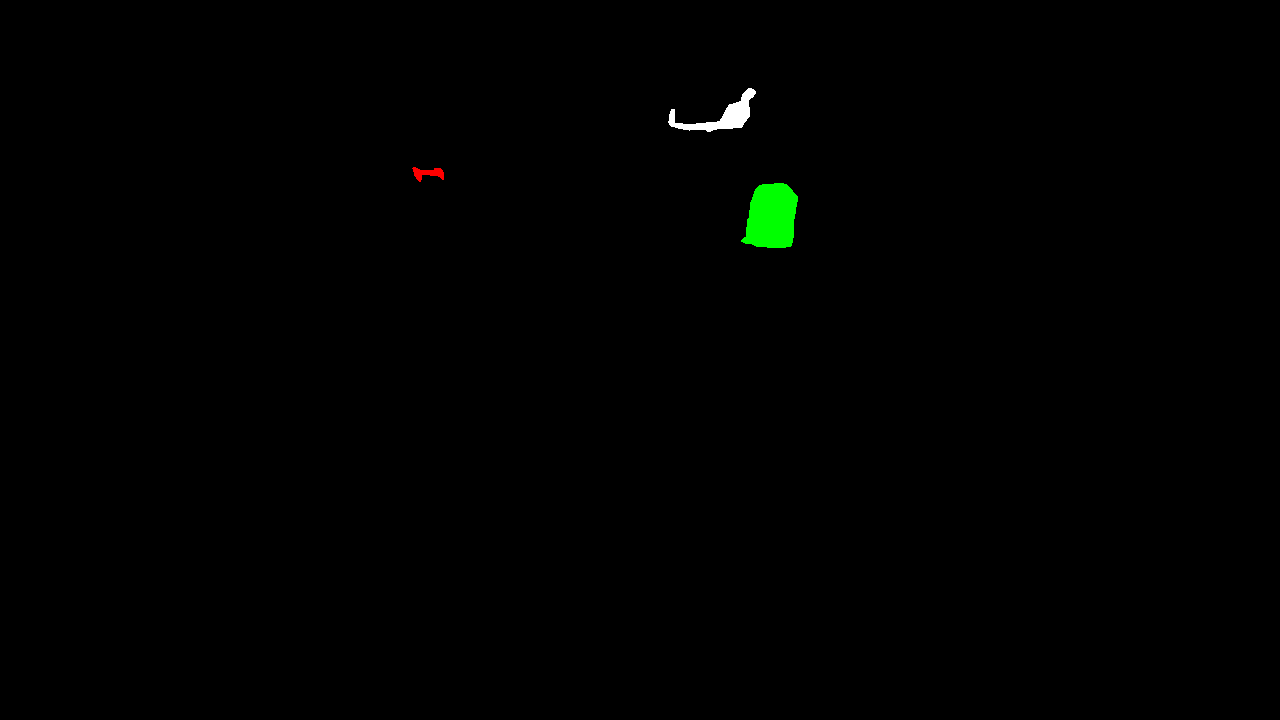}&
    \addFig{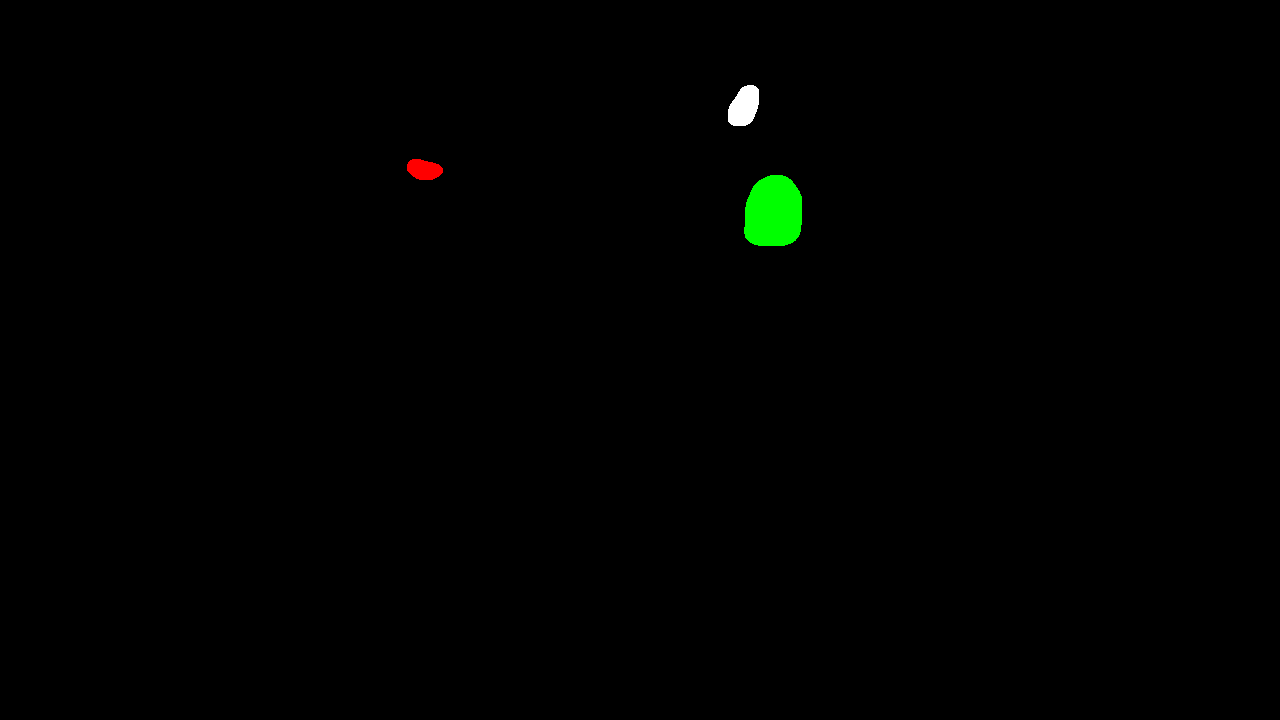}&
    \addFig{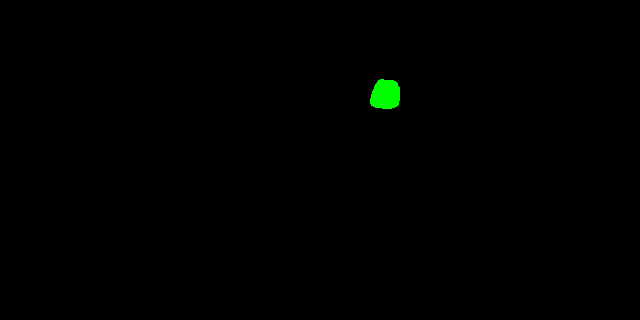}&
    \addFig{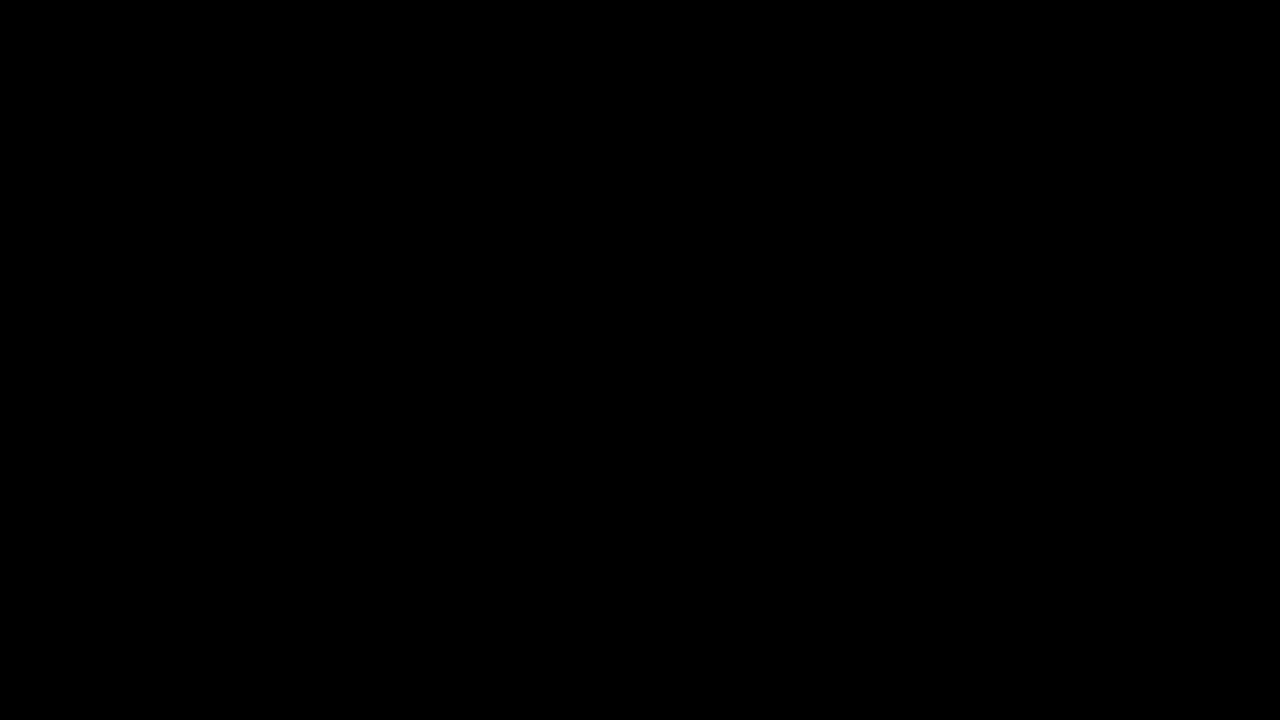}&    
    \addFig{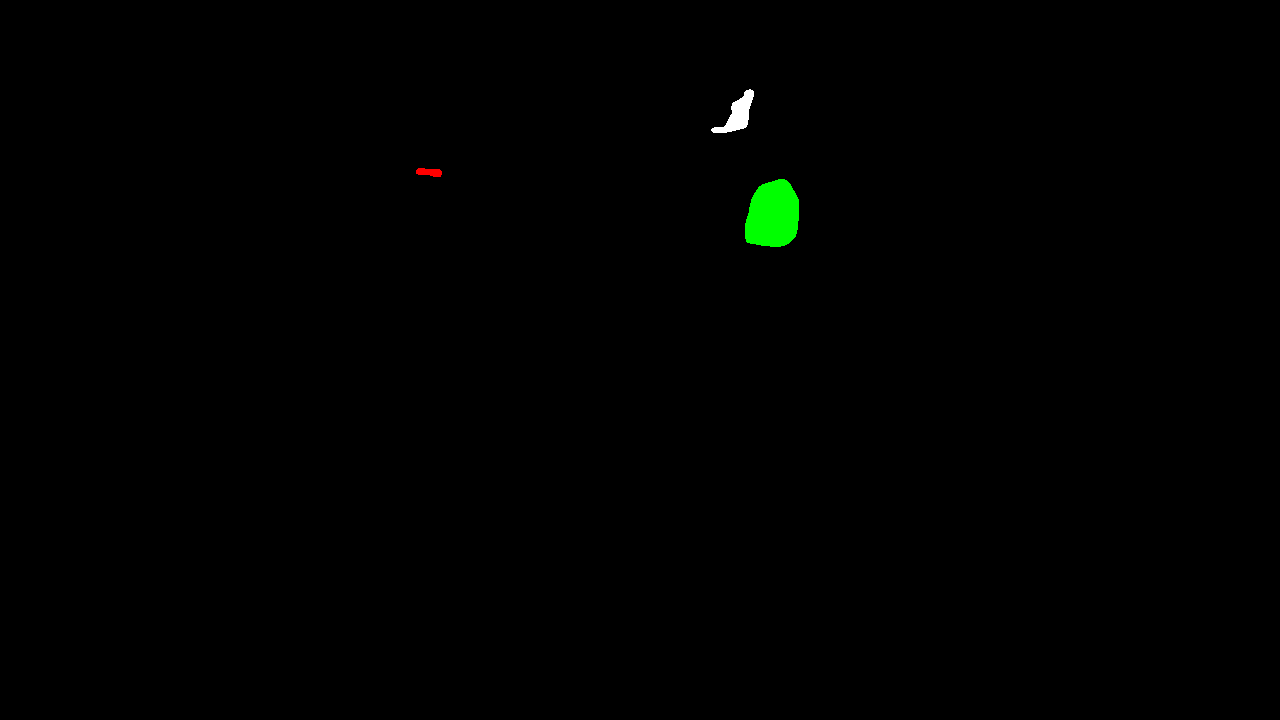}
   \\
    \addFig{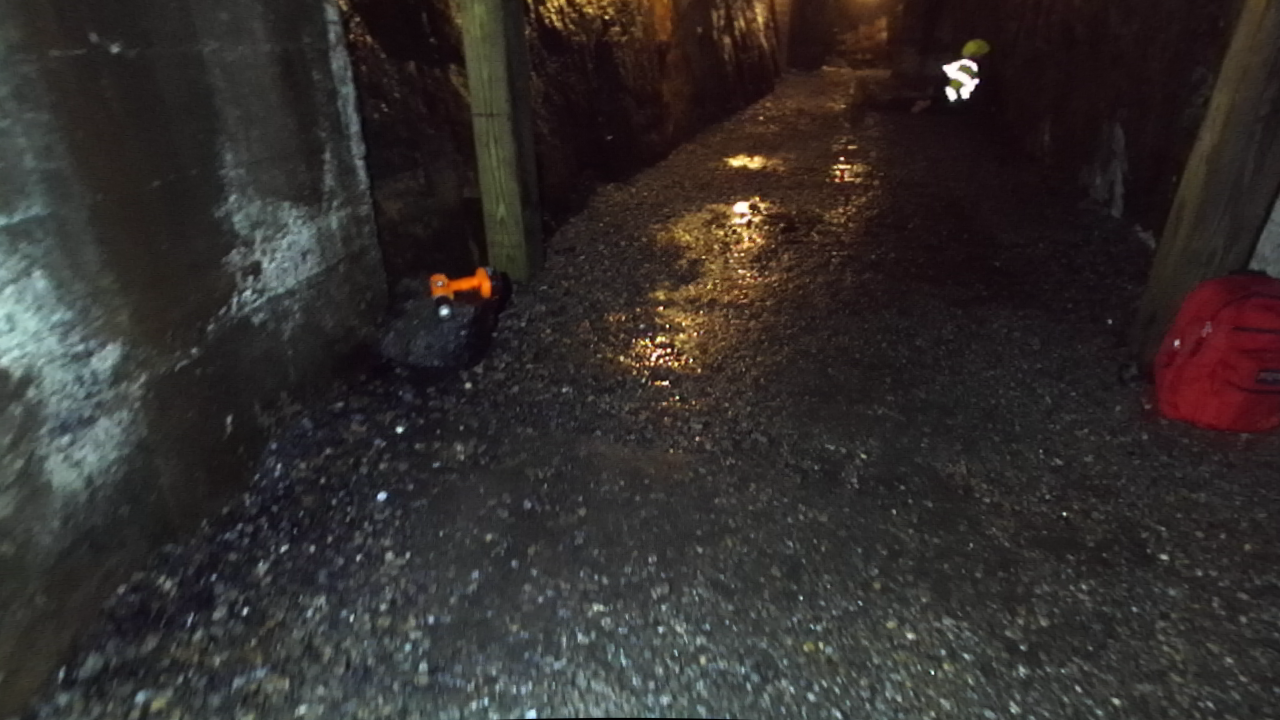}&
    \addFig{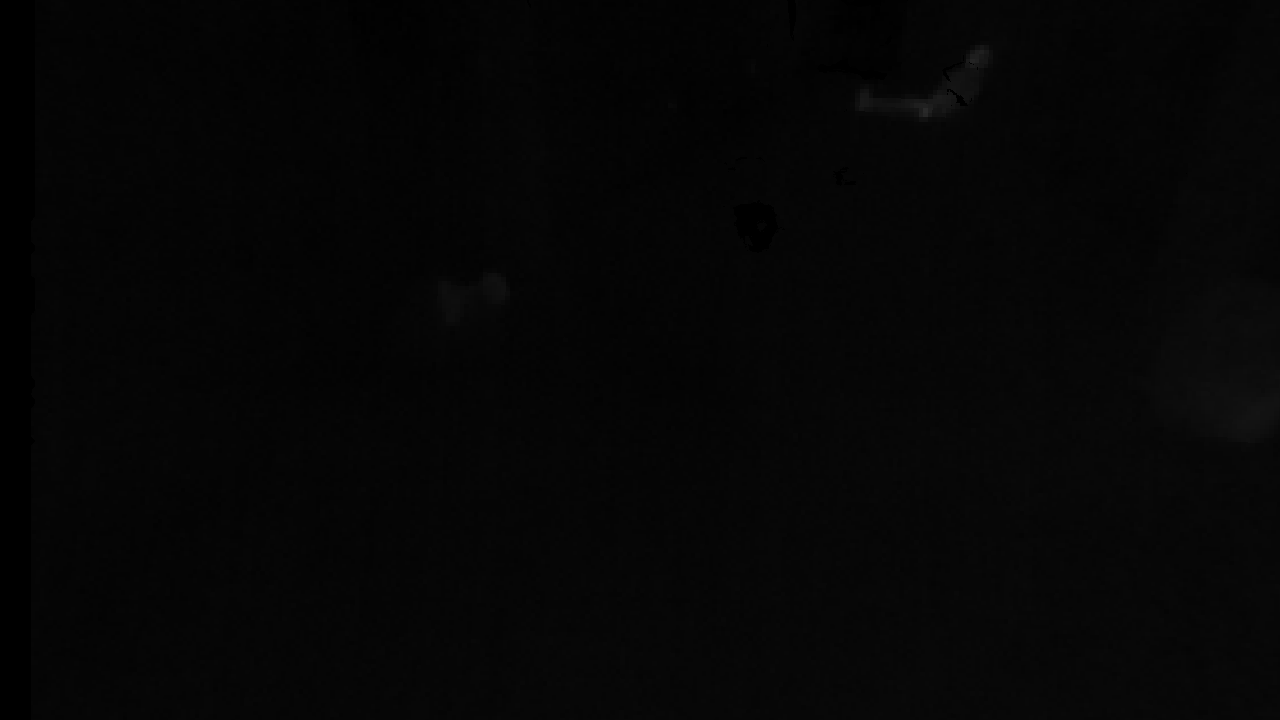}&
    \addFig{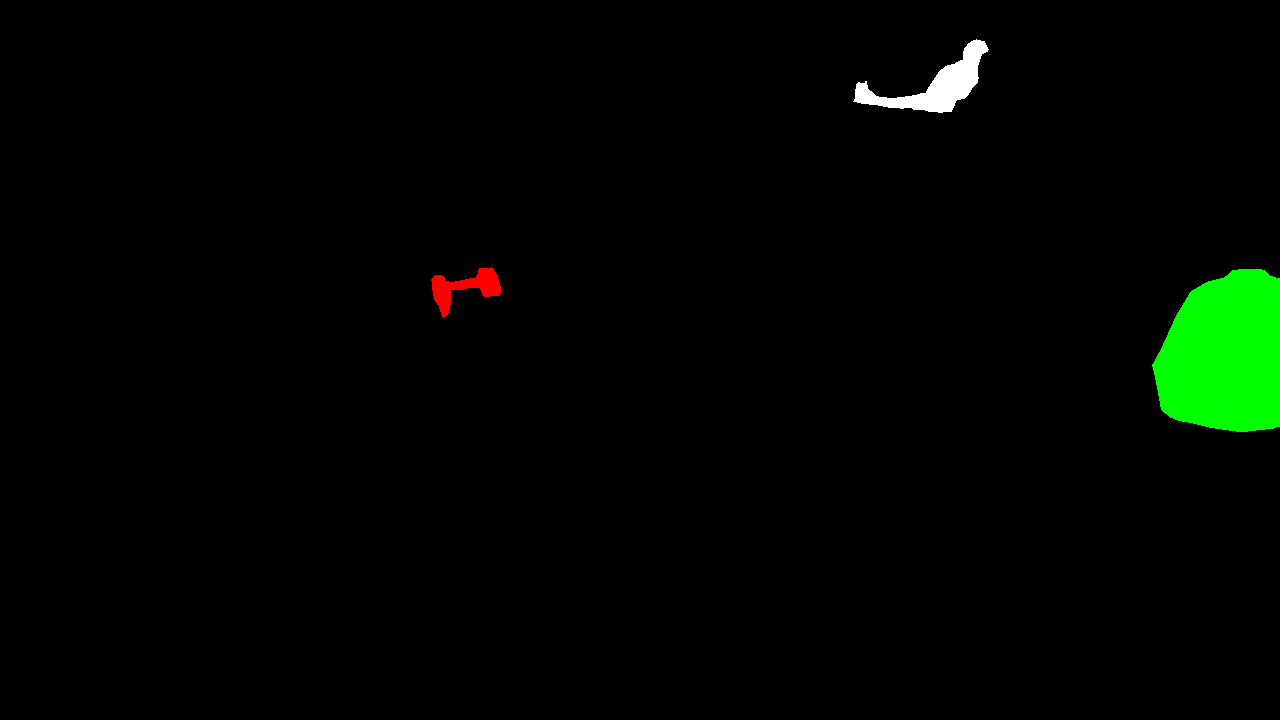}&   \addFig{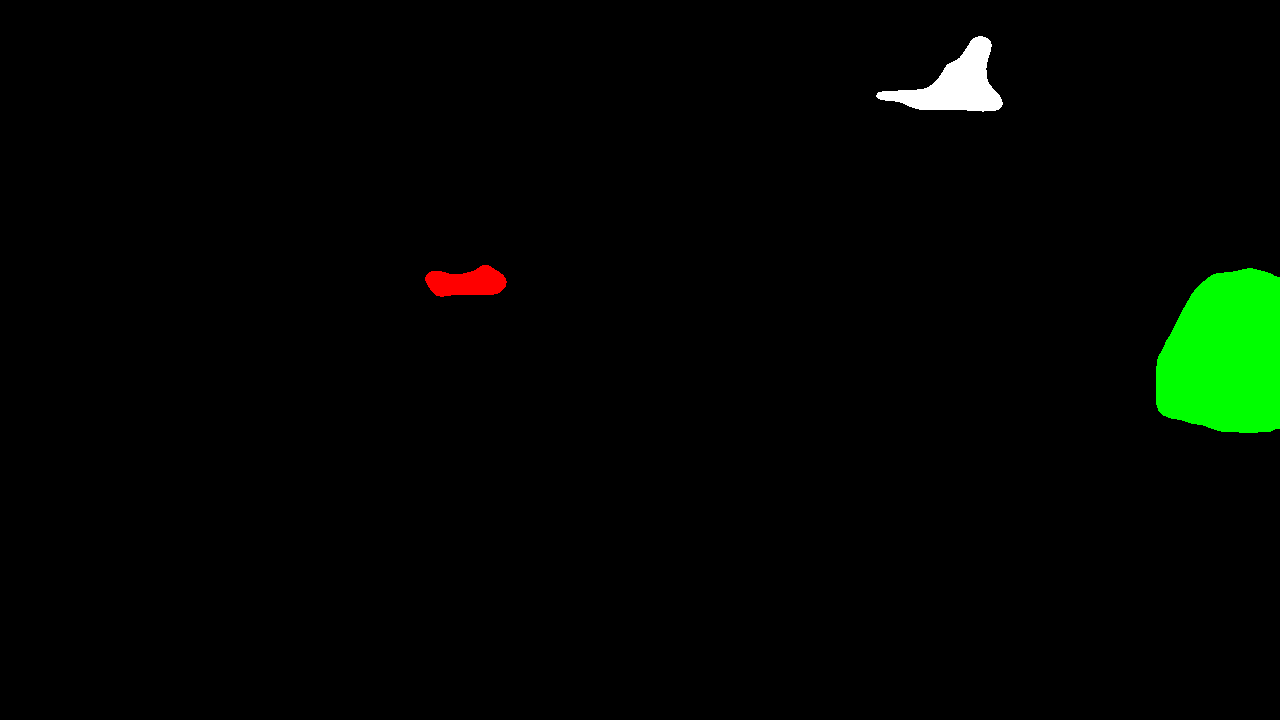}&
    \addFig{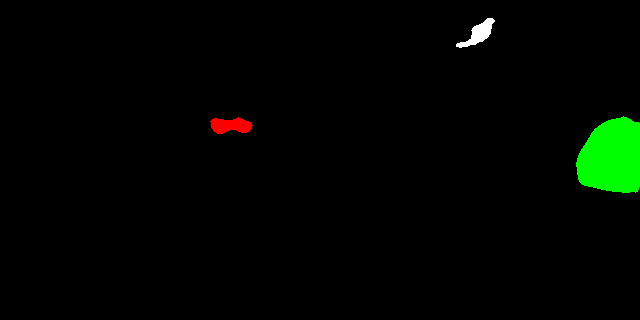}&
    \addFig{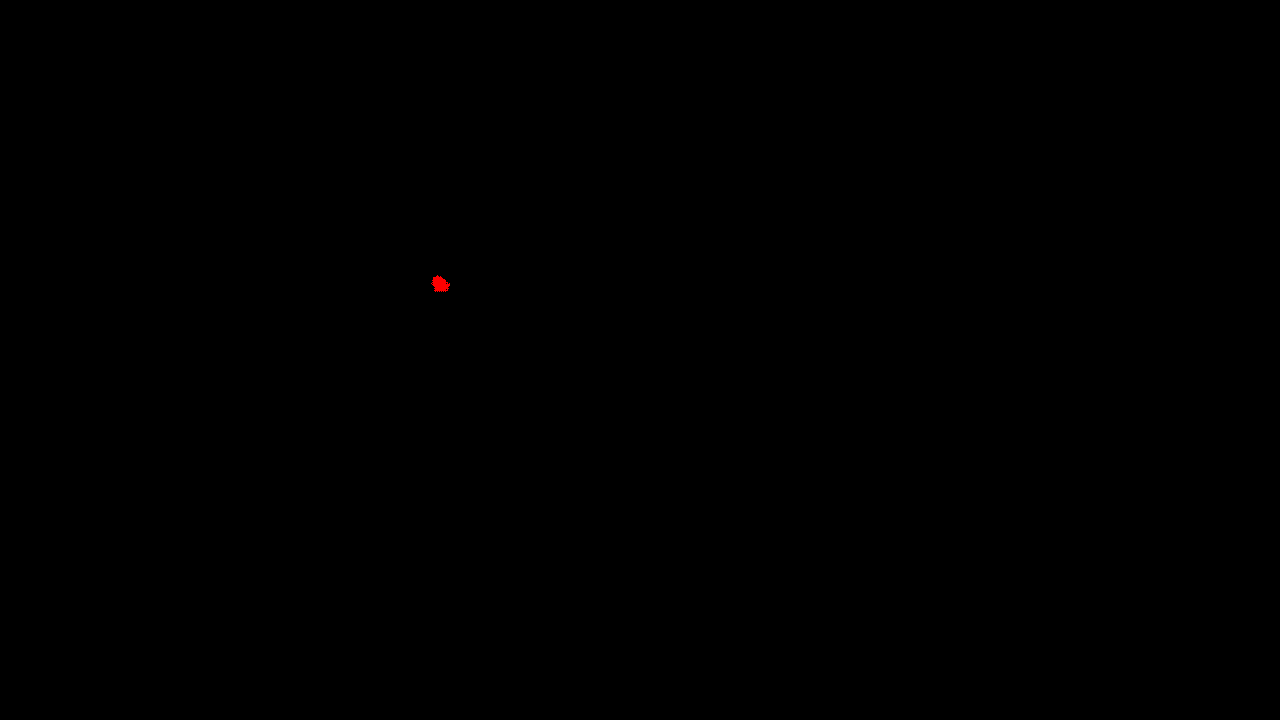}&    
    \addFig{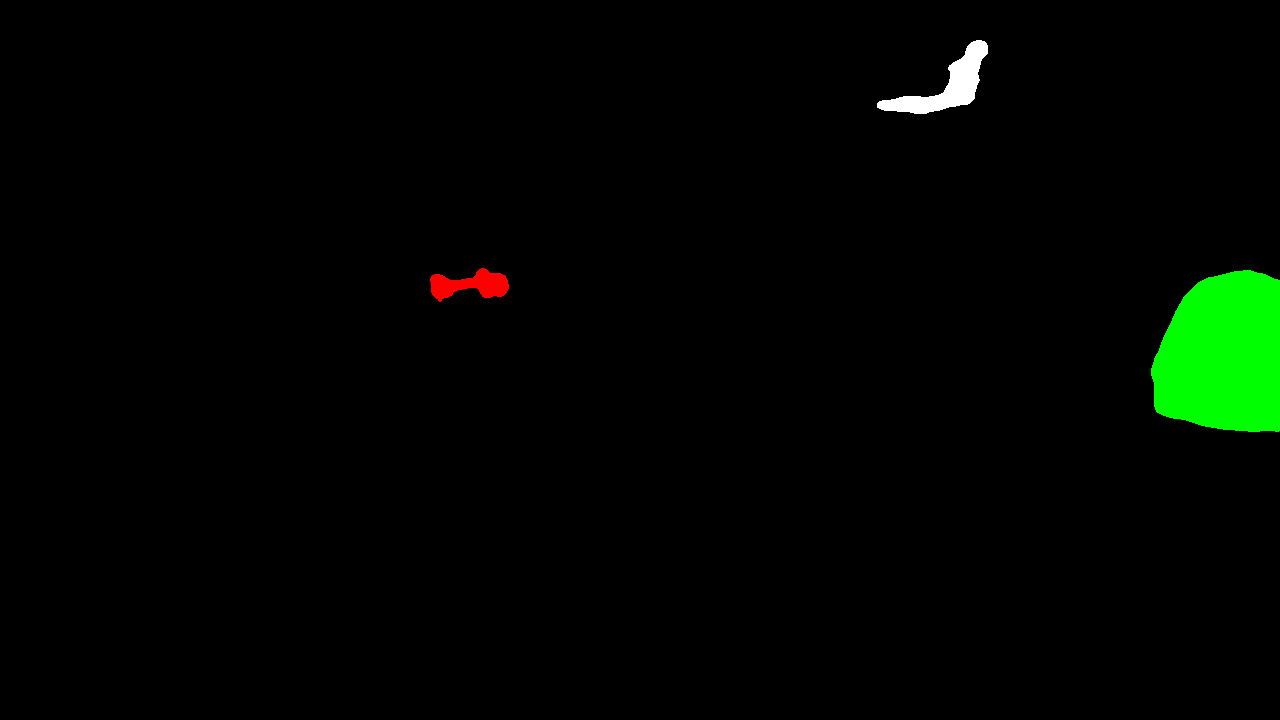}
   \\
    \addFig{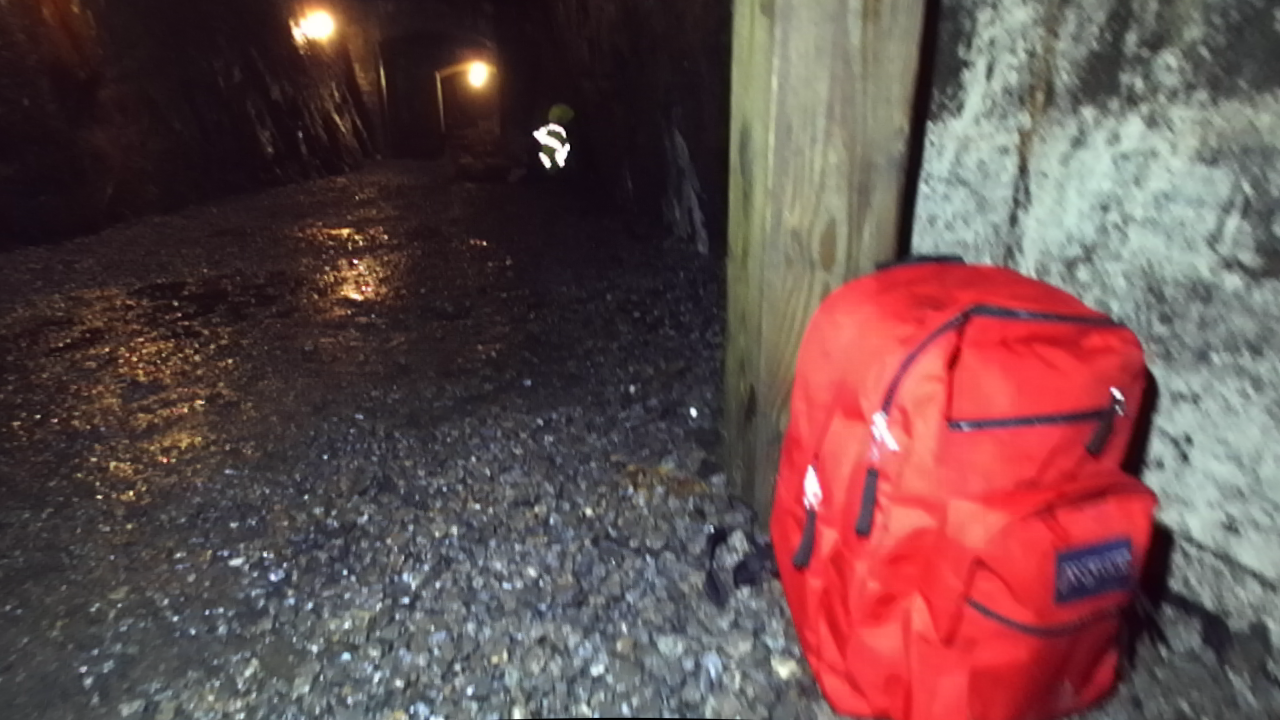}&
    \addFig{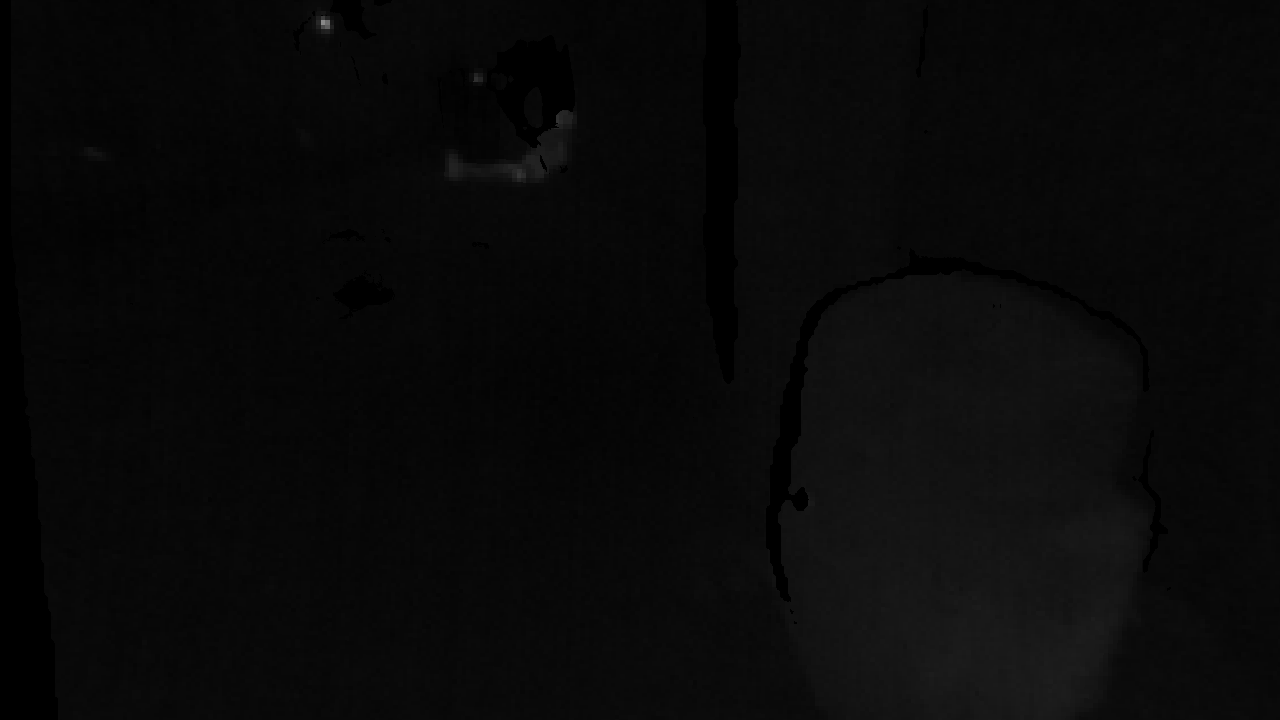}&
    \addFig{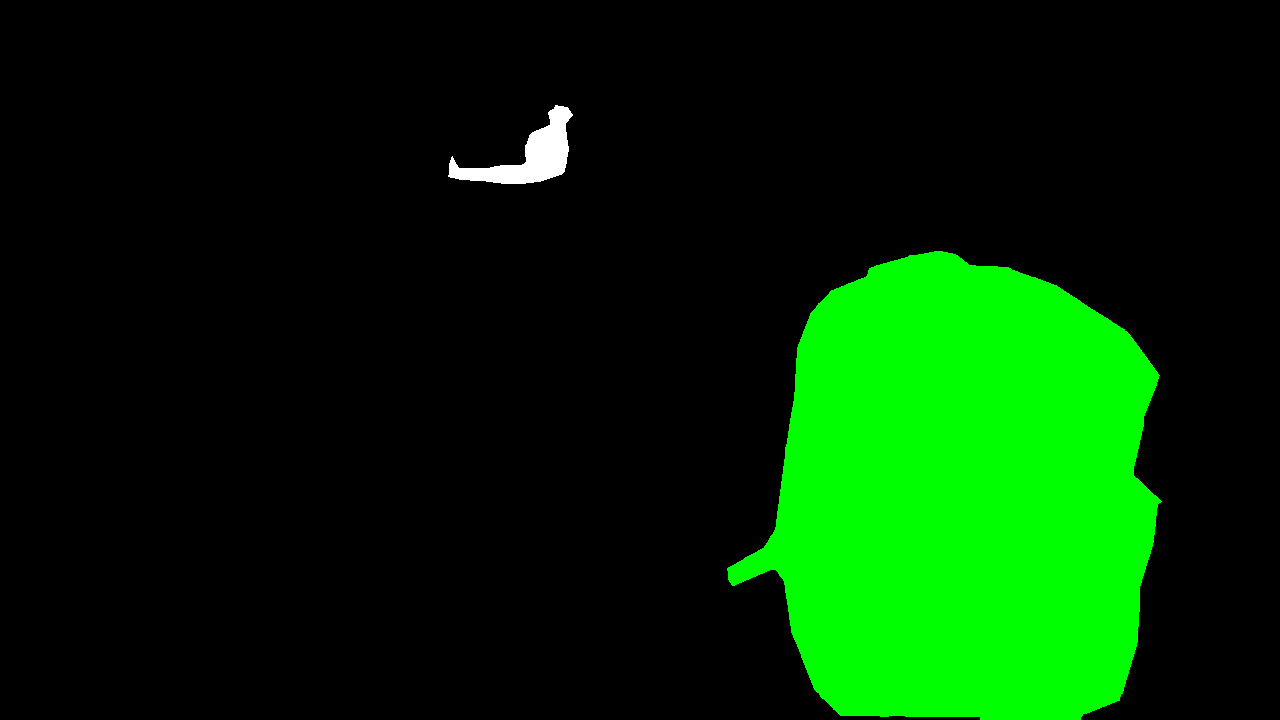}&   \addFig{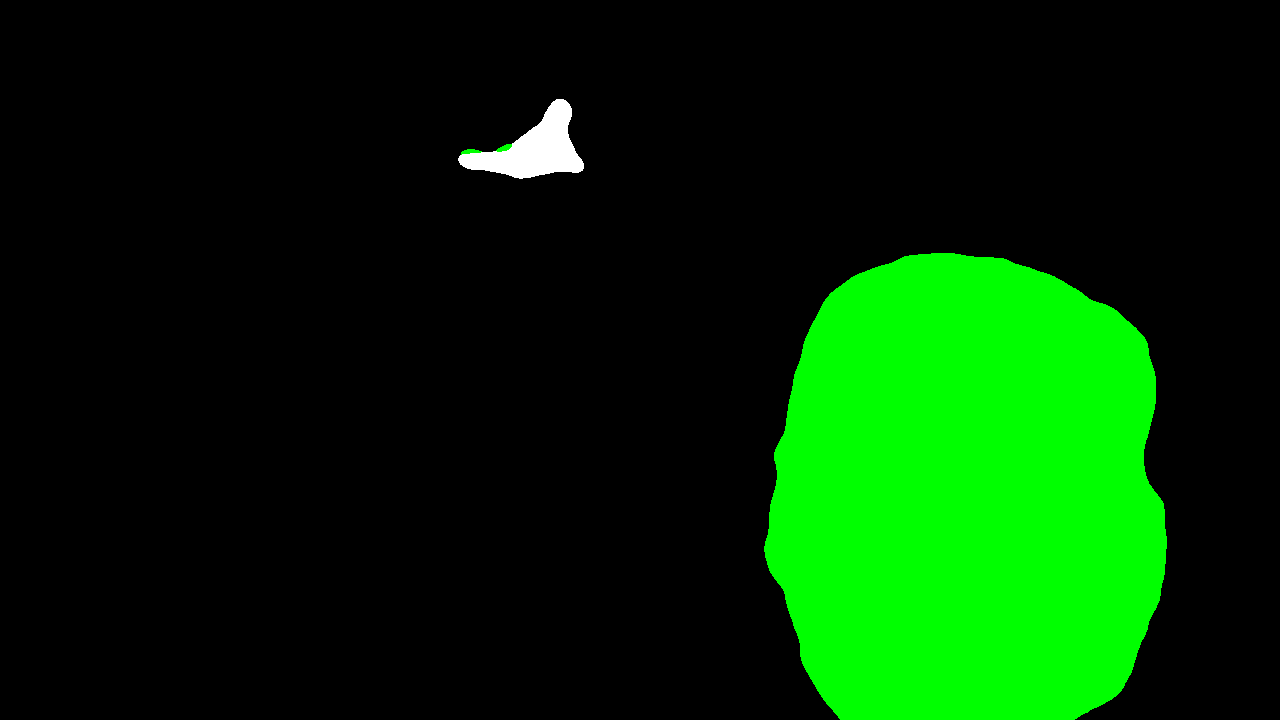}&
    \addFig{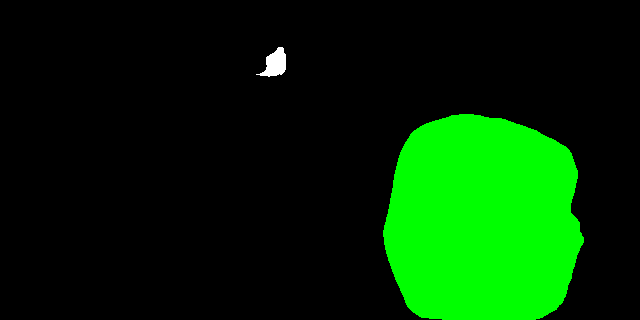}&
    \addFig{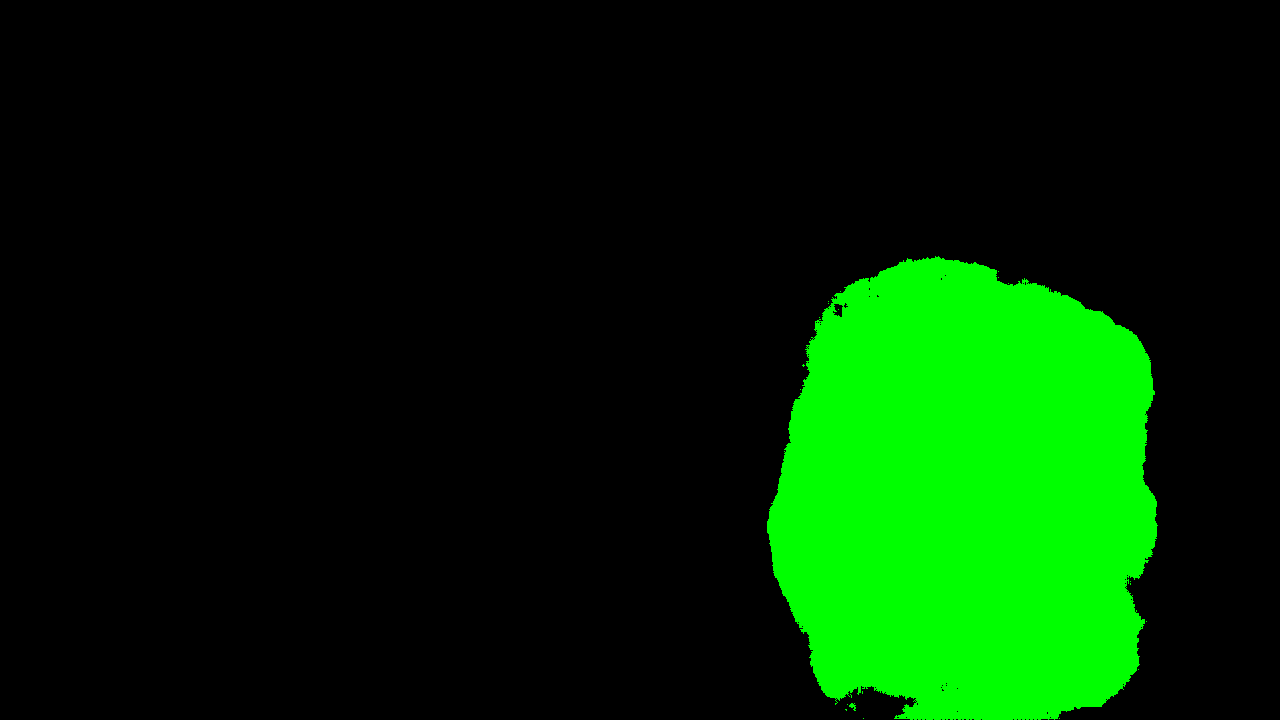}&    
    \addFig{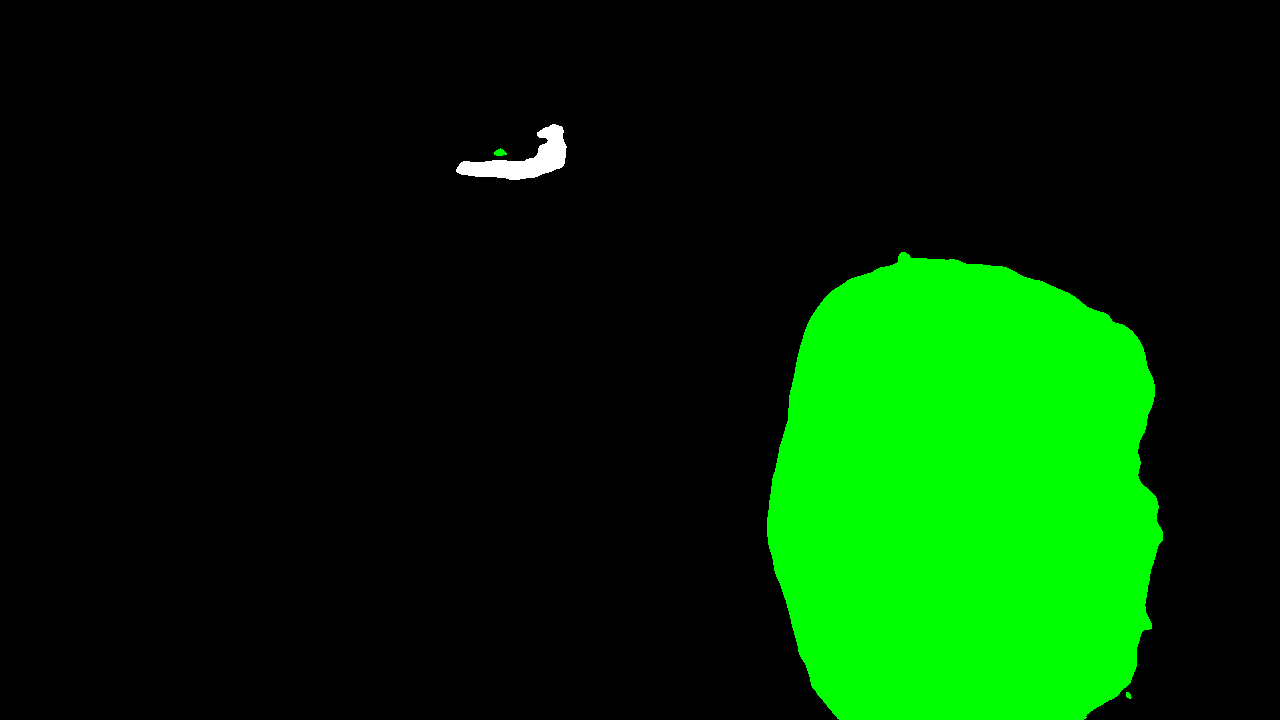}
   \\
   \addFig{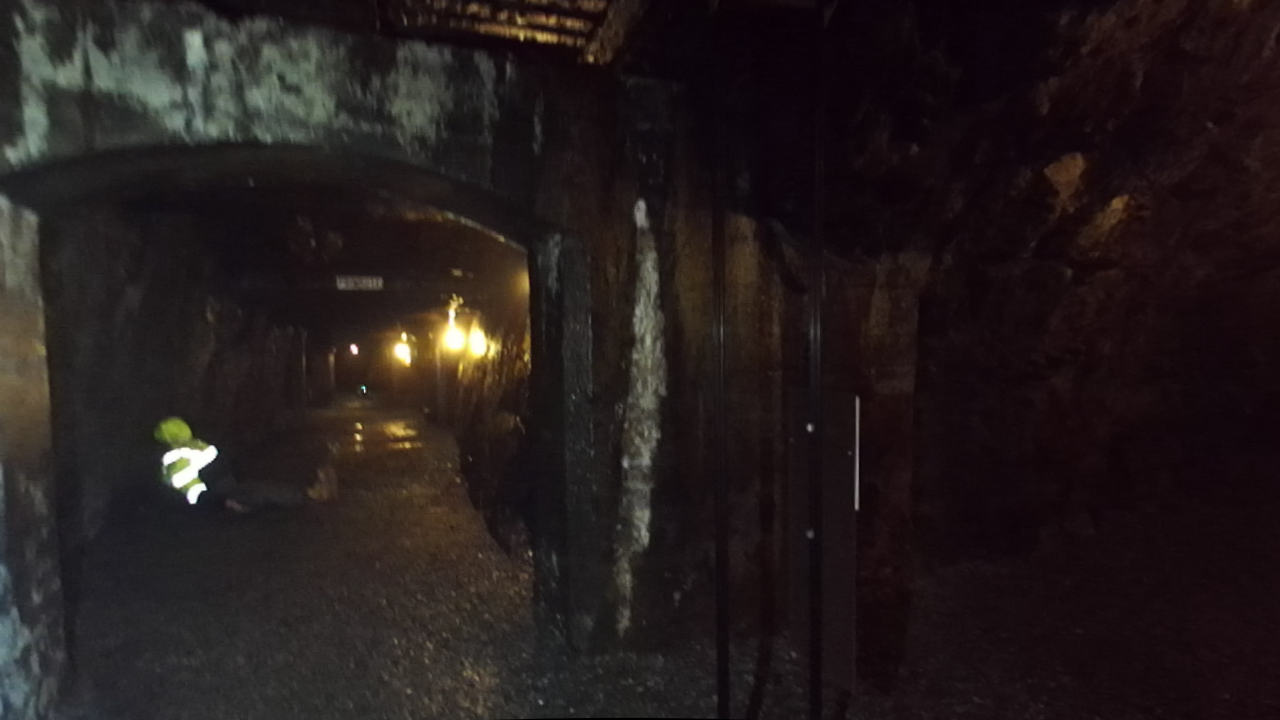}&
    \addFig{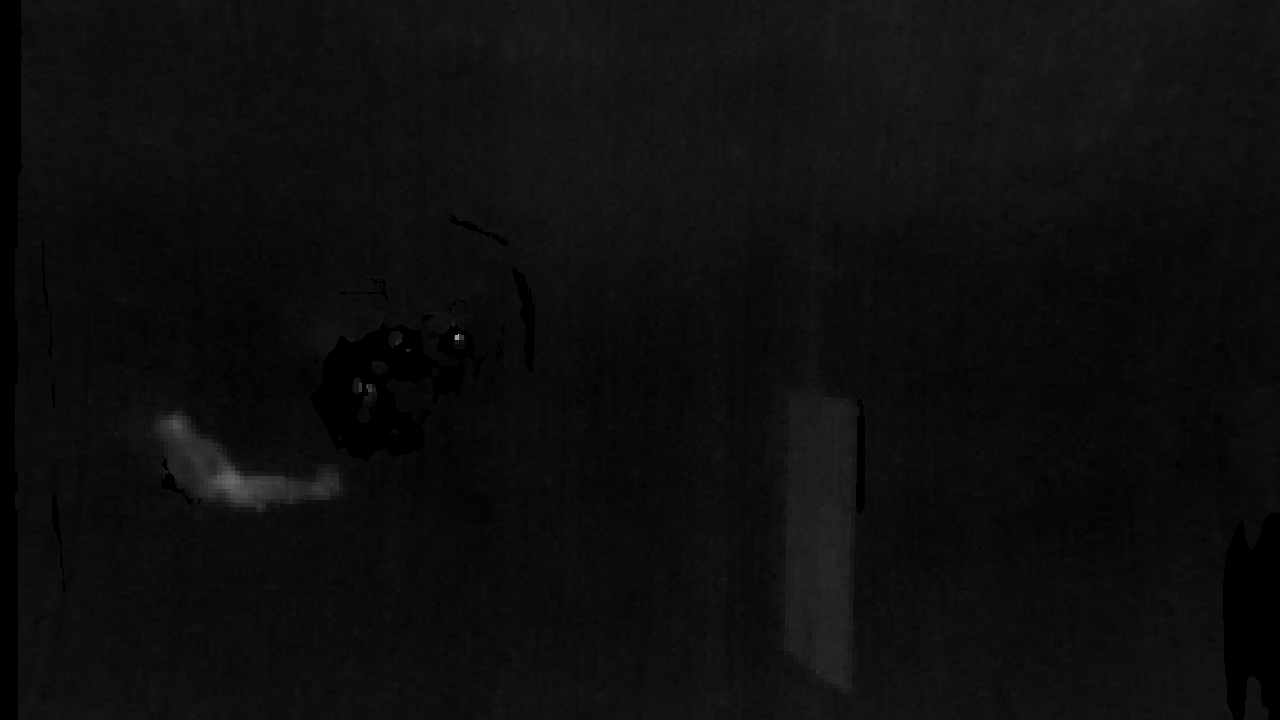}&
    \addFig{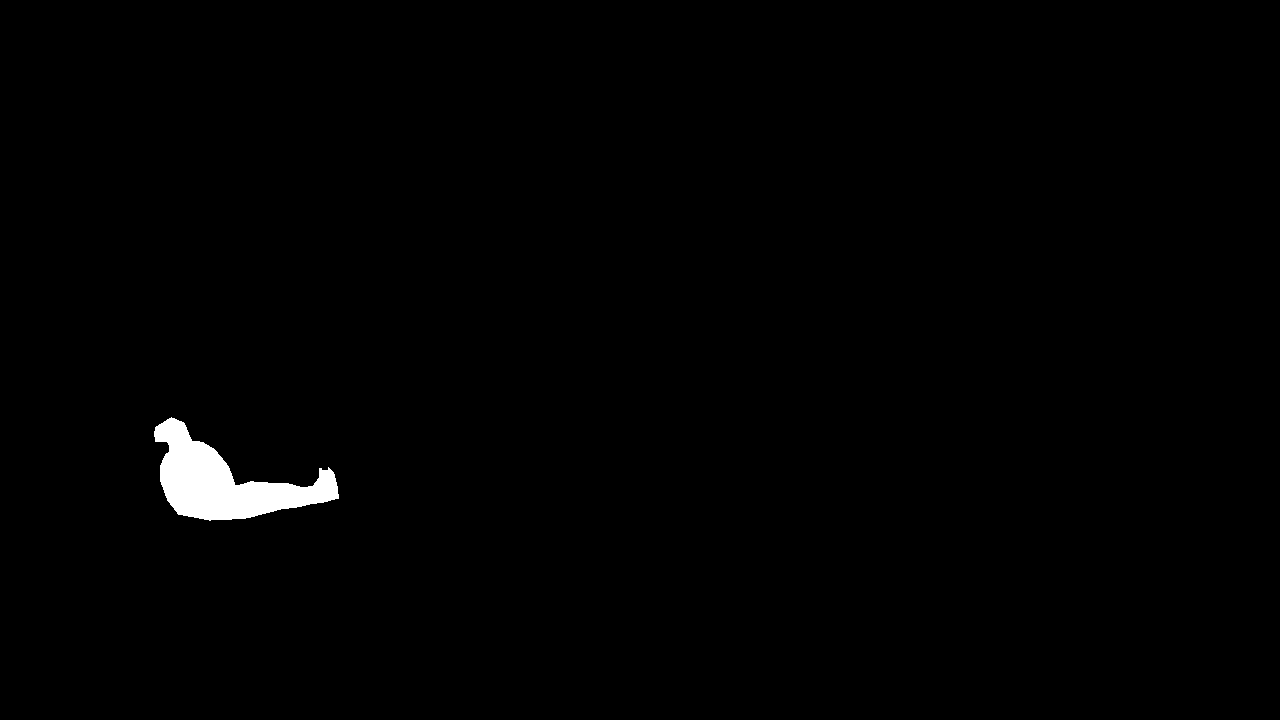}&   \addFig{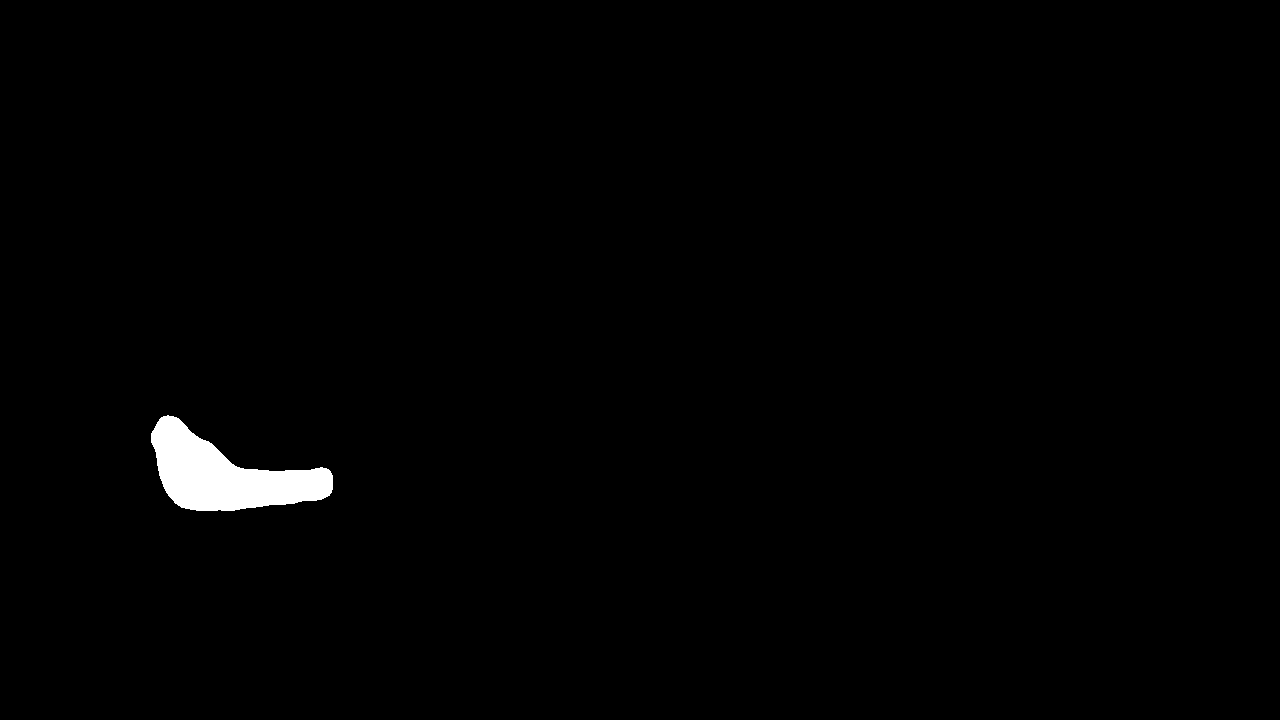}&
    \addFig{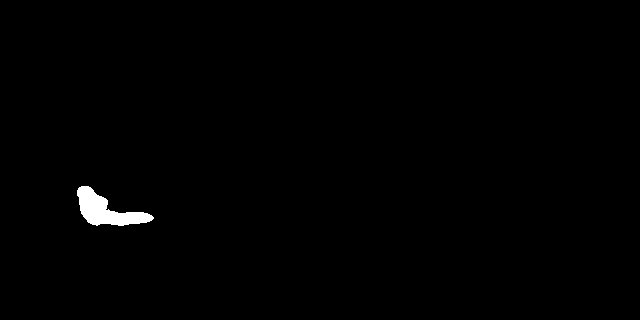}&
   \addFig{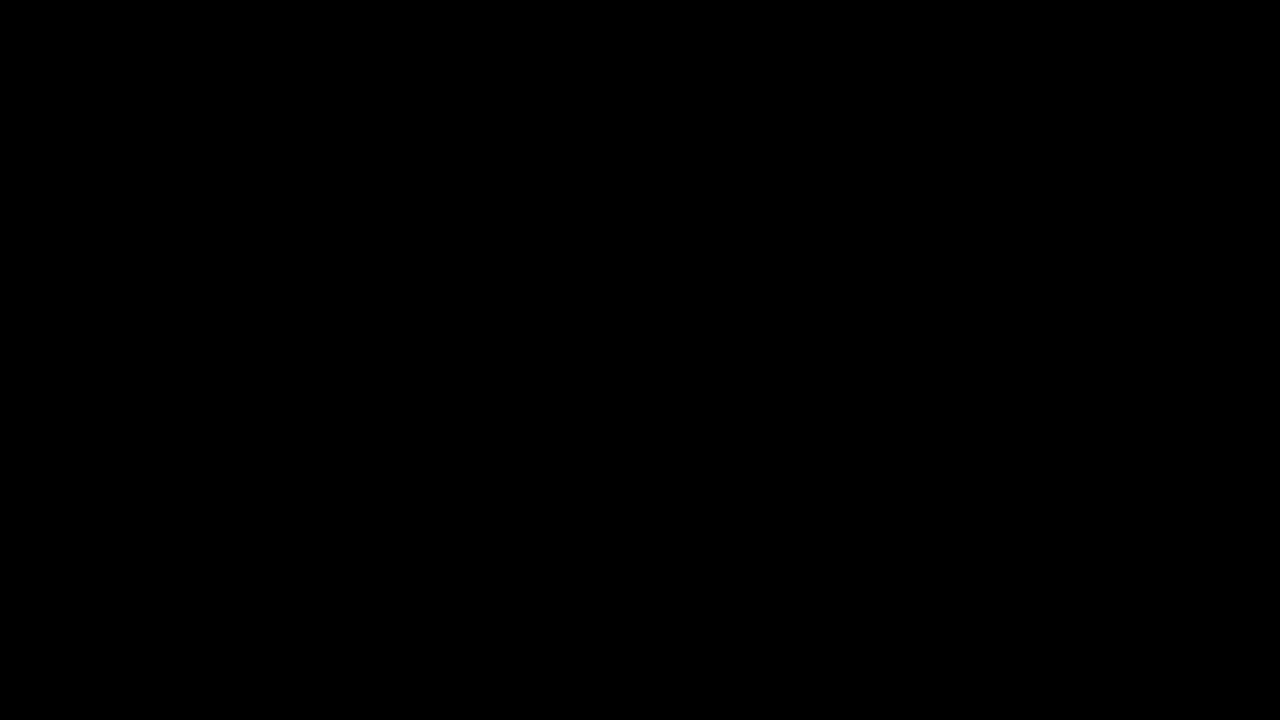}&   
    \addFig{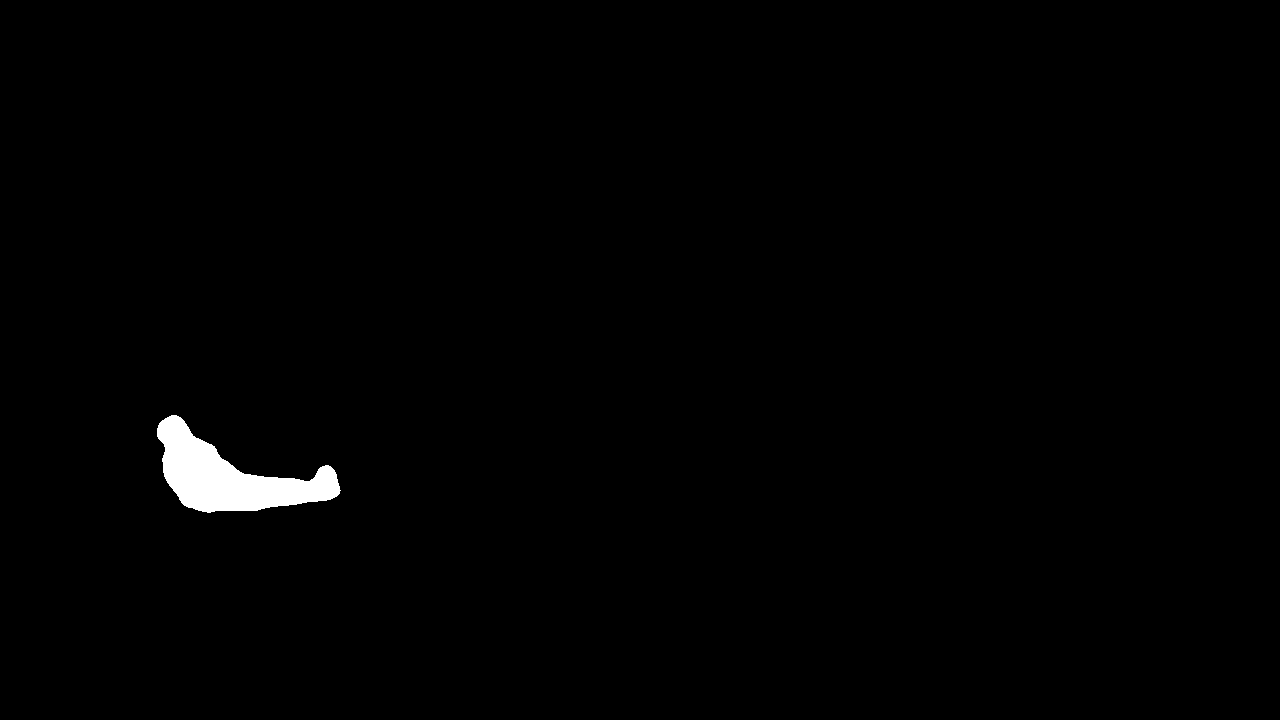}
   \\
   \addFig{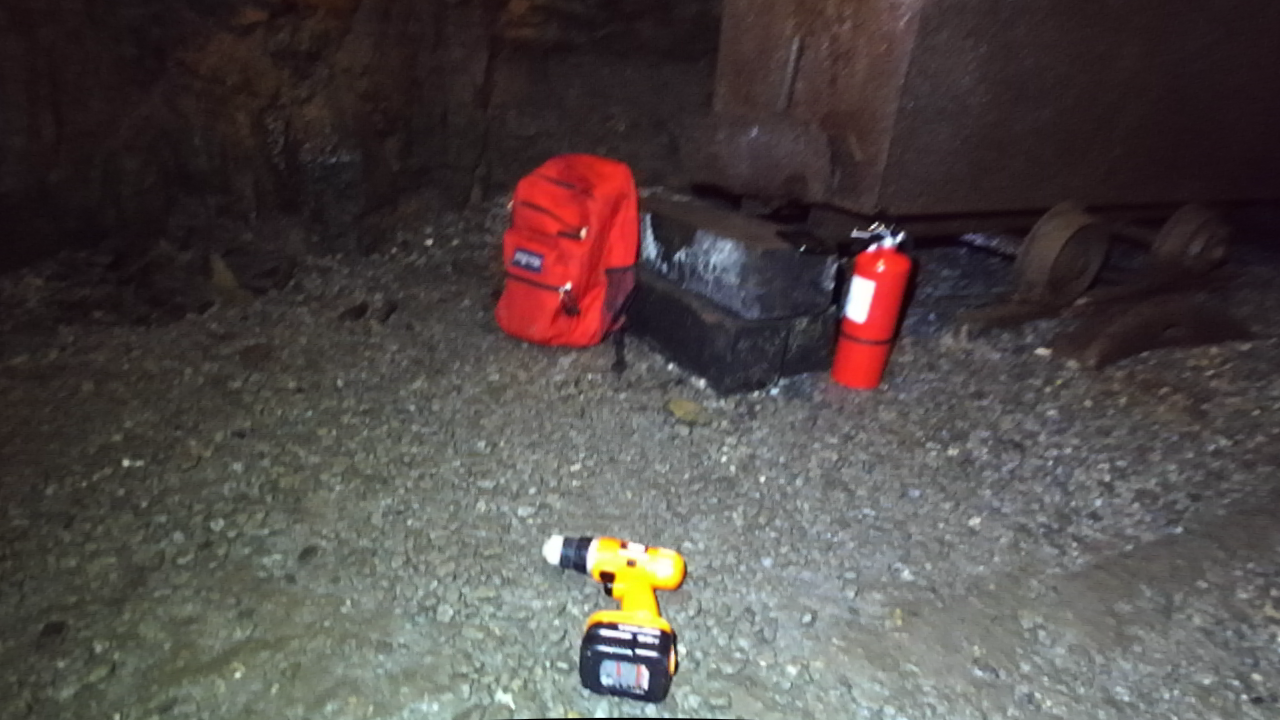}&
    \addFig{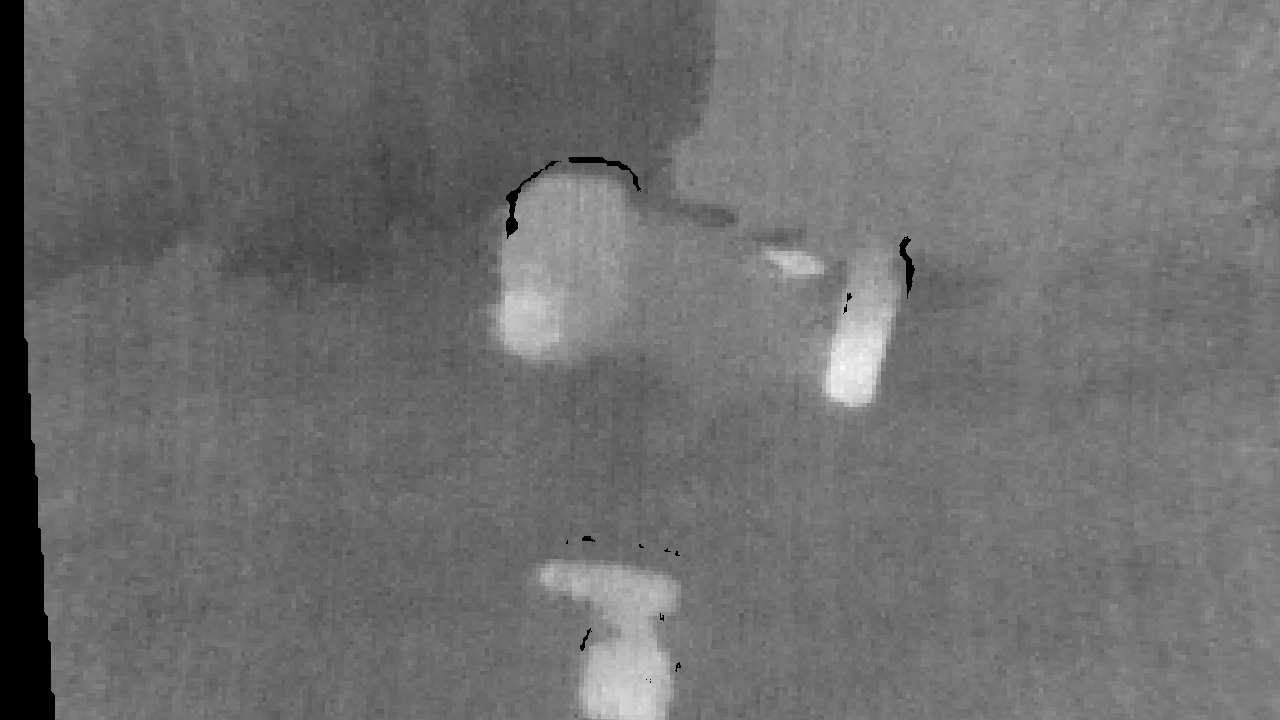}&
    \addFig{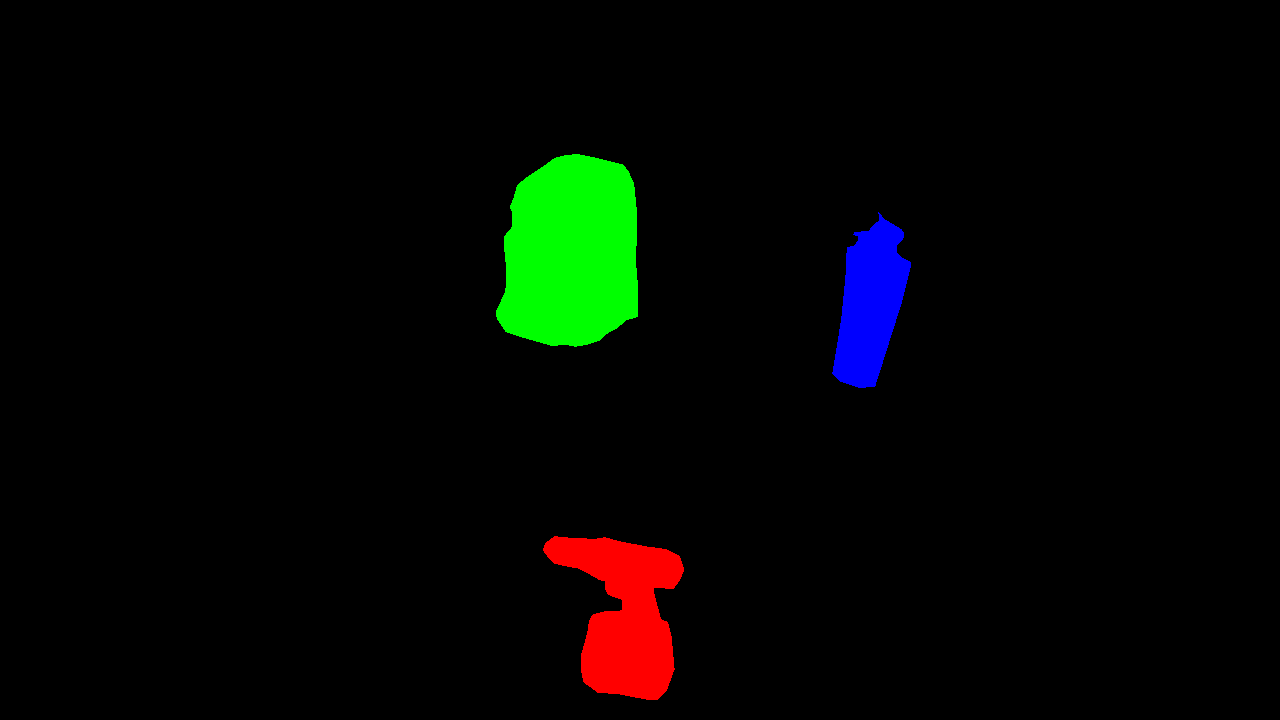}&   \addFig{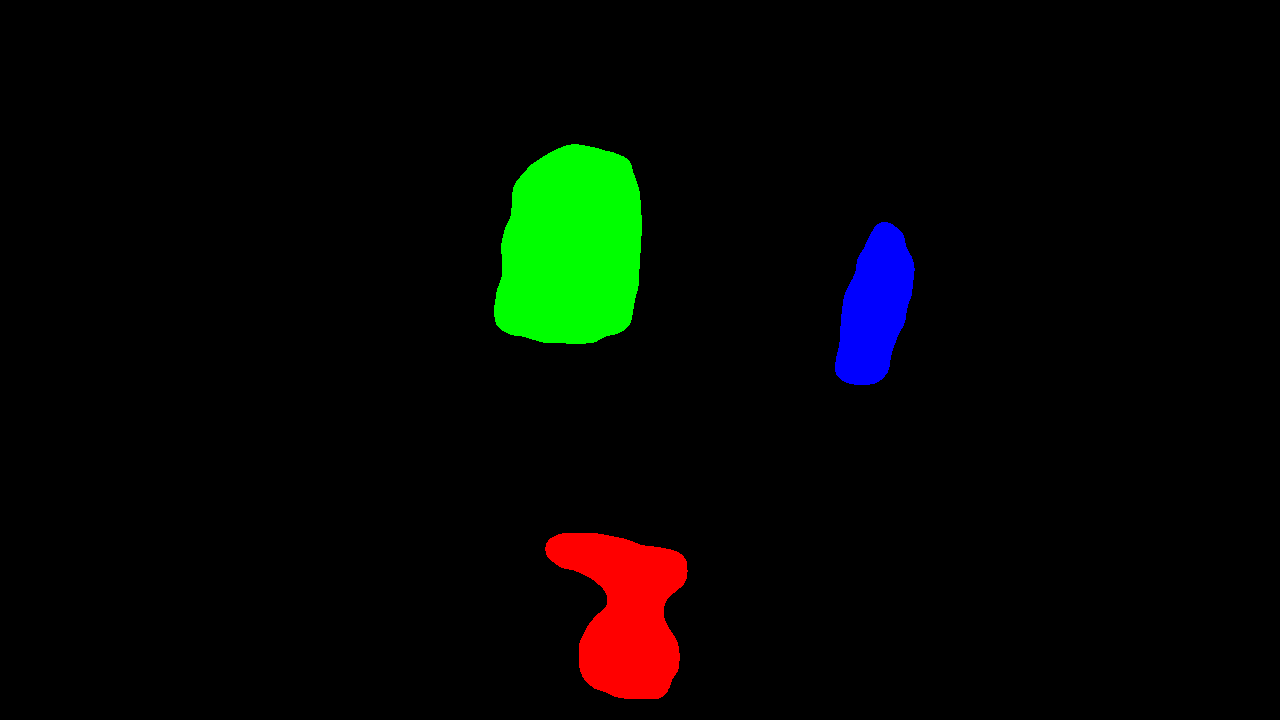}&   \addFig{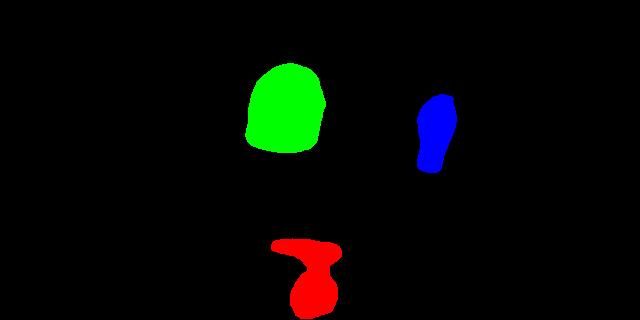}&
   \addFig{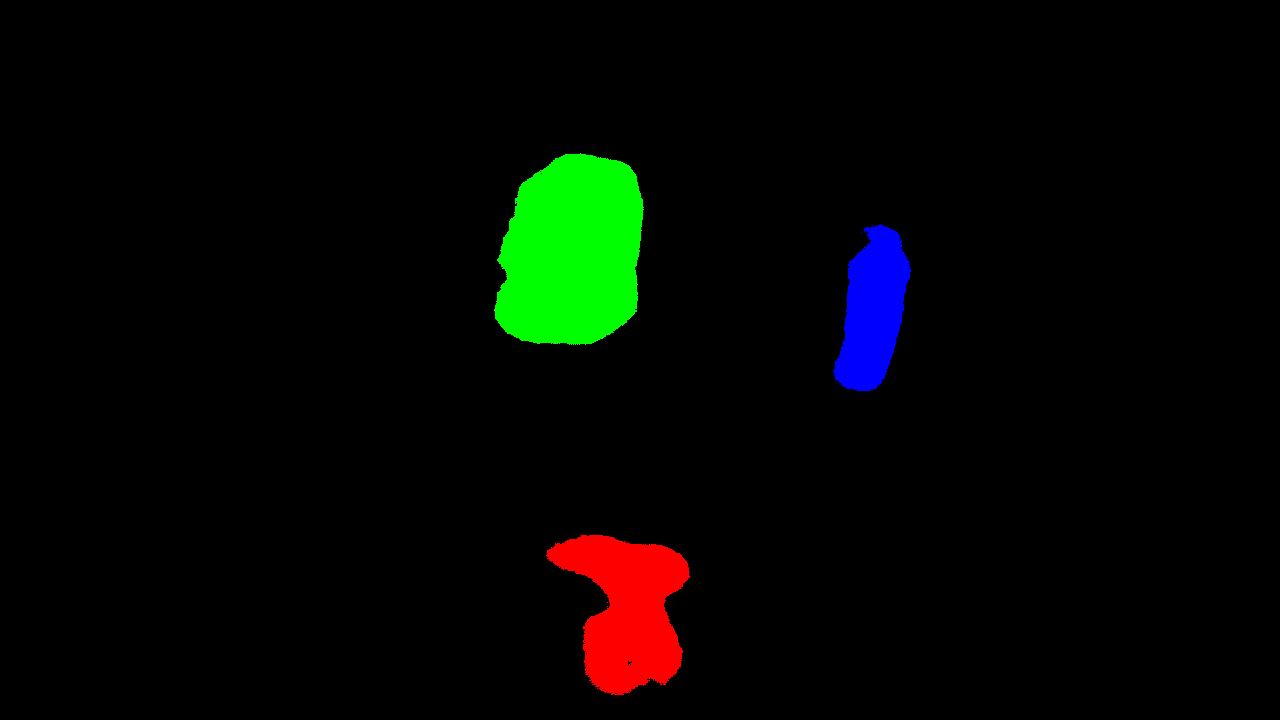}&   
    \addFig{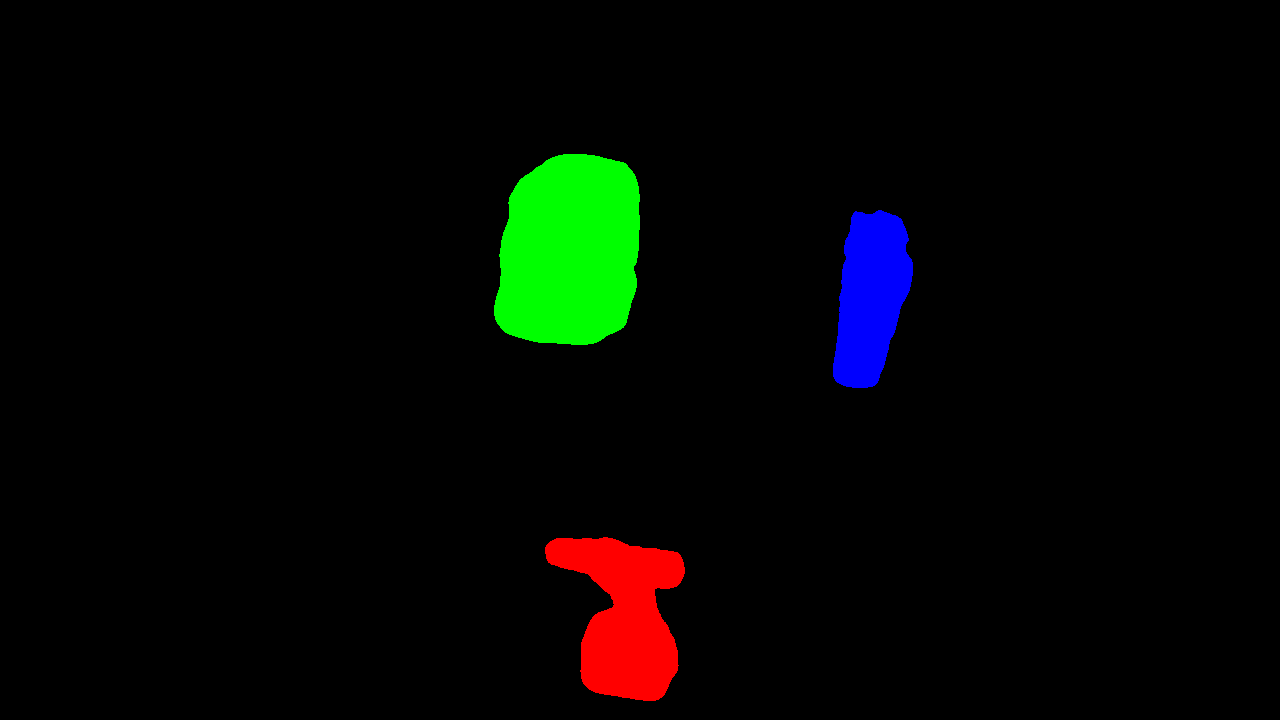}
   \\
   \addFig{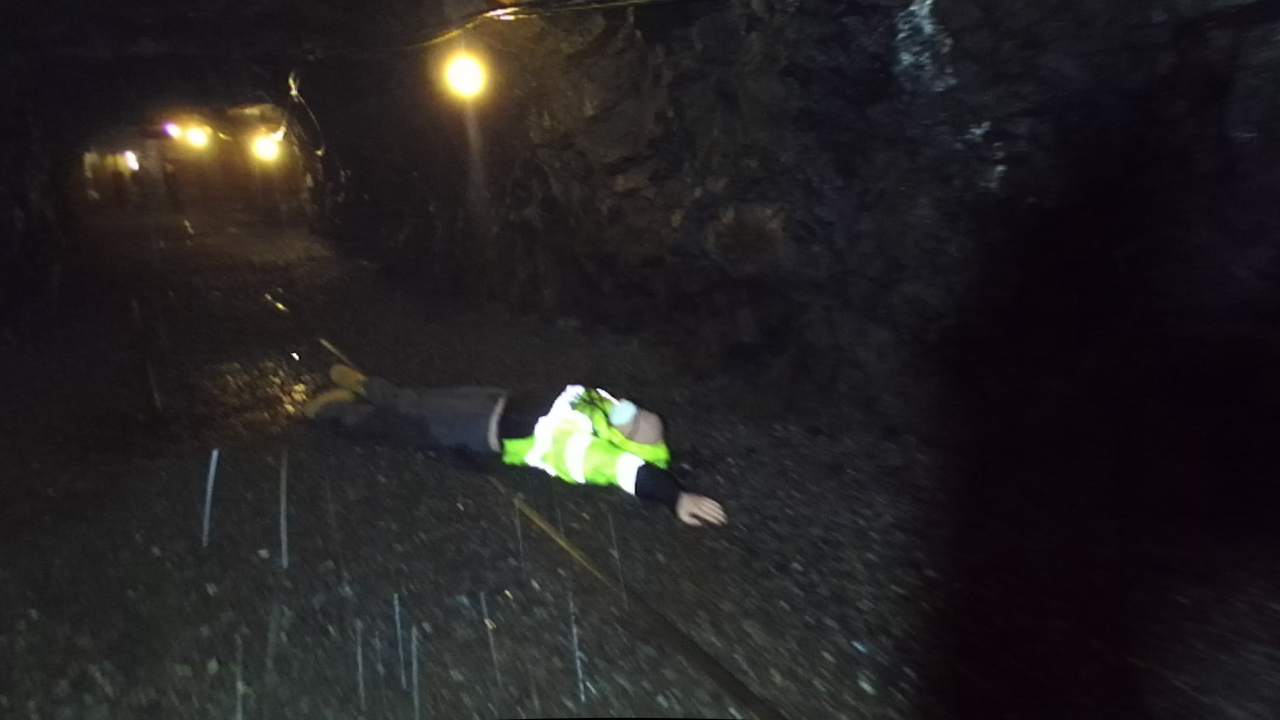}&
    \addFig{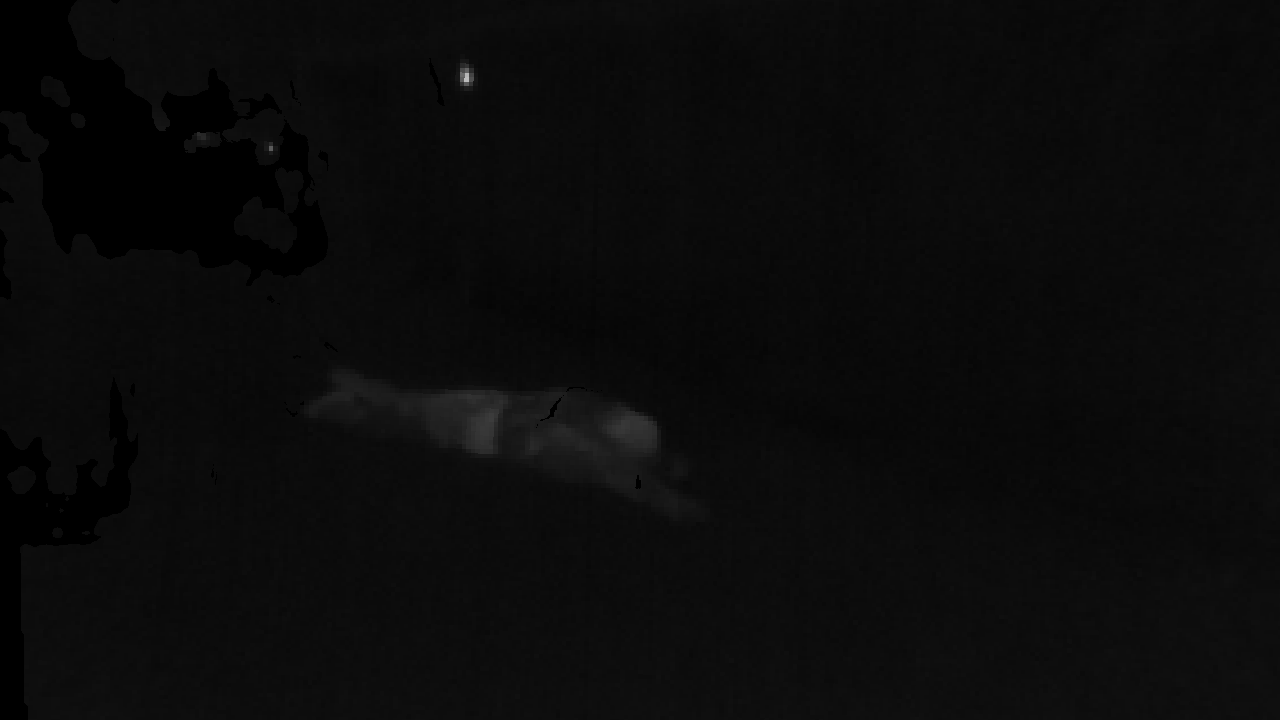}&
    \addFig{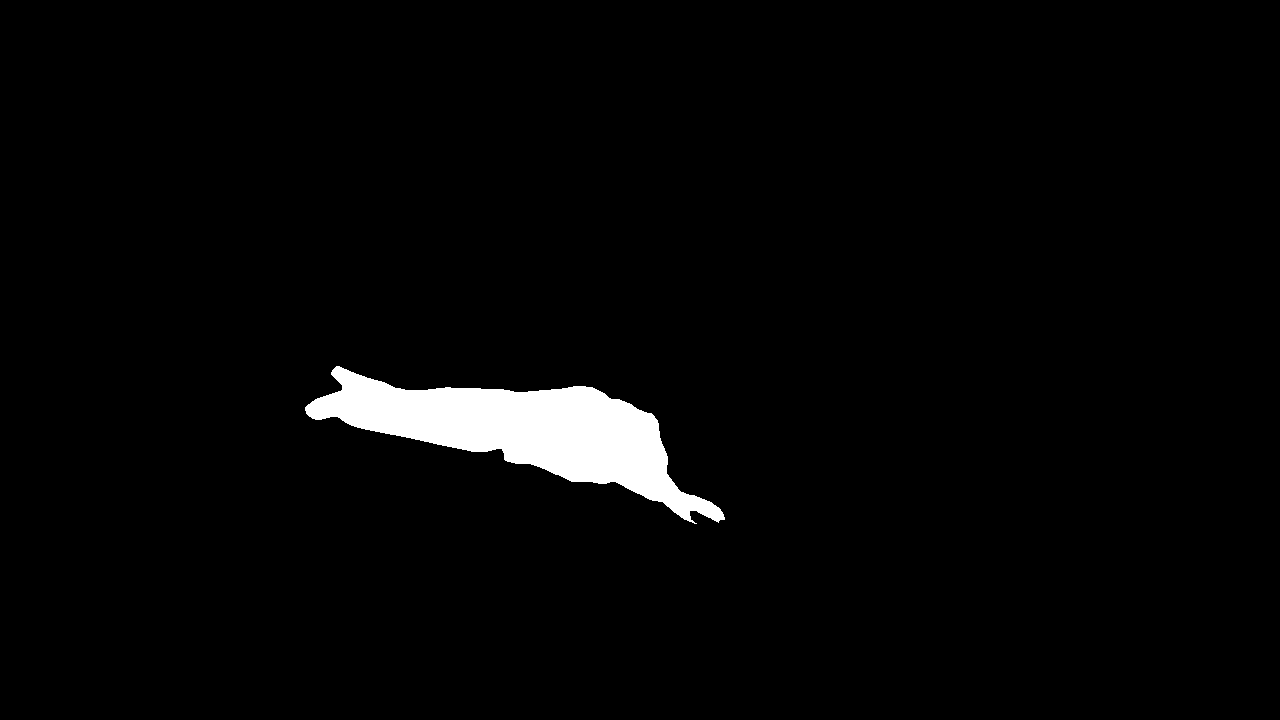}&   \addFig{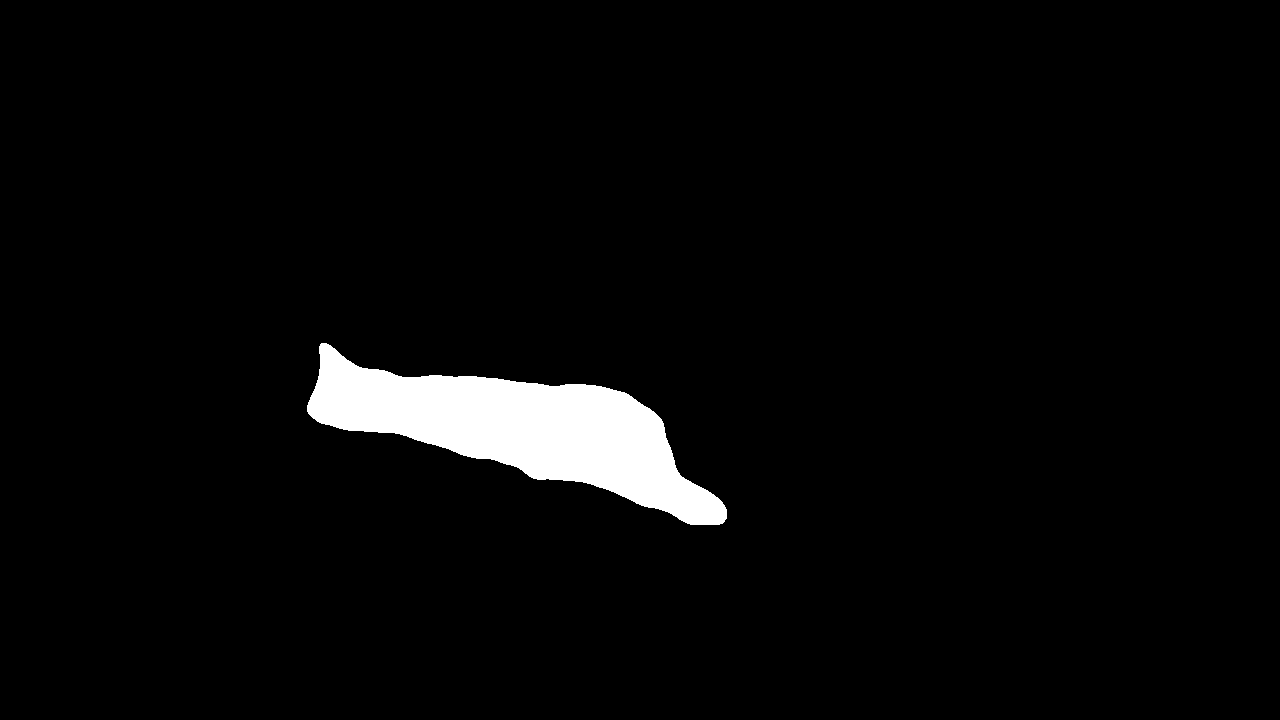}&
    \addFig{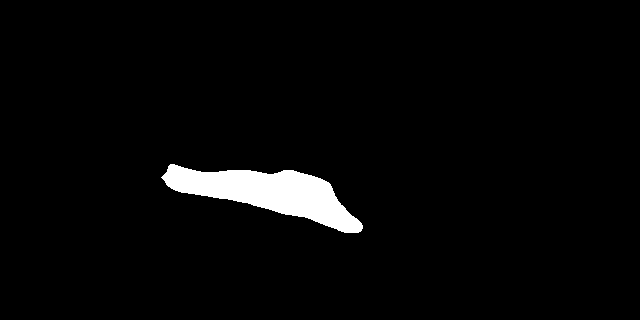}&
   \addFig{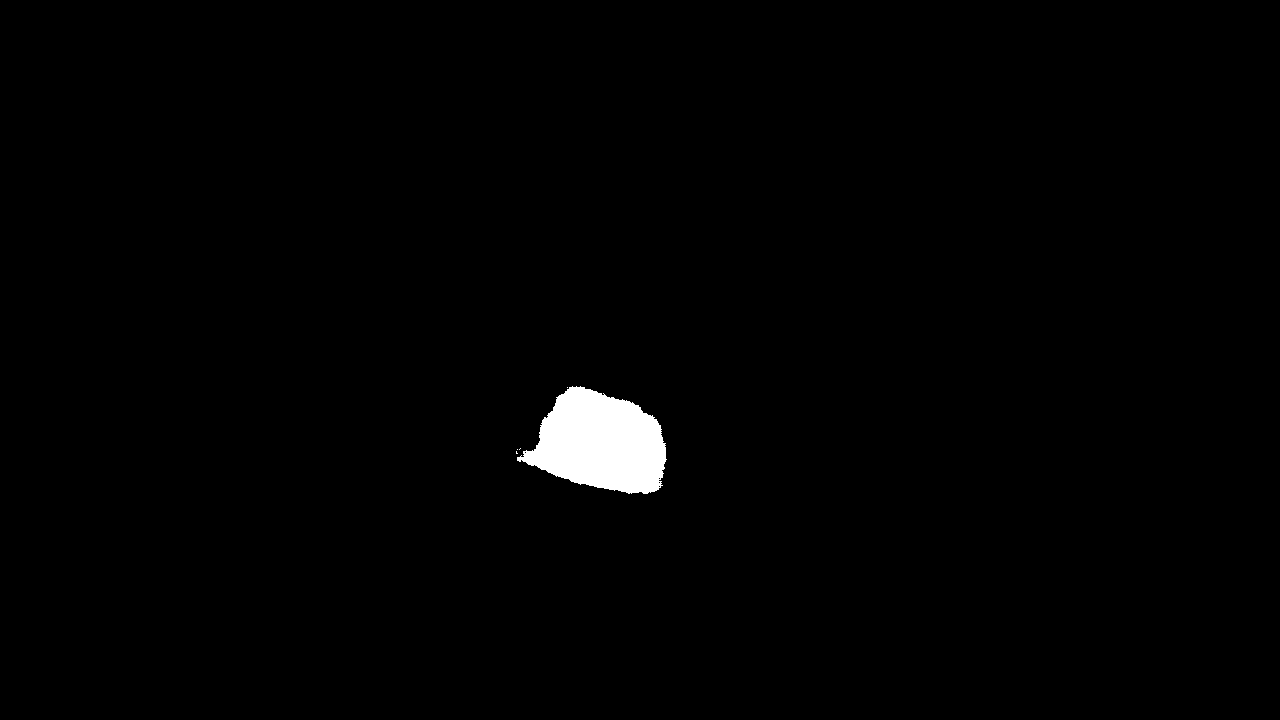}&   
    \addFig{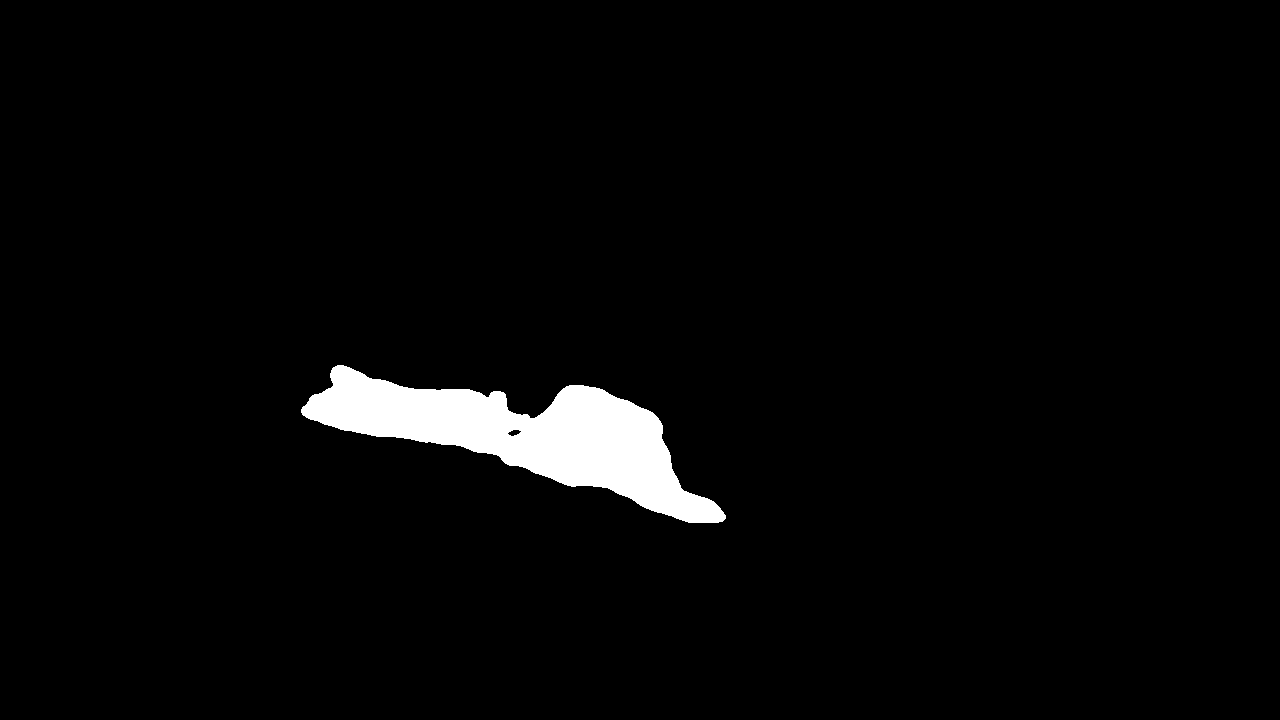}
   \\
   \addFig{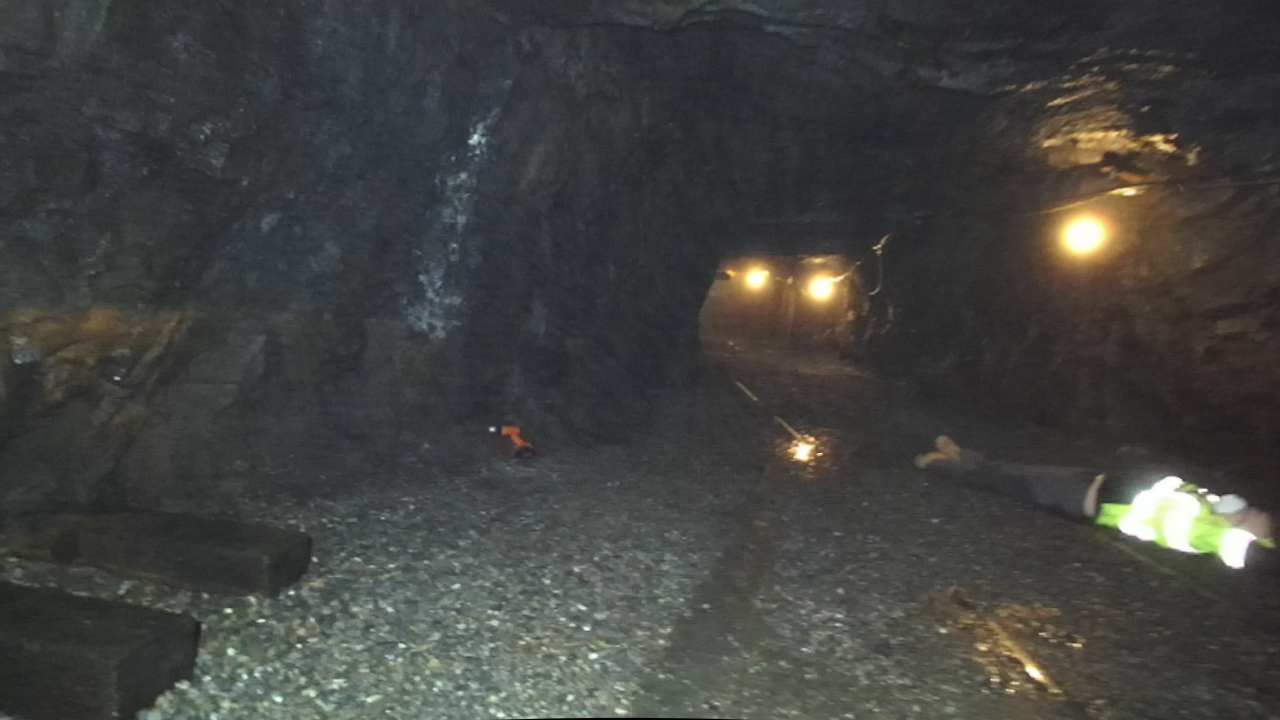}&
    \addFig{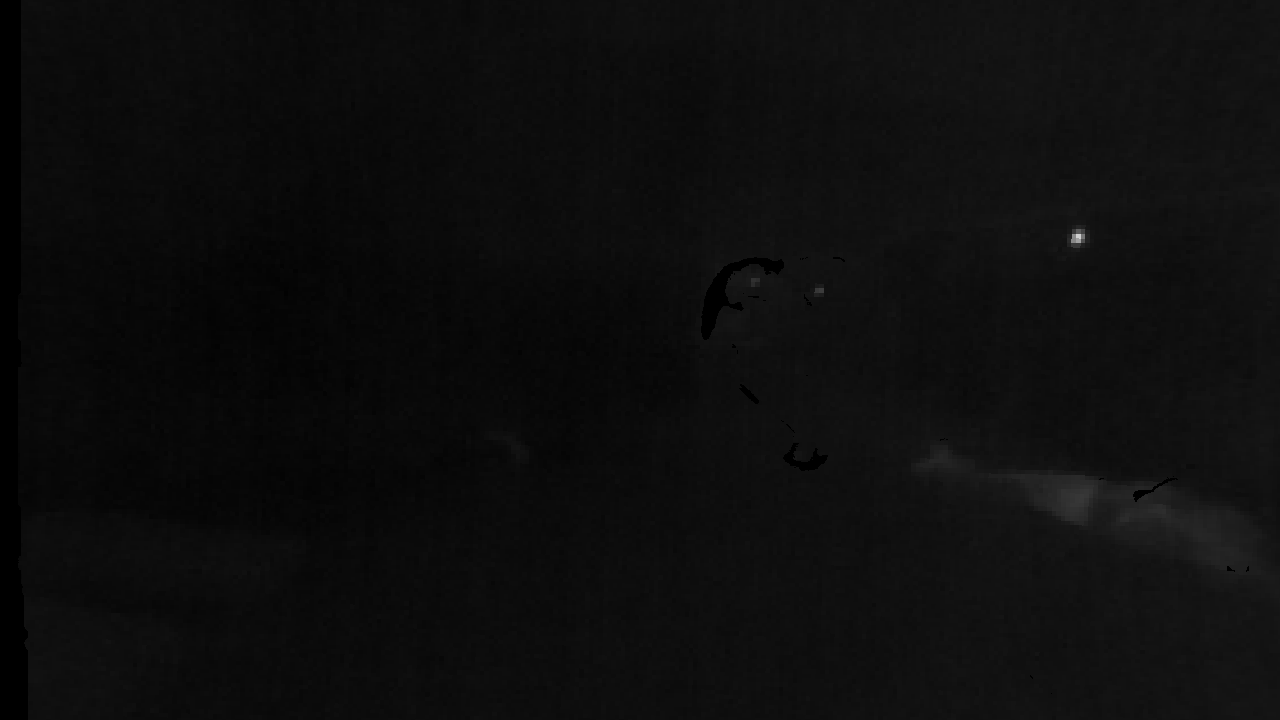}&
    \addFig{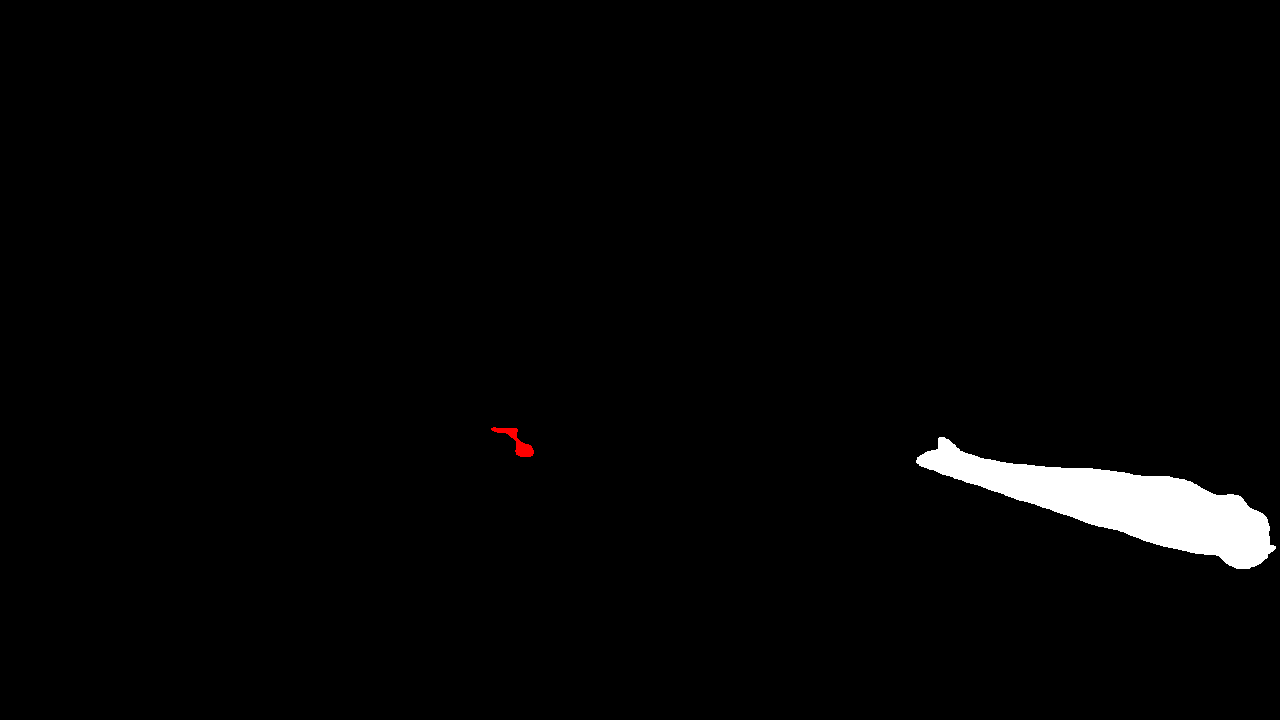}&   \addFig{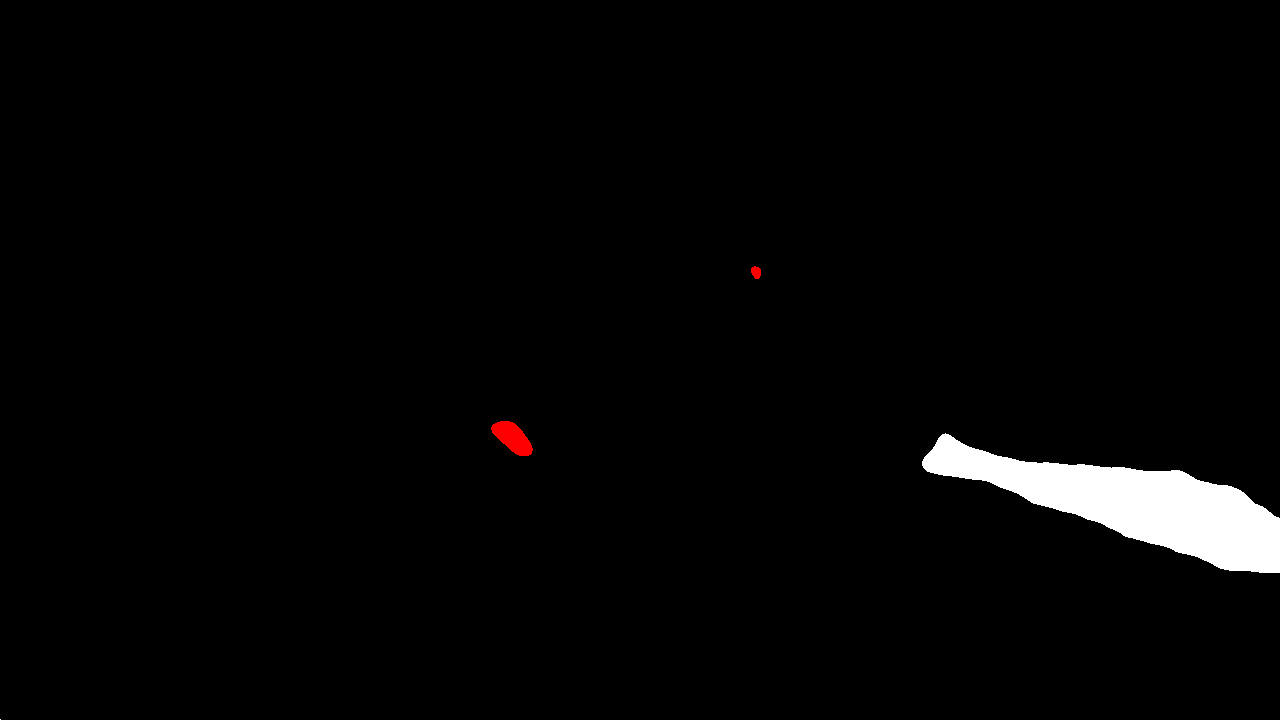}&
   \addFig{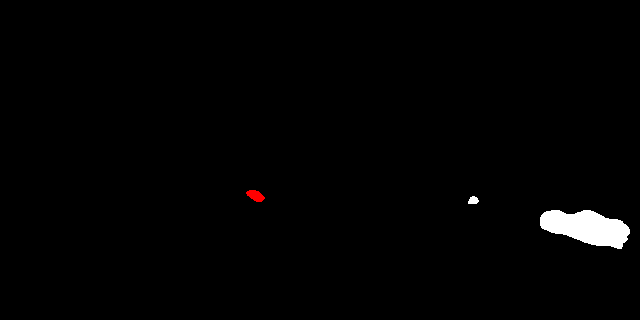}&
      \addFig{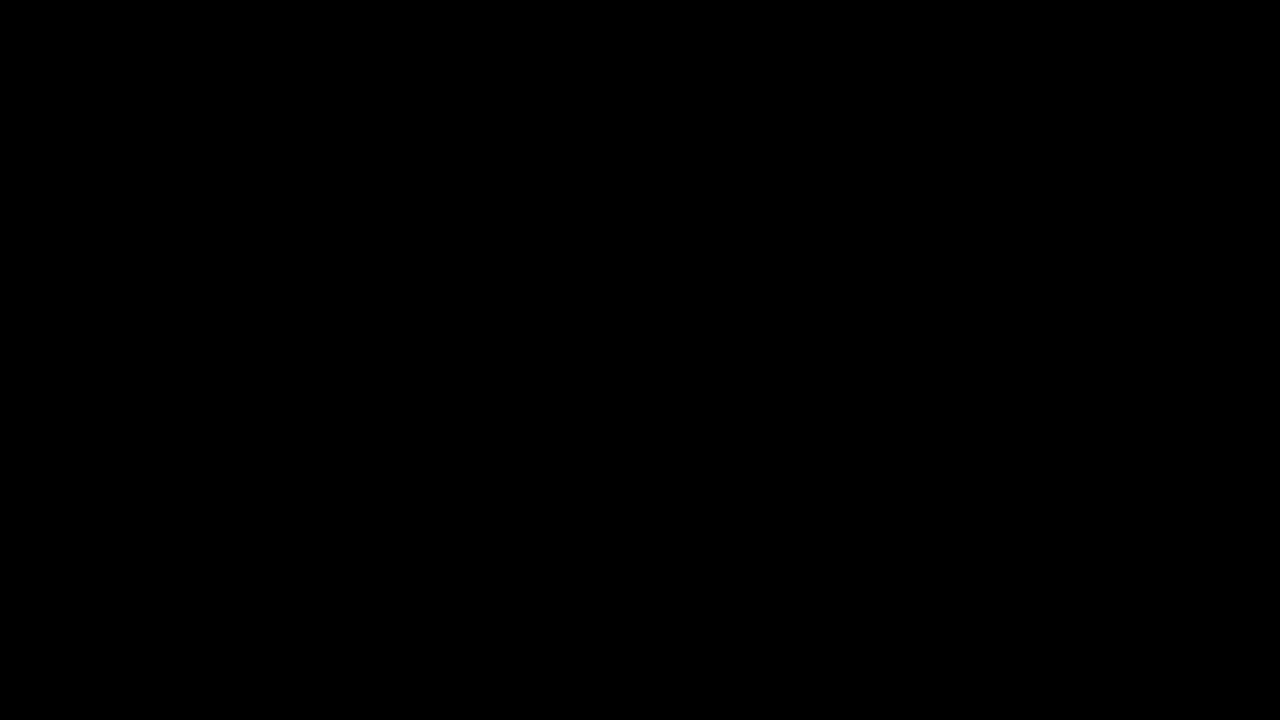}& 
    \addFig{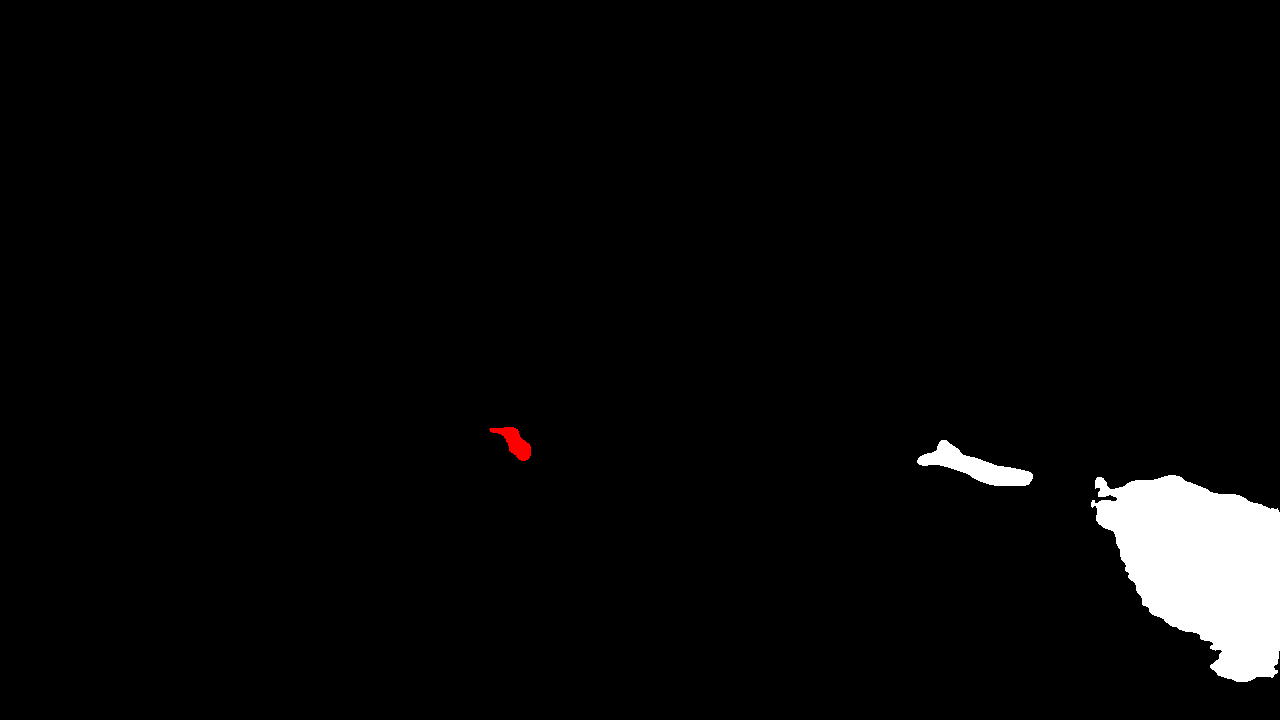}
   \\
   \addFig{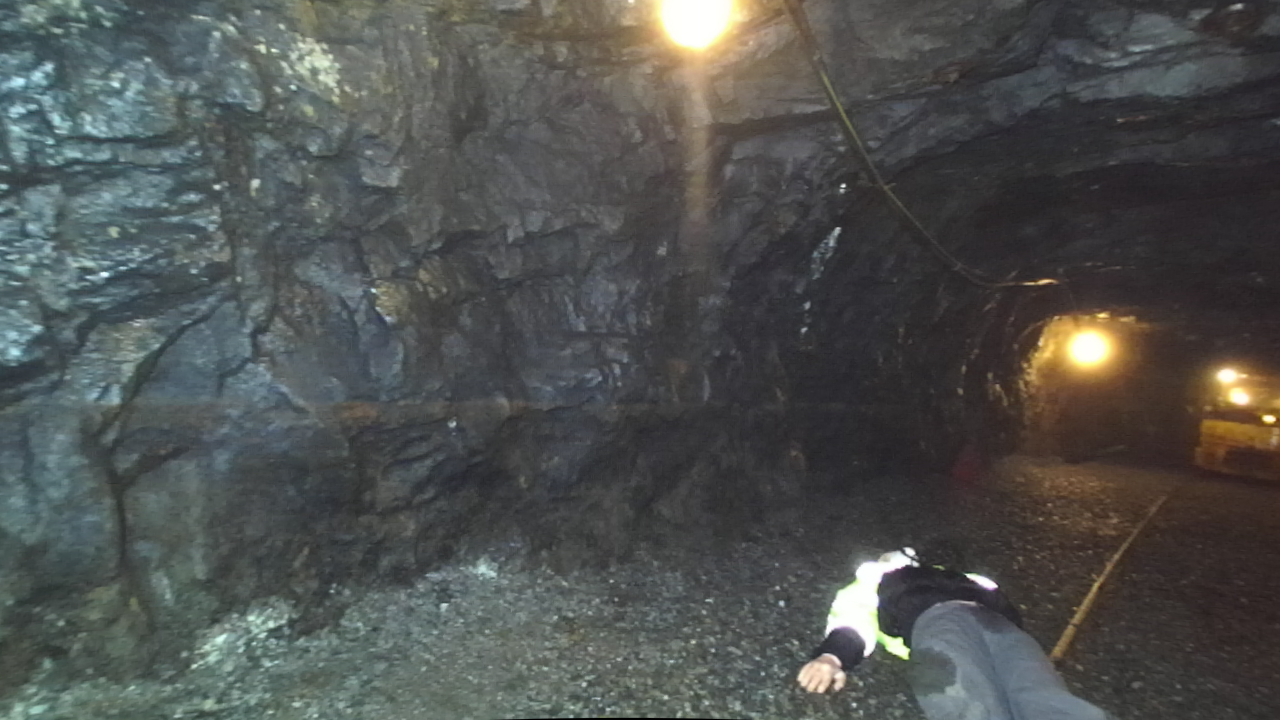}&
    \addFig{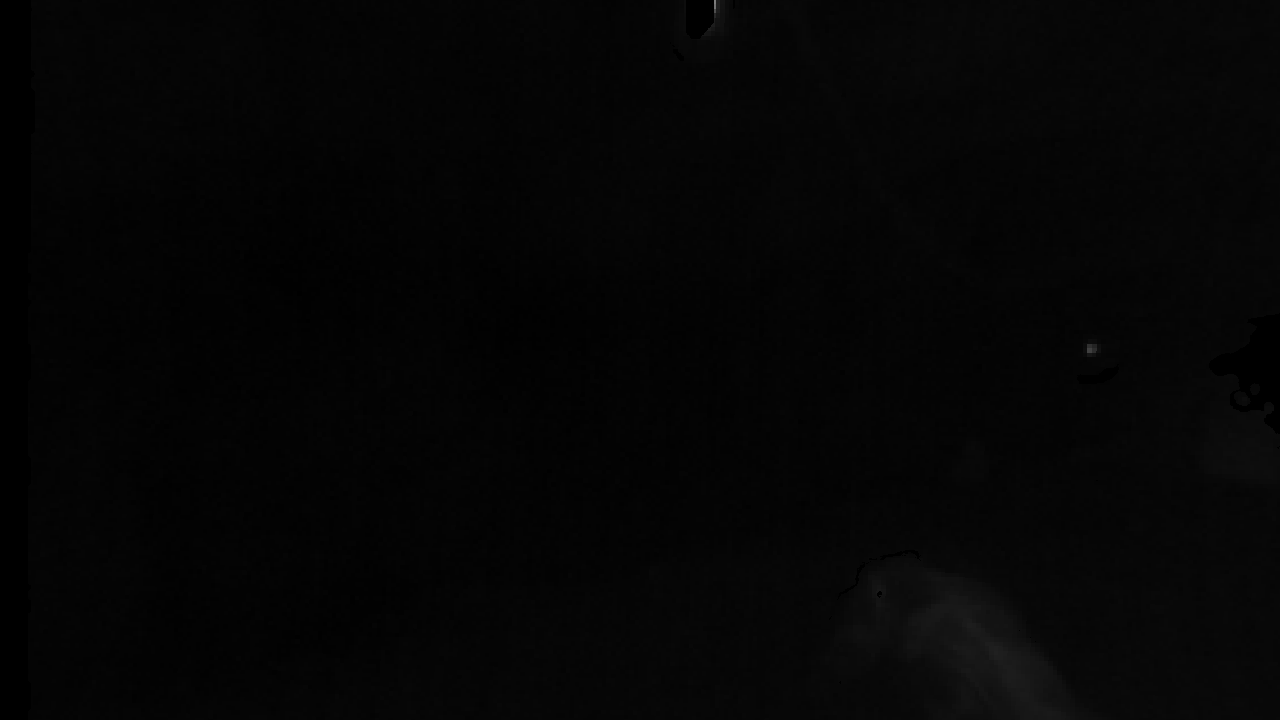}&
    \addFig{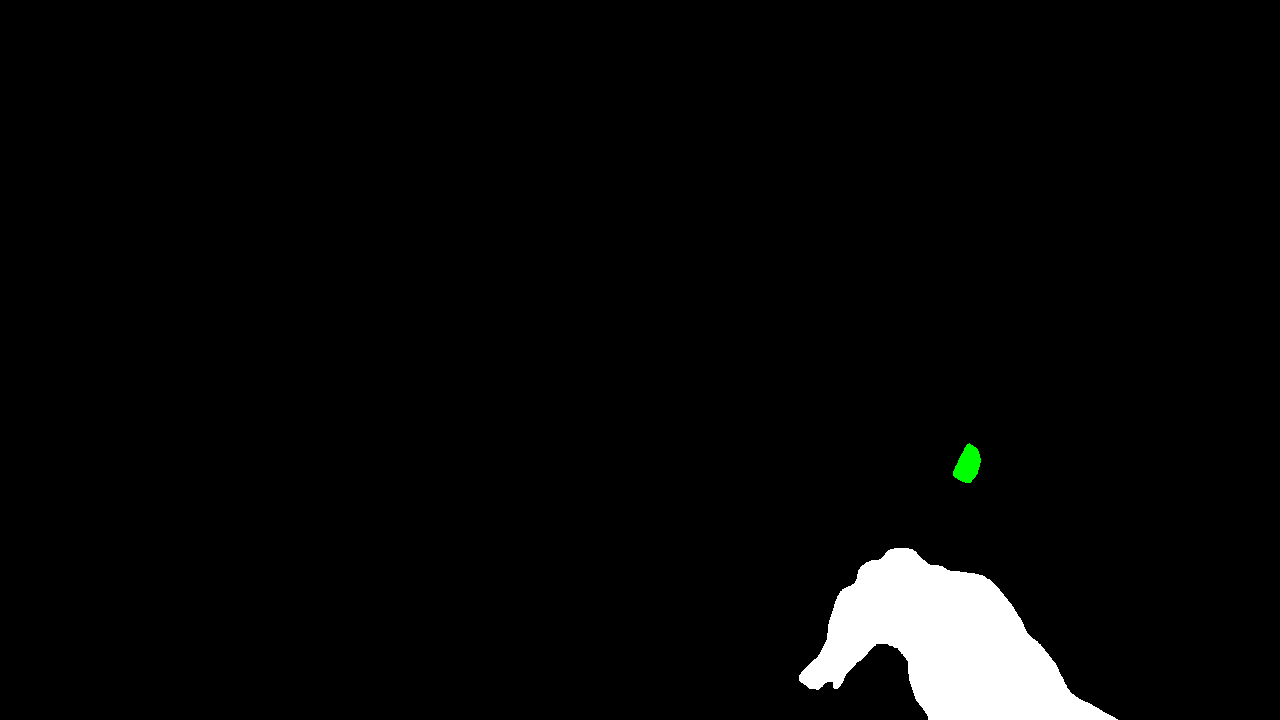}&   \addFig{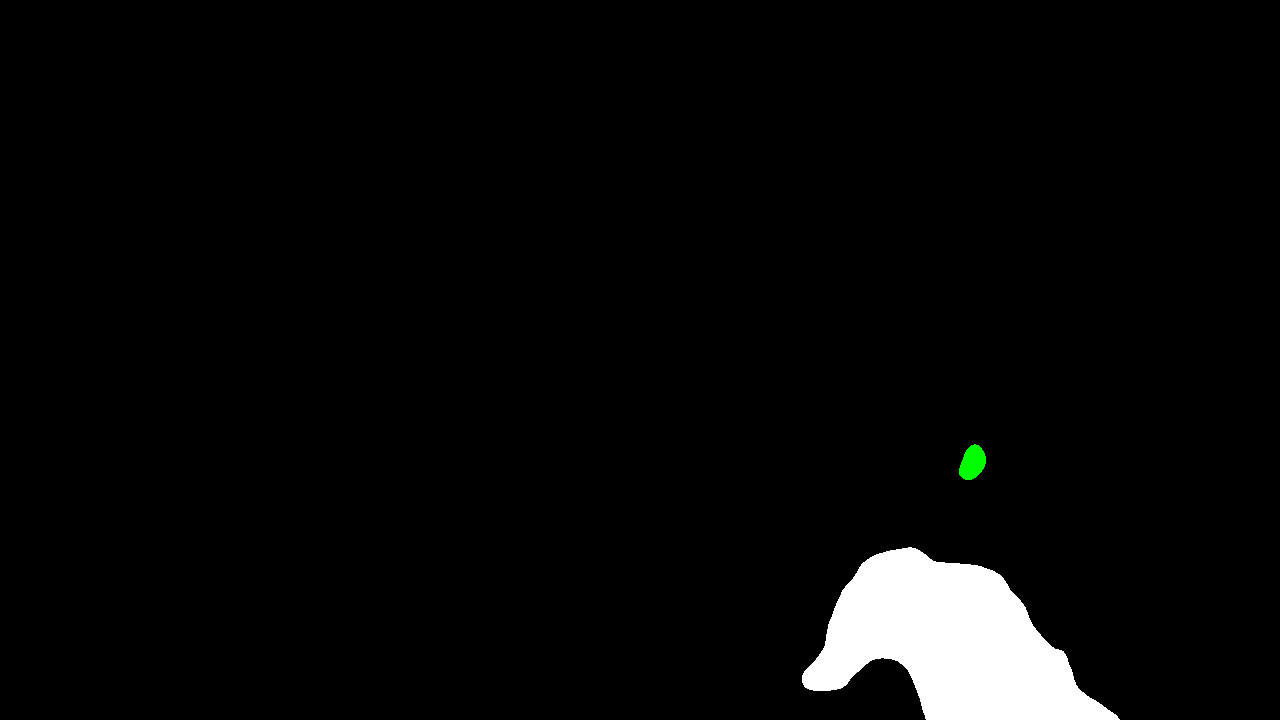}&
       \addFig{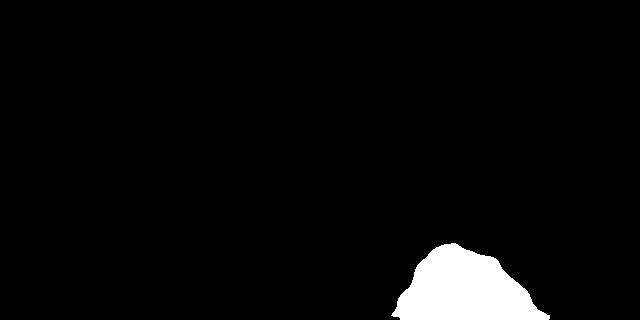}&
   \addFig{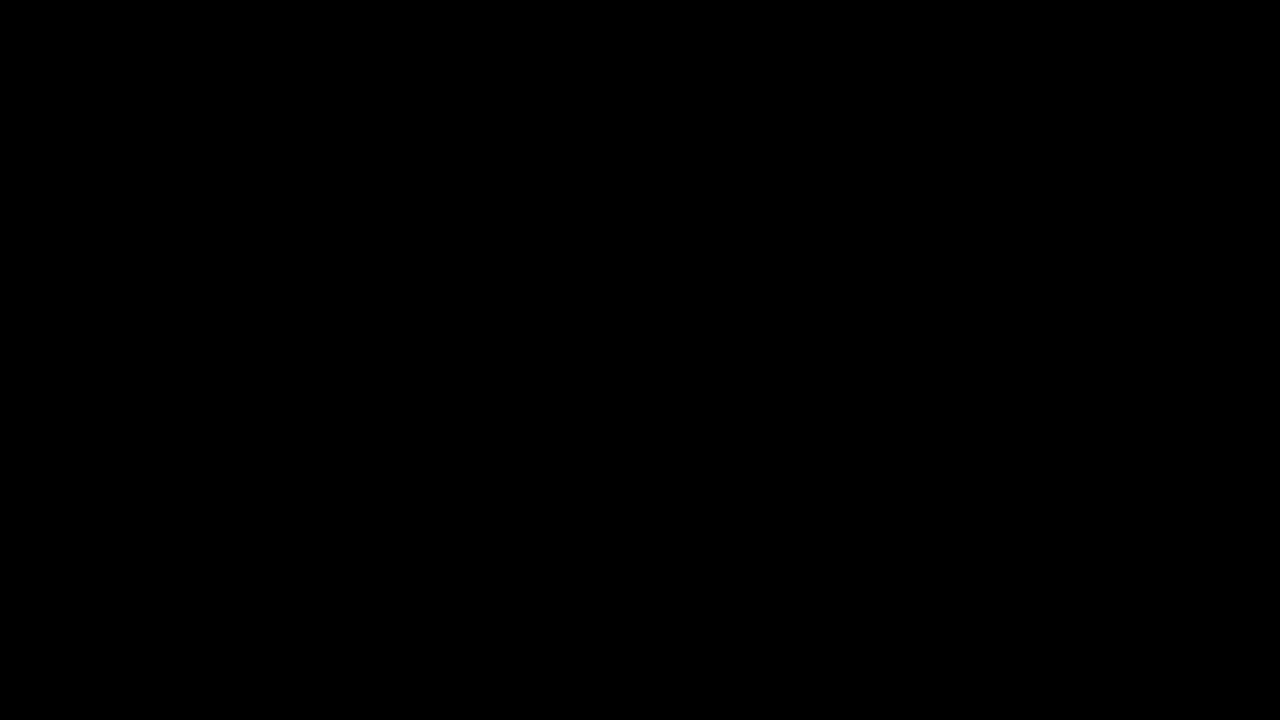}&
    \addFig{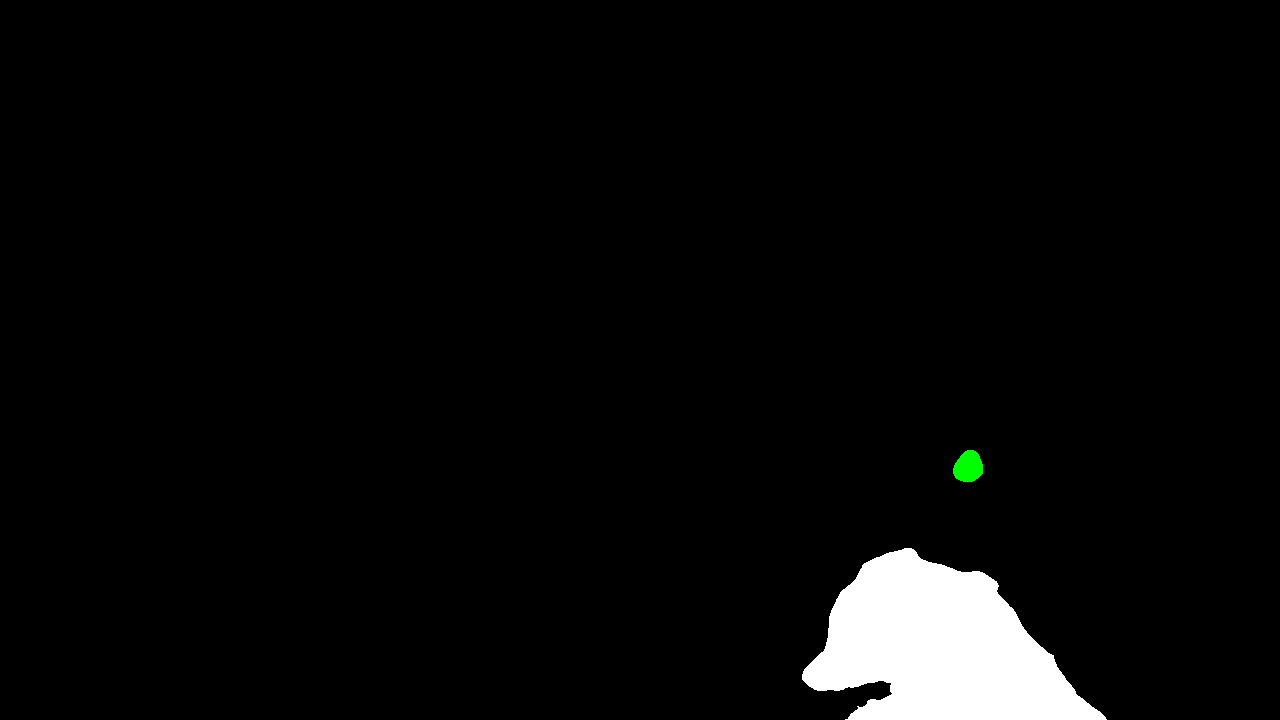}
   \\
   RGB & Thermal & GT & \textbf{Ours} &DSGBINet~\cite{2022DSGBINet}& 
   FDCNet~\cite{2022FDCNet}& 
   LASNet~\cite{2023LASNet} \\
  \end{tabular}
  \caption{Qualitative comparison for semantic segmentation of  RGB-T images on PST900~\cite{2020PSTNet} dataset.}
  \label{fig:com_pst900}
\end{figure*}

\textit{2) Evaluation on the PST900 Dataset:}
We also compare the proposed CAINet on the PST900 dataset with ten state-of-the-art methods. They are all RGB-T semantic segmentation methods, including MFNet~\cite{2017MFNet}, PSTNet~\cite{2020PSTNet}, MFFENet~\cite{2022MFFENet}, EGFNet~\cite{2022EGFNet}, MTANet~\cite{2022MTANet},
GMNet~\cite{2021GMNet},
CCFFNet~\cite{2022CCFFNet},
DSGBINet~\cite{2022DSGBINet},
FDCNet~\cite{2022FDCNet},
and LASNet~\cite{2023LASNet}.
The quantitative performance of our proposed method and the compared methods on the PST900 dataset is reported in Tab.~\ref{table:QuantitativeResults_PST900}. Our method demonstrates competitive performance on the PST900 dataset, securing six first-place and two second-place rankings across all 12 metrics. Of particular significance, our method outperforms the second-place LASNet by 0.34\% in mIoU, ranking first in this metric. Additionally, our method achieves first place in mAcc, surpassing EGFNet by 0.25\%. It is worth noting that our method exhibits significant performance improvements in the ‘Backpack’ and ‘Survivor’ categories. Our analysis underscores the effectiveness of CAINet on the PST900 dataset and its potential for generalization to different datasets. Furthermore, compared to previous methods, our approach provides accurate and reliable segmentation results, as visualized in Fig.~\ref{fig:com_pst900}.

\textit{3) Computational Complexity:}
We present a comparison of model parameters and floating-point operations (FLOPs) in Tab.~\ref{table:flops_and_params}. Compared to other models, our model achieves the highest mIoU and comparable mAcc while utilizing only 12.16M parameters and 123.62G FLOPs, showcasing the highest computational efficiency.
%
%
%

\begin{table}[!t]
\centering
\small
\caption{Quantitative results (\%) of assessing the individual and joint contributions of the four modules in CAINet on MFNet~\cite{2017MFNet} dataset.The best one is \textcolor{red}{\textbf{red}}.
  }
\label{Ablation_model}
\renewcommand{\arraystretch}{1.45}
\renewcommand{\tabcolsep}{1.6mm}
\begin{tabular}{c|ccccc||cc}
\bottomrule

 {No.} & {Baseline} & {ARLM} &   {DA}   &   {CACR}   &   {GCM}    & {mAcc}& {mIoU}  \\
\hline
\hline
1 & \Checkmark &            &            &            &            &67.2 &51.9   \\ 
2 & \Checkmark & \Checkmark &            &            &            & 72.0&53.0   \\
3 & \Checkmark & \Checkmark & \Checkmark &            &            & 64.8&54.9   \\
4 & \Checkmark & \Checkmark &            & \Checkmark &            & 70.7&55.3   \\
5 & \Checkmark & \Checkmark &            &            & \Checkmark & 69.3&55.7   \\
\hline
6 & \Checkmark & \Checkmark &            & \Checkmark & \Checkmark & \textcolor{red}{\textbf{73.8}}&56.4   \\
7 & \Checkmark & \Checkmark & \Checkmark &            & \Checkmark & 69.2&56.6   \\
8 & \Checkmark & \Checkmark & \Checkmark & \Checkmark &            & 63.3&55.7   \\
\hline
9 & \Checkmark & \Checkmark & \Checkmark & \Checkmark & \Checkmark & 73.2&\textcolor{red}{\textbf{58.6}}  \\
\toprule
\end{tabular}
\end{table}

\begin{table}[!t]
\centering
\small
\caption{Quantitative results (\%) of assessing the individual and joint contributions of the five kinds of supervision in CAINet on MFNet~\cite{2017MFNet} dataset.The best one is \textcolor{red}{\textbf{red}}.
  }
\label{Ablation_supervision}
\renewcommand{\arraystretch}{1.45}
\renewcommand{\tabcolsep}{1.0mm}
\begin{tabular}{c|ccccc||cc}
\bottomrule

 {No.} & {$L_{decoder}$} & {$L_{target}$} & {$L_{Att}$} & {$L_{binary}$} & {$L_{boundary}$} & mAcc& mIoU \\
\hline
\hline
1 & \Checkmark &            &            &            &            &70.5& 54.3   \\ 
2 & \Checkmark & \Checkmark &            &            &            &60.4& 54.9   \\
3 & \Checkmark & \Checkmark & \Checkmark &            &            & 70.6&55.7   \\
4 & \Checkmark & \Checkmark &            & \Checkmark &            & 69.0&55.1   \\
5 & \Checkmark & \Checkmark &            &            & \Checkmark & 70.5&54.7   \\
\hline
6 & \Checkmark & \Checkmark &            & \Checkmark & \Checkmark & 71.2&55.9   \\
7 & \Checkmark & \Checkmark & \Checkmark &            & \Checkmark & \textcolor{red}{\textbf{74.8}}&56.7   \\
8 & \Checkmark & \Checkmark & \Checkmark & \Checkmark &            & 70.4&56.9   \\
\hline
9 & \Checkmark & \Checkmark & \Checkmark & \Checkmark & \Checkmark &73.2& \textcolor{red}{\textbf{58.6}}  \\
\toprule
\end{tabular}
\end{table}
%
%
\subsection{Ablation Studies}
\label{Ablation Studies}
To explore the effectiveness of our proposed modules on RGB-T semantic segmentation, we perform ablation studies on the MFNet dataset using the same hyperparameters as in Section~\ref{sec:ExpProtocol}. Specifically, we evaluate the effectiveness of the proposed four modules of ARLM, DA, CACR and GCM with individual and joint contributions, as well as multiple auxiliary supervision and target supervision.

\textit{1) The Individual and Joint Contributions of Four Modules:}
We delete and replace the above mentioned modules, and provide five variants to evaluate the individual contributions of the four modules. 1) Baseline, 2) Baseline+ARLM, 
3) Baseline+ARLM+DA, 
4) Baseline+ARLM+CACR, and 
5) Baseline+ARLM+GCM. 
For “Baseline”, we retain three supervised decoders where RGB and thermal image features were fused at the end using the ASPP module~\cite{2018DenseASPP}. 
Among them, ARLM is special in the framework, and the other modules must be combined with ARLM to function, so ARLM cannot be removed. For convenience, the deleted modules are replaced with ASPP modules. 
Quantitative results are shown in Tab.~\ref{Ablation_model}, where ``\Checkmark"  indicates that the corresponding module is retained, and no mark represents that the corresponding module is removed.
“Baseline” only achieves 51.9\% in mIoU, which is 6.7\% lower than our vanilla CAINet, indicating that these three modules can also improve the accuracy of segmentation when acting together. 
With the help of DA, CACR or GCM in conjunction with ARLM, the No.2, No.3, No.4 and No.5 variants improve the performance considerably compared to “Baseline”. 

In addition, we provide three variants to evaluate the joint contribution of the three modules, as shown in Tab.~\ref{Ablation_model}: 6) Baseline+ARLM+CACR+GCM, 7) Baseline+ARLM+DA+GCM, and 8) Baseline+ARLM+DA+CACR. 
With the cooperation of the modules, the performance of the above three variants is further improved compared to the single module. We find that with the fusion of the CACR and GCM modules, the results are higher than the results of the other two joint modules, which confirms the boost of the RGB-T semantic segmentation performance by explicit auxiliary supervision with attention maps. The perfect cooperation of the three modules results in an excellent full CAINet, reaching 58.6\%.

\textit{2) The Effectiveness of Auxiliary Supervision and Target Supervision:} 
The multi-task supervision model is widely used in industry because it can improve the model performance and accelerate the convergence of model training without increasing the reasoning time and computational complexity of the model. 
To evaluate the effectiveness of the ARLM module in conjunction with DA, CACR, and GCM for explicit auxiliary supervision and target supervision, we provide eight combinations of supervision in Tab.~\ref{Ablation_supervision}, including 1) three supervisions of the decoder at the end of the model, 2) four target supervisions of residual learning branch for four semantic segmentation loss $L_{seg}$, 3) retaining supervision of two attention maps, 4) keeping supervision of only the binary map, 5) retaining supervision of only the boundary map, 6) joint supervision of the binary and the boundary map, 7) joint supervision of the attention and the boundary map, and 8) joint supervision of the attention map and the binary map. 
We observe that discarding one or all of the supervision is detrimental to the final segmentation performance. With the effective and explicit guidance of multi-auxiliary supervision, the role of each functional module can be fully exploited. The GCM interacts as a global context-guided feature of the CACR in an explicit paradigm and the DA can enhance the representation of target edges and contour surfaces, which helps to refine the CAINet final segmentation results. The joint supervision achieves our CAINet final segmentation results in up to 58.6\% outperforming other results.


\begin{table}[!t]
\begin{center}
  \centering
  \small
  \renewcommand{\arraystretch}{1.4}
  \renewcommand{\tabcolsep}{5.0mm}
  \caption{quantitative comparison result(\%) on the test set of the  NYU-depth V2 dataset. The best results are highlighted in boldface.}
  \label{table:rgbd}

\begin{tabular}{ c | c c }
\hline
Methods &mAcc & mIoU \\
\hline
\hline
 D-CNN~\cite{2018D-CNN}  &   56.3 & 43.9 \\
 3D Graph~\cite{3dgroup} &   55.7 & 43.1  \\
 LSTM~\cite{lstm-cf} &   60.0 & 45.9  \\
 RefineNet~\cite{2017RefineNet} &    58.9 & 46.5 \\
 RDFNet~\cite{2017RDFNet} &    62.8 & 50.1 \\
 RTFNet~\cite{2019RTFNet} & 64.8  & 49.1 \\
 SA-Gate~\cite{2020SA-Gate} & -  & 52.4 \\
 ECGFNet~\cite{zhou2023embedded} & 65.2   & 51.5 \\
 Ours & \textbf{65.9}  & \textbf{52.6} \\
\hline
\end{tabular}
\end{center}
\end{table}

\renewcommand{\addFig}[1]{{\includegraphics[height=0.0586\textwidth]{com_nyu/#1}}}
\begin{figure}[t]
  \renewcommand{\arraystretch}{0.5}
  \setlength{\tabcolsep}{0.3mm}
  \begin{tabular}{cccccc}
    \addFig{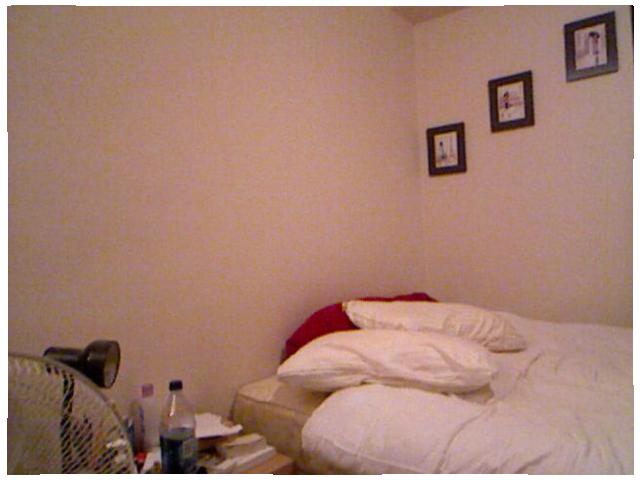} &
    \addFig{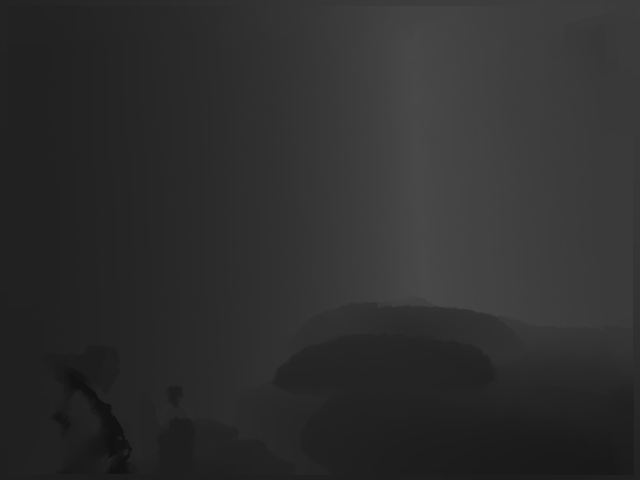} &
    \addFig{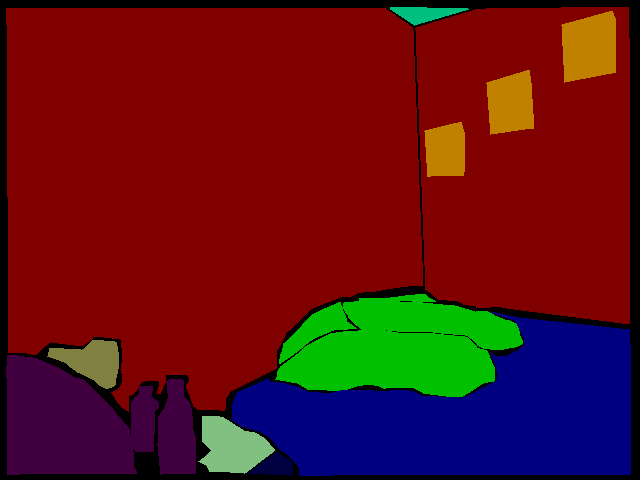} &
    \addFig{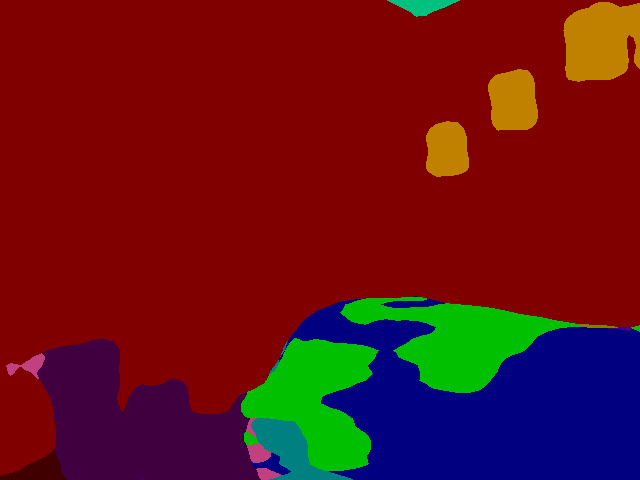} &
    \addFig{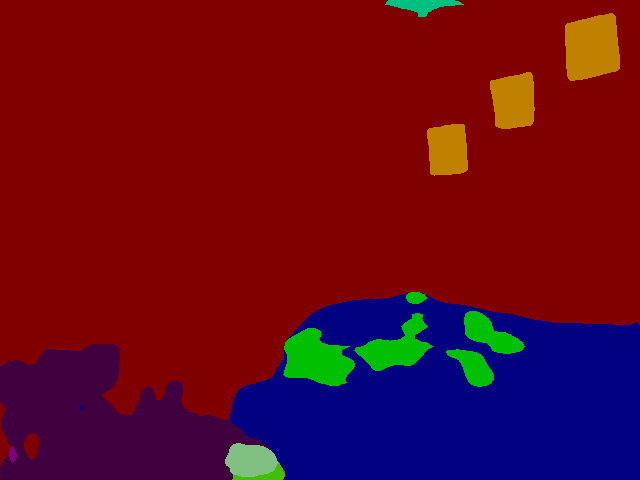} &
    \addFig{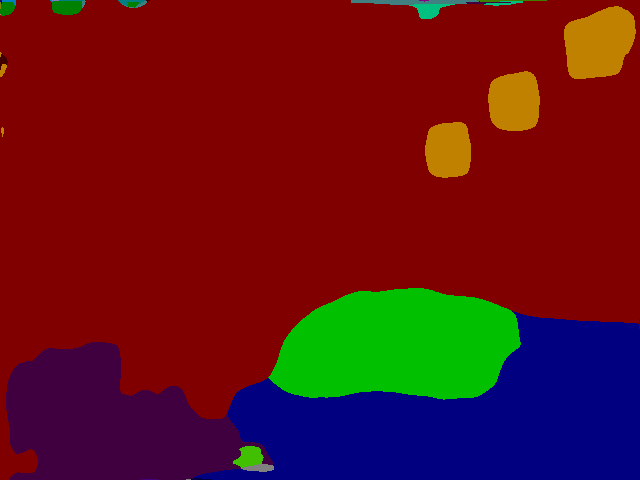} \\
    \addFig{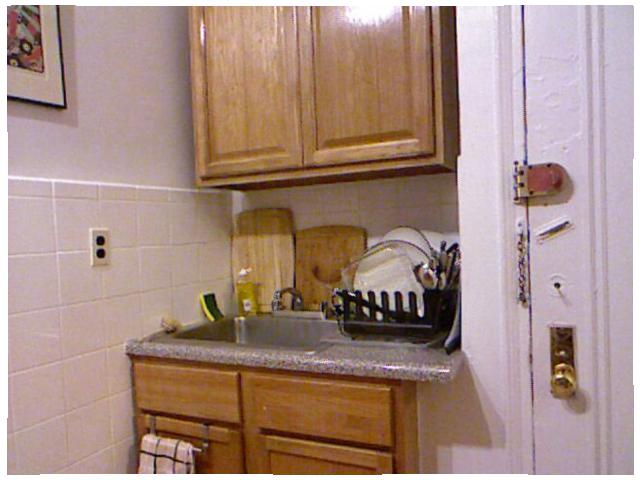} &
    \addFig{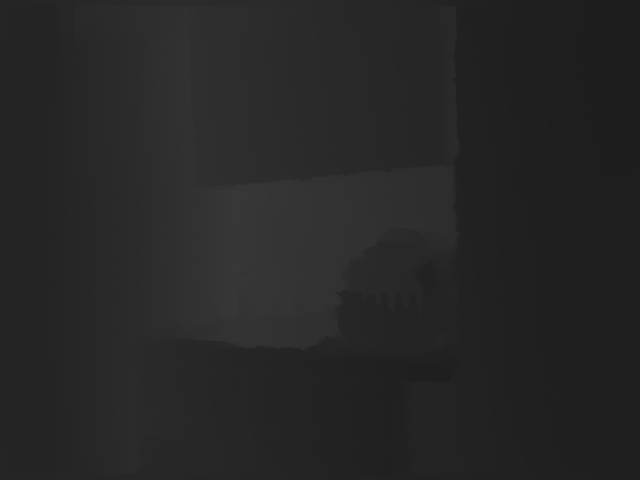} &
    \addFig{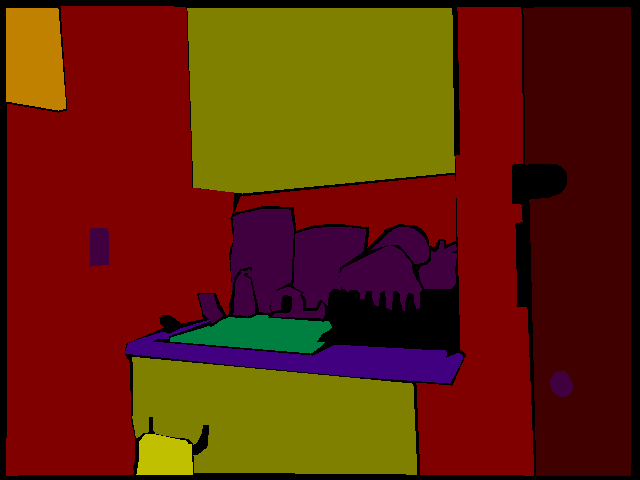} &
    \addFig{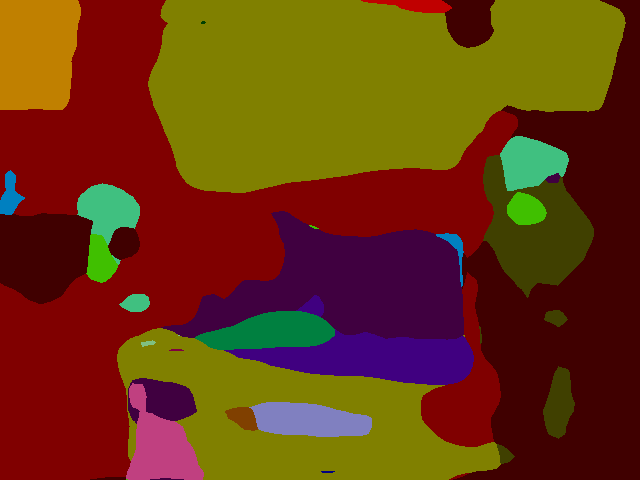} &
    \addFig{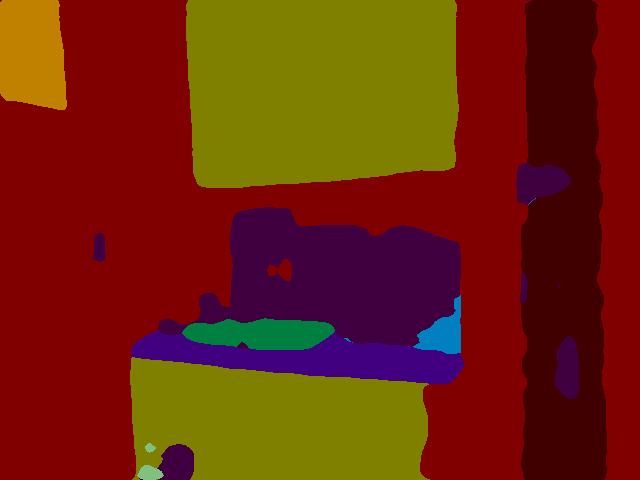} &
    \addFig{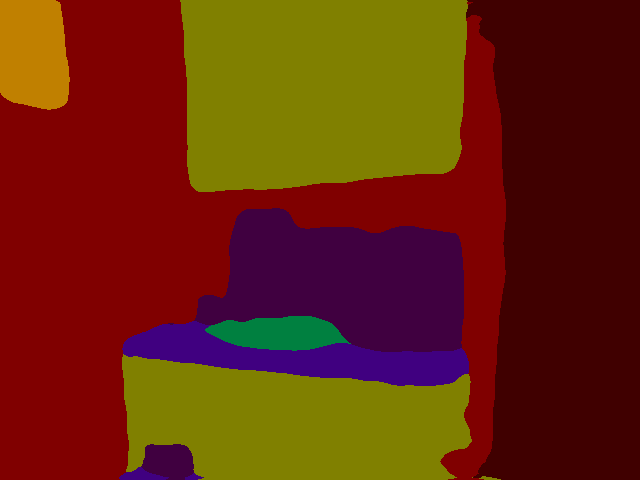} \\
    \addFig{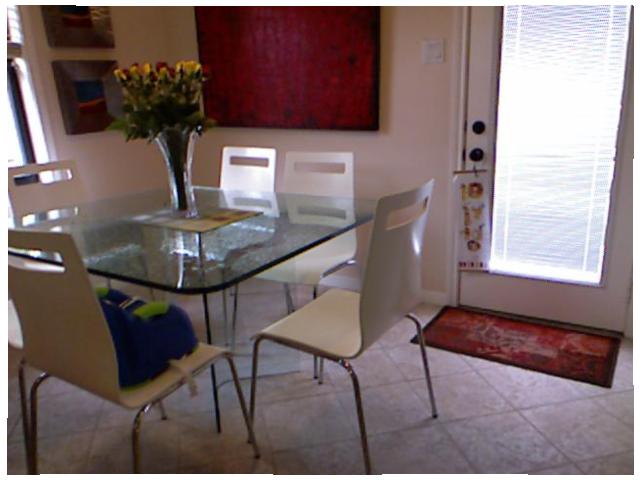} &
    \addFig{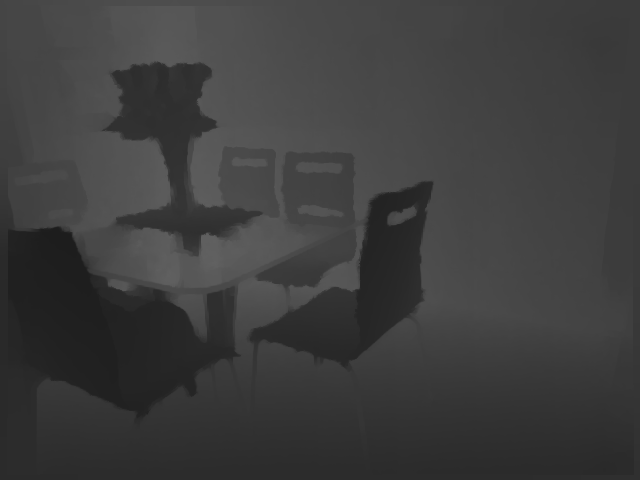} &
    \addFig{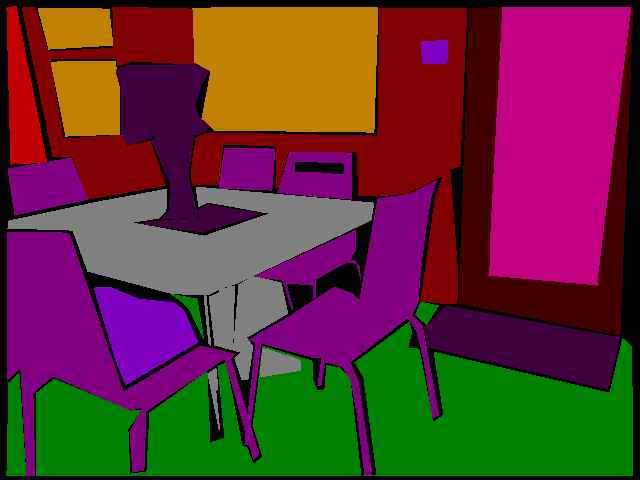} &
    \addFig{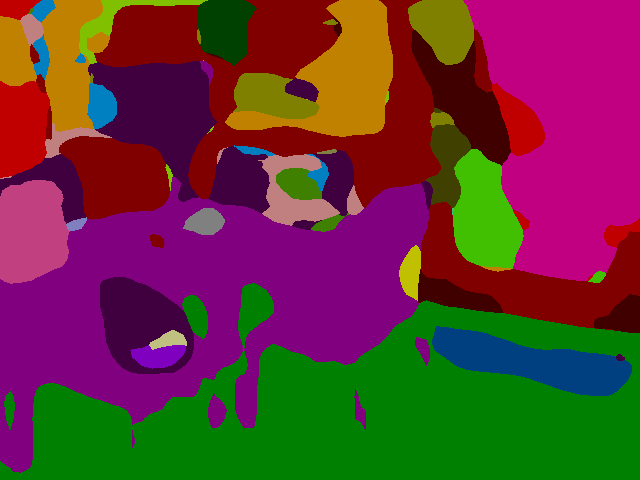} &
    \addFig{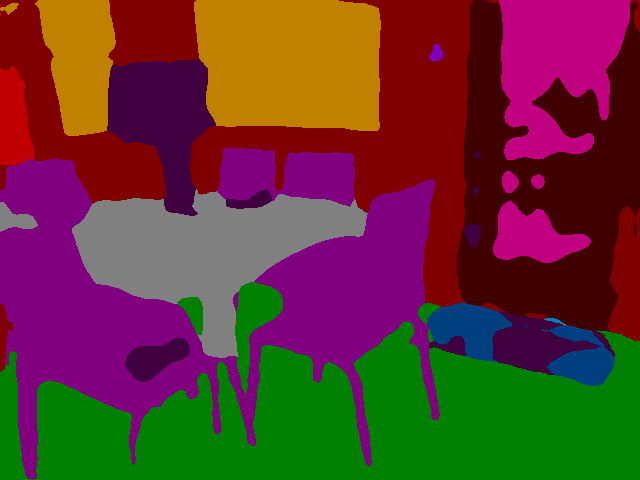} &
    \addFig{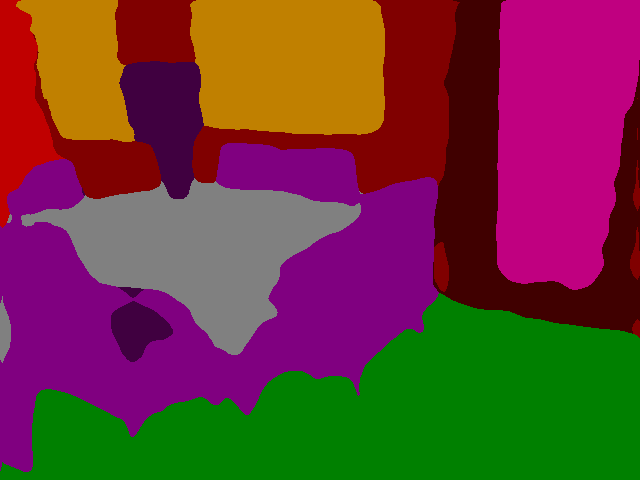} \\
    \addFig{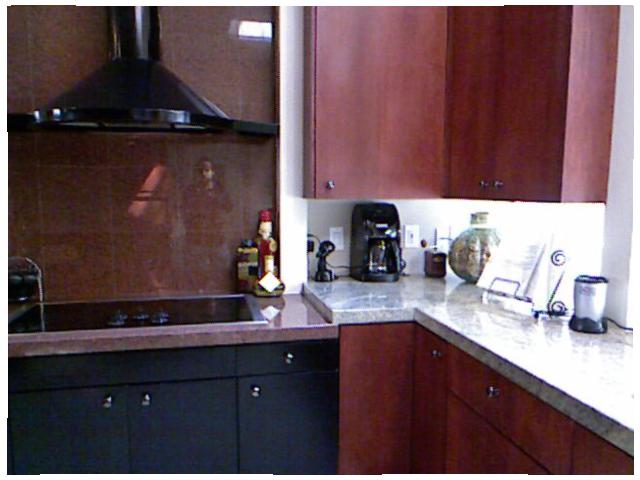} &
    \addFig{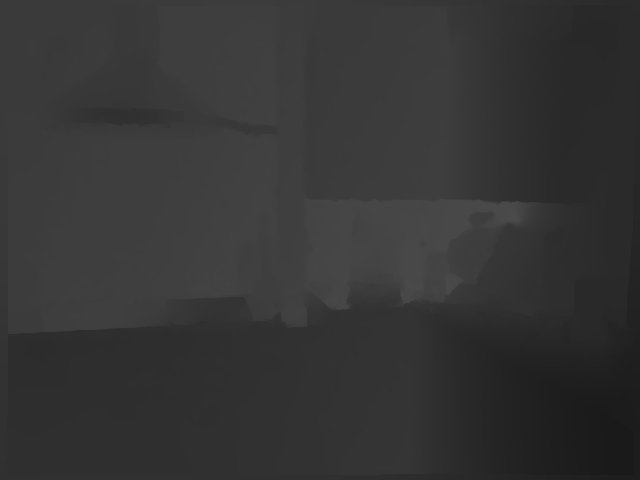} &
    \addFig{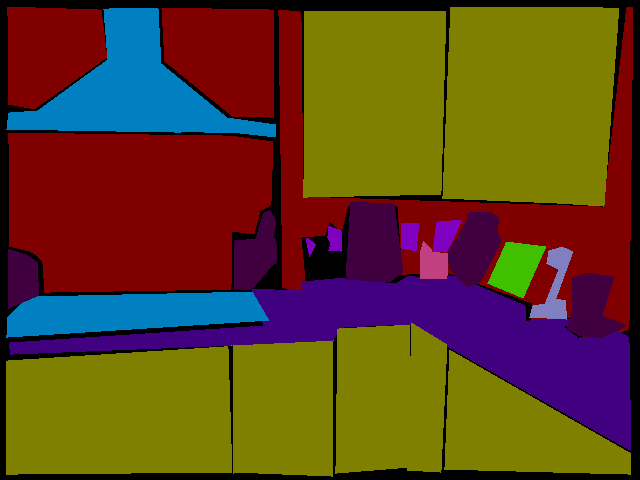} &
    \addFig{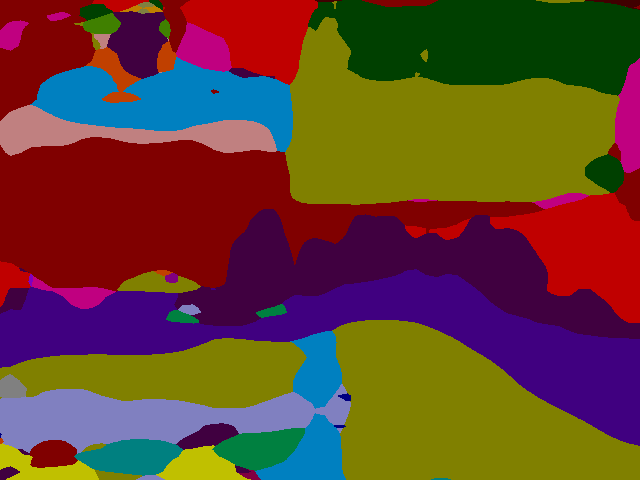} &
    \addFig{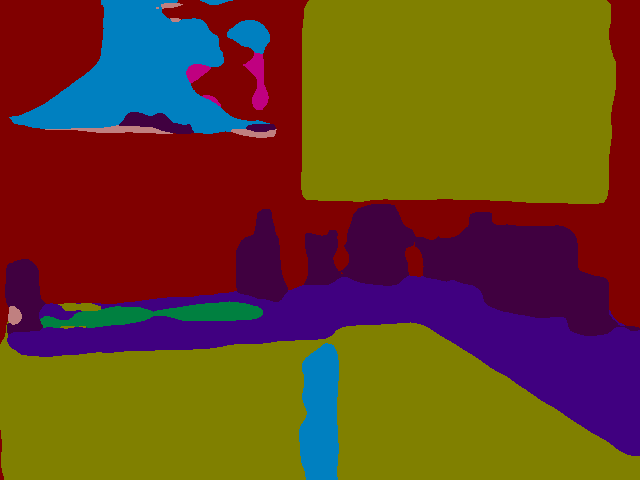} &
    \addFig{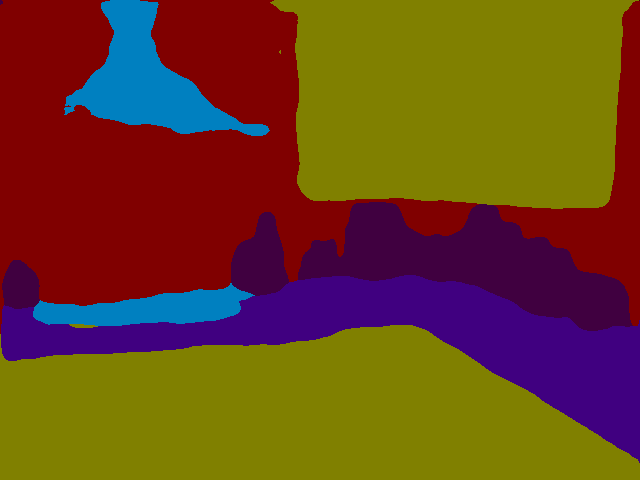} \\
    %
     (a) &(b) & (c) & (d) & (e) & (f)
    \end{tabular}
    \caption{Segmentation results for RGB-D images. The proposed CAINet delivers superior segmentation across various scenes compared to the other methods on the test set of the NYU-Depth V2 dataset. (a) RGB image, (b) Depth image, (c) GT, (d) RDFNet~\cite{2017RDFNet}, (e) SA-Gate~\cite{2020SA-Gate}, (f) CAINet.}
\label{fig:nuy}
\end{figure}

\subsection{Generalization to RGB-D Data}
\label{Generalization}
To assess the generalization ability of CAINet, we conduct training and testing using the NYU-Depth V2 dataset~\cite{silberman2012indoor}. 
In addition, we compare our results with those of other multimodal semantic segmentation methods, taking depth images instead of thermal images as input. The comparative results are presented in Tab.~\ref{table:rgbd}. CAINet demonstrates superior performance, indicating its applicability to RGB-D data. Many RGB-D segmentation methods struggle with issues stemming from the limitations of depth image capture devices, such as sensitivity to lighting conditions, which often results in suboptimal segmentation. The proposed CAINet, with its CACR and GCM modules, effectively addresses these challenges. Some segmentation results are shown in Fig.~\ref{fig:nuy} for a visual comparison.

\section{Conclusion}
\label{sec:con}
In this paper, we review existing state-of-the-art multimodal fusion methods for RGB-T semantic segmentation, including feature fusion and feature interaction paradigms. By considering the advantages of both, we propose a new fusion paradigm, the context-aware interaction network (CAINet), in which the interaction space is constructed to exploit auxiliary tasks and global contexts for explicitly guided learning. Compared to previous work, it is more effective in tapping the complementary relationship between multimodal regions in terms of long-term dependencies in spatial and channel dimensions. Moreover, the CACR module views the inputs as a set of features and performs long-term dependencies in the interaction space. This mitigates cross-modal divergence, greatly improving the semantic segmentation performance. Furthermore, with our proposed GCM and DA modules, global context information and boundary details are well-explored. 
%
Extensive experiments and analysis demonstrate our method outperforms state-of-the-art methods on MFNet and PST900 datasets. Ablation experimental results verify the efficacy of our method. While we achieve better results with CAINet, there is room for improvement in terms of the number of parameters and computational complexity. Furthermore, if we intend to deploy CAINet on mobile devices, a more lightweight model is required. Therefore, in our future work, we will focus on investigating lightweight algorithms to better adapt to embedded platforms.



\ifCLASSOPTIONcaptionsoff
  \newpage
\fi

\bibliographystyle{IEEEtran}
\bibliography{RGBTSSref}
\vspace{-10mm}

\begin{IEEEbiography}
[{\includegraphics[width=1in,height=1.25in,clip,keepaspectratio]{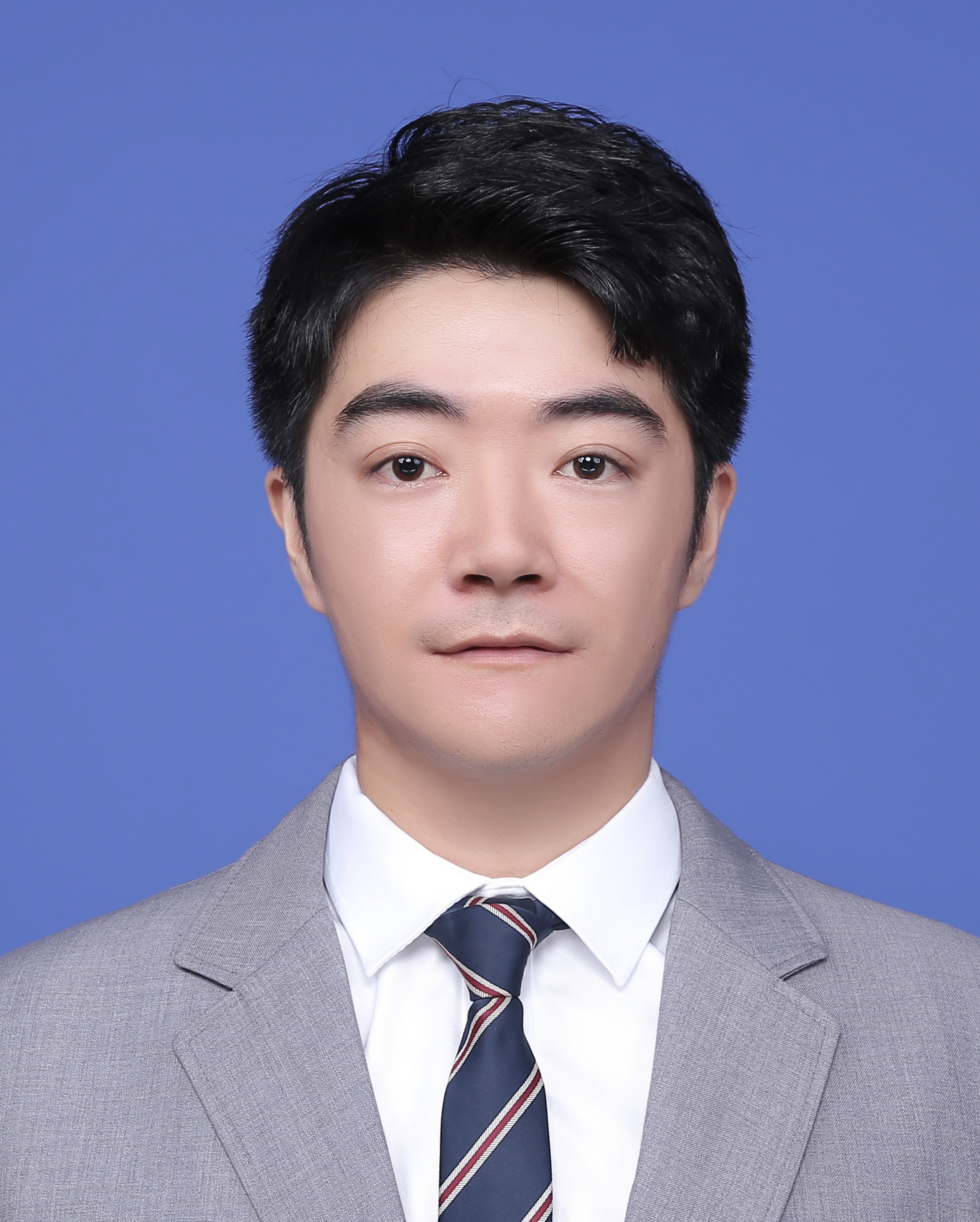}}]
{Ying Lv} is currently a Ph.D. student with the School of Communication and Information Engineering, Shanghai University, Shanghai, China. 
His main research interests include multimodal semantic segmentation and audio-visual segmentation.
\end{IEEEbiography}

\vspace{-10mm}
\begin{IEEEbiography}
[{\includegraphics[width=1in,height=1.25in,clip,keepaspectratio]{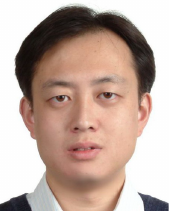}}]
{Zhi Liu} (M'07-SM'15) received the B.E. and M.E. degrees from Tianjin University, Tianjin, China, and the Ph.D. degree from the Institute of Image Processing and Pattern Recognition, Shanghai Jiao Tong University, Shanghai, China, in 1999, 2002 and 2005, respectively. He is currently a Professor at the School of Communication and Information Engineering, Shanghai University, Shanghai, China. From August 2012 to August 2014, he was a Visiting Researcher with the SIROCCO Team, IRISA/INRIA-Rennes, France, with the support by EU FP7 Marie Curie Actions. He has published more than 200 refereed technical papers in international journals and conferences. His research interests include image/video processing, machine learning, computer vision, and multimedia communication. He is an Area Editor of \textit{Signal Processing: Image Communication} and served as a Guest Editor for the special issue on \textit{Recent Advances in Saliency Models, Applications and Evaluations} in \textit{Signal Processing: Image Communication}.
\end{IEEEbiography}

\vspace{-10mm}
\begin{IEEEbiography}
[{\includegraphics[width=1in,height=1.25in,clip,keepaspectratio]{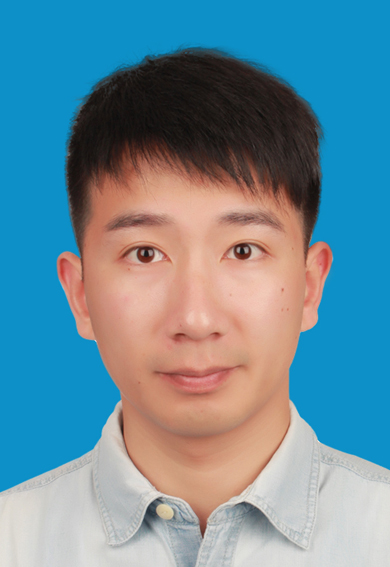}}]
{Gongyang Li} received the Ph.D. degree from Shanghai University, Shanghai, China, in 2022.
From July 2021 to June 2022, he was a Visiting Ph.D. Student with the School of Computer Science and Engineering, Nanyang Technological University, Singapore.
He is currently a Postdoctoral Fellow with the School of Communication and Information Engineering, Shanghai University, Shanghai, China. 
His research interests include image/video object segmentation, semantic segmentation, and saliency detection.
\end{IEEEbiography}

\end{document}